\documentclass[5p,times]{elsarticle}

\usepackage[utf8]{inputenc}

\journal{Robotics and Autonomous Systems}

\usepackage{graphicx}
\usepackage{setspace}

\usepackage{hyperref}
\usepackage{epstopdf}
\usepackage{float}
\usepackage{multirow,tabularx,makecell}
\usepackage{booktabs}
\usepackage{xcolor}
\usepackage[ruled,vlined]{algorithm2e}

\usepackage{fancyvrb} 

\usepackage{calc}
\makeatletter
\newcommand{\algrule}[1][.5pt]{\par\vskip.2\baselineskip\hrule height #1 width \linewidth \vskip.2\baselineskip}

\usepackage{cellspace}
\newcolumntype{D}{>{\hfill}N{3}{2}<{\hfill}}

\usepackage{wrapfig}
\usepackage[font={small}]{caption}

\usepackage{subfig}
\usepackage[export]{adjustbox}

\usepackage{color}
\usepackage{url}

\usepackage{amsmath,amsfonts,amssymb,bm}
\usepackage{nicefrac}

\usepackage[binary-units=true]{siunitx}
\DeclareSIUnit\pixel{px}
\DeclareSIUnit\milimeter{mm}
\DeclareSIUnit\radian{rad}

\usepackage[nolist,nohyperlinks]{acronym}
\acrodef{GPS}[GPS]{Global Positioning System}
\acrodef{SLAM}[SLAM]{Simultaneous Localization And Mapping}
\acrodef{SLAMs}[SLAMs]{Simultaneous Localization And Mapping systems}
\acrodef{GPS}[GPS]{Global Positioning System}
\acrodef{RTK}[RTK]{Real-time Kinematics}
\acrodef{GNSS}[GNSS]{Global Navigation Satellite System}
\acrodef{ROS}[ROS]{Robot Operating System}
\acrodef{API}[API]{Application Programming Interface}
\acrodef{UAV}[UAV]{Unmanned Aerial Vehicle}
\acrodef{UGV}[UGV]{Unmanned Ground Vehicle}
\acrodef{UV}[UV]{Ultra-Violet}
\acrodef{LED}[LED]{Light-emitting Diode}
\acrodef{MBZIRC}[MBZIRC]{Mohamed Bin Zayed International Robotics Challenge}
\acrodef{DARPA}[DARPA]{Defense Advanced Research Projects Agency}
\acrodef{IMU}[IMU]{Inertial Measurement Unit}
\acrodef{LTI}[LTI]{Linear time-invariant}
\acrodef{MPC}[MPC]{Model Predictive Control}
\acrodef{UVDAR}[UVDAR]{Ultra-Violet Direction And Ranging}
\acrodef{DOF}[DOF]{degree of freedom}
\acrodef{DOFs}[DOFs]{degrees of freedom}
\acrodef{LiDAR}[LiDAR]{Light Detection and Ranging}
\acrodef{ESC}[ESC]{Electronic Speed Controller}
\acrodef{LKF}[LKF]{Linear Kalman Filter}
\acrodef{UKF}[UKF]{Unscented Kalman Filter}
\acrodef{EKF}[EKF]{Extended Kalman Filter}
\acrodef{RAS}[RAS]{Robotics and Automation Society}
\acrodef{IEEE}[IEEE]{Institute of Electrical and Electronics Engineers}
\acrodef{MRS}[MRS]{Multi-robot Systems Group}
\acrodef{COTS}[COTS]{commercial off-the-shelf}
\acrodef{RGBD}[RGB-D]{Red-Green-Blue-Depth}
\acrodef{CTU}[CTU]{Czech Technical University}
\acrodef{UPenn}[UPenn]{University of Pennsylvania}
\acrodef{NYU}[NYU]{New York University}
\acrodef{GCS}[GCS]{Ground Control Station}
\acrodef{FOV}[FOV]{Field-of-View}
\acrodef{RMS}[RMS]{root mean square}
\acrodef{fps}[fps]{frames per second}
\acrodef{GMM}[GMM]{Gaussian Mixture Model}
\acrodef{PCA}[PCA]{Principal Component Analysis}
\acrodef{HSV}[HSV]{Hue, Saturation, Value}
\acrodef{EPM}[EPM]{Electro Permanent Magnet}
\acrodef{NMPC}[NMPC]{Nonlinear Model Predictive Controller}

\usepackage[tight]{units}

\newcommand{\minus}{\scalebox{0.75}[1.0]{$-$}}

\newcommand{\reffig}[1]{Fig.~\ref{#1}}
\newcommand{\refalg}[1]{Alg.~\ref{#1}}
\newcommand{\refsec}[1]{Sec.~\ref{#1}}
\newcommand{\reftab}[1]{Table~\ref{#1}}

\usepackage{tikz}
\usepackage{pgfplots}
\usepackage{pgfgantt}
\usetikzlibrary{backgrounds,arrows,automata,shapes,positioning,calc,through,math}
\pgfdeclarelayer{background}
\pgfdeclarelayer{foreground}
\pgfsetlayers{background,main,foreground}
\tikzset{
  imgletter/.style={
    rectangle,
    inner sep=2pt,
    rounded corners=.1em,
    text=black,
    minimum height=1em,
    text centered,
    fill=white,
    fill opacity=.7,
    text opacity=1,
    anchor=south west,
  },
}

\usepackage{hyperref}
\newcounter{node}
\makeatletter
\newcommand{\customlabel}[2]{%
\protected@write \@auxout {}{\string \newlabel {#1}{{#2}{\thepage}{#2}{#1}{}} }%
\hypertarget{#1}{#2}
}
\newcommand{\nodelabel}[1]{%
\protect\customlabel{#1}{\scriptsize[S\arabic{node}]}}
\makeatother

\begin{document}\sloppy


\begin{frontmatter}



\title{Autonomous Cooperative Wall Building by a Team of Unmanned Aerial Vehicles in the MBZIRC 2020 Competition}


\author[add,corr,eq]{Tomas Baca}
\author[add,eq]{Robert Penicka}
\author[add,eq]{Petr Stepan}
\author[add]{Matej Petrlik}
\author[add]{Vojtech Spurny}
\author[add]{Daniel Hert}
\author[add]{Martin Saska}

\address[add]{Authors are with the Faculty of Electrical Engineering, Czech Technical University in Prague, Technicka 2, Prague 160 00.}
\address[eq]{Corresponding author, e-mail: tomas.baca@fel.cvut.cz.}
\address[eq]{Authors contributed equally.}

\begin{abstract}
  This paper presents a system for autonomous cooperative wall building with a team of \acp{UAV}.
  The system was developed for Challenge~2 of the \ac{MBZIRC} 2020.
  The wall building scenario of Challenge~2 featured an initial stack of bricks and wall structure where the individual bricks had to be placed by a team of three \acp{UAV}.
  The objective of the task was to maximize collected points for placing the bricks within the restricted construction time while following the prescribed wall pattern.
  The proposed approach uses initial scanning to find a priori unknown locations of the bricks and the wall structure.
  Each \ac{UAV} is then assigned to individual bricks and wall placing locations and further perform grasping and placement using onboard resources only.
  The developed system consists of methods for scanning a given area, \acs{RGBD} detection of bricks and wall placement locations, precise grasping and placing of bricks, and coordination of multiple \acp{UAV}.
  The paper describes the overall system, individual components, experimental verification in demanding outdoor conditions, the achieved results in the competition, and lessons learned.
  The presented CTU-UPenn-NYU approach achieved the overall best performance among all participants to won the \ac{MBZIRC} competition by collecting the highest number of points by correct placement of a high number of bricks.
\end{abstract}







\end{frontmatter}



\section{Introduction}
\label{sec:introduction}

\begin{figure*}
  \centering
  \subfloat {
    \begin{tikzpicture}
      \node[anchor=south west,inner sep=0] (img) at (0,0) {\includegraphics[width=0.32\textwidth]{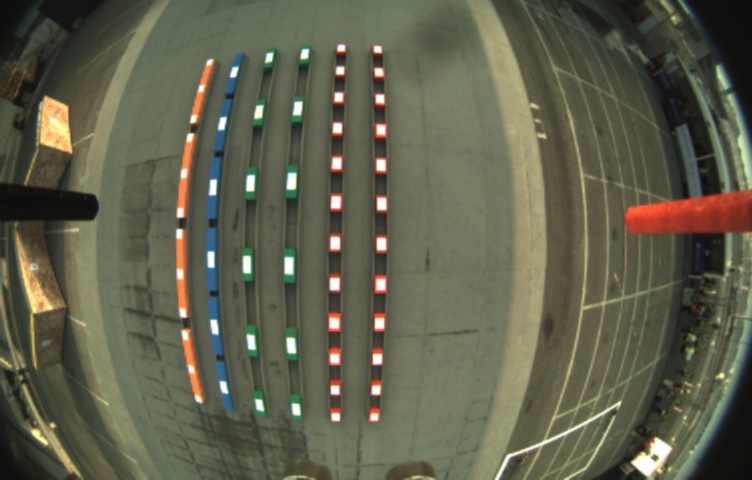}};
      \begin{scope}[x={(img.south east)},y={(img.north west)}]
        \fill[draw=black, draw opacity=0.5, fill opacity=0] (0,0) rectangle (1, 1);
        \node[imgletter,text=black] (label) at (img.south west) {(a)};
      \end{scope}
    \end{tikzpicture}}
  \hfill%
  \subfloat {\begin{tikzpicture}
    \node[anchor=south west,inner sep=0] (img) at (0,0) {\includegraphics[width=0.32\textwidth]{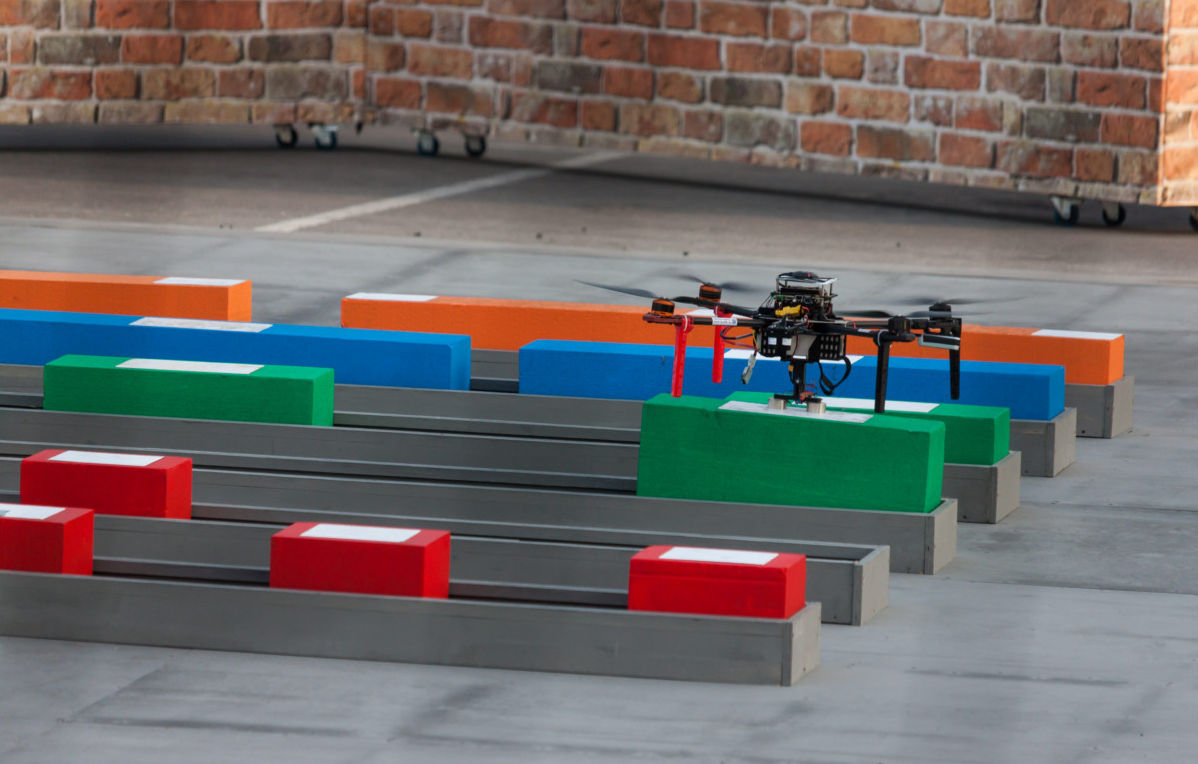}};
    \begin{scope}[x={(img.south east)},y={(img.north west)}]
      \fill[draw=black, draw opacity=0.5, fill opacity=0] (0,0) rectangle (1, 1);
      \node[imgletter,text=black] (label) at (img.south west) {(b)};
    \end{scope}
  \end{tikzpicture}}
  \hfill%
  \subfloat {\begin{tikzpicture}
    \node[anchor=south west,inner sep=0] (img) at (0,0) {\includegraphics[width=0.32\textwidth]{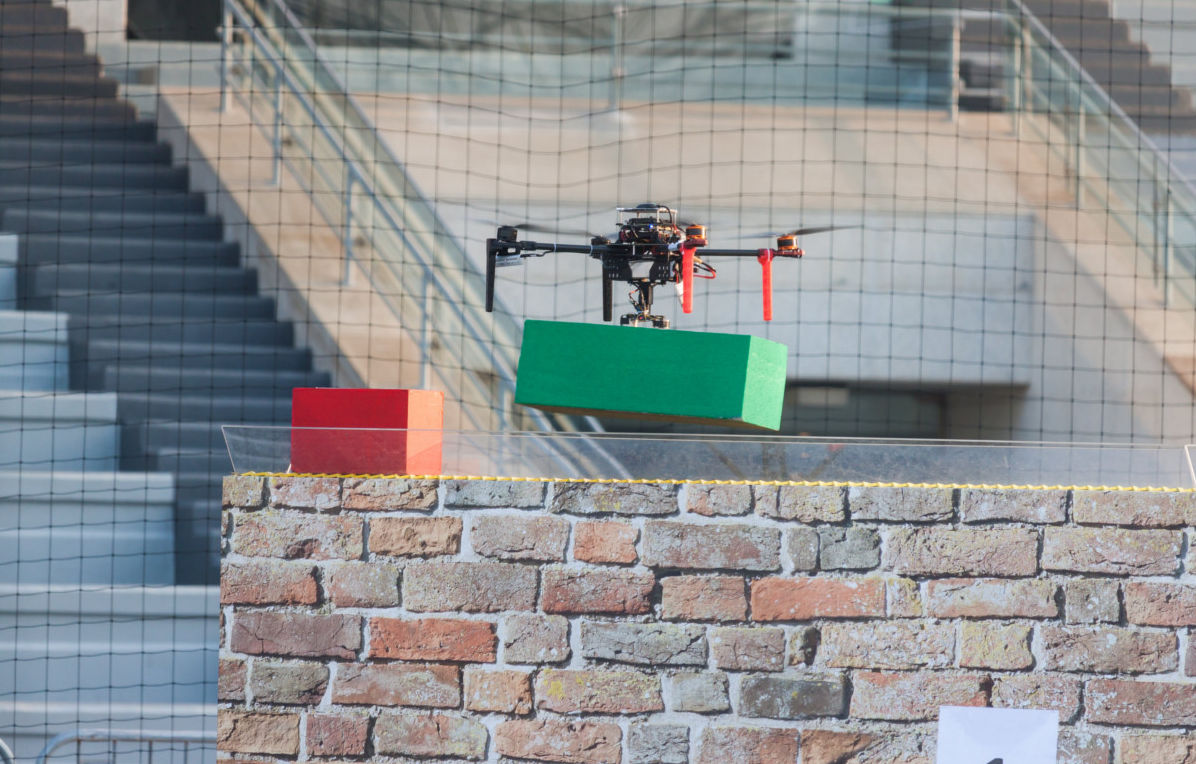}};
    \begin{scope}[x={(img.south east)},y={(img.north west)}]
      \fill[draw=black, draw opacity=0.5, fill opacity=0] (0,0) rectangle (1, 1);
      \node[imgletter,text=black] (label) at (img.south west) {(c)};
    \end{scope}
  \end{tikzpicture}}
  \caption{Illustration of the proposed \ac{UAV} system for the wall building task showing images captured from the onboard camera during scanning (a) and photos of \ac{UAV} during brick grasping (b) and placing (c).\label{fig:intro_gal}}
\end{figure*}
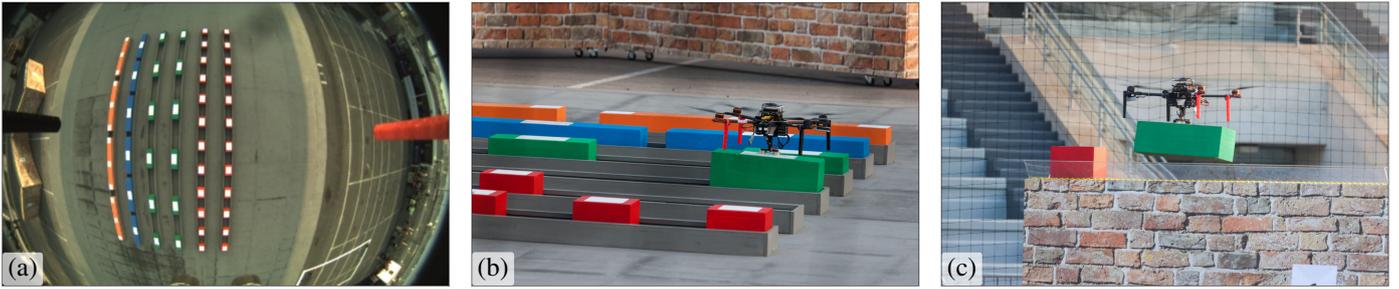

\acp{UAV} belong to one of the most studied topics in the field of robotics due to the numerous possible applications.
One of the possible areas of the \ac{UAV} deployment is in construction~\cite{Tatum2017UASapplicationsinConstruction} where the \acp{UAV} can, for example, visually inspect existing construction sites, survey areas before construction starts, or monitor security and safety of the sites~\cite{ham2016visual, Howard2018uavinconstruction}.
This paper goes beyond these works to present a fully autonomous system enabling physical interaction and not only inspection.
The \acp{UAV} directly take part in the construction and are used for building walls.
The proposed multi-robot system is designed in order to autonomously build walls with only a little a priori knowledge of the construction site.
The system uses onboard detection of bricks and wall structure locations using carried camera and depth sensors.
An initial scan of the construction area is conducted using one \ac{UAV} to find the locations of the brick stack and of the wall building site.
Afterwards, each \ac{UAV} is assigned to a particular stack part and wall segment to then cooperatively build the wall according to a given wall pattern.
During the building process, each \ac{UAV} repeatedly attempts to grasp a brick from the assigned stack, delivers the brick above the designated segment of the wall, and then precisely places the brick on the wall.
Figure~\ref{fig:intro_gal} illustrates the scanning, grasping, and placing subtasks of the wall building.


The proposed system was developed by the joint CTU-UPenn-NYU\footnote{Collaboration of the Czech Technical University in Prague, University of Pennsylvania, and New York University.} team for the participation in Challenge 2 of the \ac{MBZIRC} 2020~\cite{mbzirc_description_page}.
Challenge 2 consisted of the wall building task where three \acp{UAV} and one \ac{UGV} were assigned to autonomously build two walls --- one by the \ac{UGV} and one by the \acp{UAV}.
This paper presents details of the system used for the \ac{UAV} part of the challenge.
The challenge featured brick stacking for the \acp{UAV} containing 46 bricks, each with ferromagnetic plate on top to facilitate grasping.
Four types of bricks were present, each with different color, weight, length, and earned points for placement.
The future wall structure for the UAVs consisted of four segments arranged in a `W' letter shape located \SI{1.7}{\meter} above ground, capable of containing all the bricks from the stack in only two layers.
The goal of the challenge was to maximize collecting points by autonomously placing the bricks on the wall according to a given wall pattern in a given time limit.

The wall building approach by the CTU-UPenn-NYU team exhibited the best performance among all participants of the \ac{MBZIRC} 2020 Challenge 2.
During the two competition trials, each with a duration of 25 minutes, the \acp{UAV} were able to grasp a total number of 17 bricks and successfully place ten of them.
The UGV helped by placing one brick to fulfill the requirements for winning the challenge.
The CTU-UPenn-NYU team was thus able to place the most bricks among the participants to achieve a score of 8.24- far higher than the second best team with a score of 1.33 points.

The solution proposed for the wall building task consists of three main autonomous capabilities of the \acp{UAV}.
The first is the scanning of the arena to find locations of the brick stack and the wall.
All detections from the one \ac{UAV} performing the scanning are used to create a topological map of the arena which is created by employing statistical analysis of the detections using known sizes and shapes of the brick stack and the wall segments.
Distribution of the wall building task is then based on sharing the topological map among the \acp{UAV} using Wi-Fi communication and deterministic assignment of the brick stack and wall parts to individual \acp{UAV}.
Each \ac{UAV} then creates a plan to grasp and place according to a given wall pattern and assigned part, and afterward repeats grasping and placing until battery depletion or plan fulfillment.

The second primary capability is for brick grasping, which requires precise navigation of the UAV to the center of the brick marked by a white ferromagnetic plate.
The most essential part of ensuring precise grasping is robust and fast brick detection.
The color and \ac{RGBD} cameras provide sufficient information about brick position from altitudes above the bricks and their fusion improves the robustness of detection.
The duration of brick detection for grasping was no longer than \SI{7}{\milli\second} and thus allowed the use of a visual servoing technique during the final approach to increase grasping precision.
A grasping state machine is used to govern various stages of approaching the brick, e.g., decides when to switch from \ac{GPS} localization to visual servoing.
Finally, both the \ac{UAV} estimated mass and attitude are checked during grasping by the UAV control system to abort grasping in close brick interactions when, e.g., the brick is grasped far from the center of mass or the \ac{UAV} mass is transferred to the ground by its landing gear.

The last main autonomous capability is the placement of a brick to a desired position on the wall structure.
This task is challenging as the grasped brick may influence wall detection due to the sensors possibly being obscured by the brick. The brick may further influence the UAV control system as additional brick mass could generate torque to the UAV if not grasped exactly above the center of mass.
The brick concealing a significant amount both sensors' views is compensated immediately after successful grasping by removing such parts of sensory data during consequent placing.
As only the dimensions of the wall structure were known a priori and not its position or orientation, the \ac{RGBD} camera alone is used for wall detection.
The wall detection and computation of the placing position on the wall takes up to \SI{10}{\milli\second}.
The brick placing uses its own dedicated state machine to manage various placement stages and considers that, e.g, only a part of the wall segment can be visible and the placement is planned to be on the leftmost free position on the wall.

The visual detection for autonomous wall building with drones has to be robust, fast, and with minimal computation demands.
Detection during the scanning of the arena (i.e. looking for both bricks and walls) takes up to \SI{15}{\milli\second}, detection of brick with known color takes up to \SI{5}{\milli\second}, and the computation of brick placement on the wall takes up to \SI{10}{\milli\second}.
The wall detection pipeline first detects the ground in the \ac{RGBD} data by creating a histogram of measured distances to ground plane transformed from \ac{RGBD} data using measurements from the \ac{IMU}.
A number of highest distance values in the histogram are used as altitude measured by the \ac{RGBD} sensor.
Thresholding of the distances to ground plane using the measured altitude is then used to a create a binary image with possible wall detections.
Finally, the wall segments are verified by examining contour lines of the possible detections to be parallel and in distance approximately equal to wall width.
Visual recognition of the bricks is mainly based on white plate detection using color segmentation applied to \ac{HSV} image from the color camera.
The contours of white segments are transformed to a plane parallel to the ground plane in altitude equal to brick height.
Such transformed contours are then checked for the size of the white plates.
Finally, additional color thresholding of the \ac{HSV} image is used to identify different types of the bricks.
All detection functions take only one thread on the onboard computer, and therefore allow enough computation power for the rest of the system, e.g., control algorithms.

Automatic control of the UAV motion is vital for the precise grasping and placing of bricks.
We build upon our success from the first \ac{MBZIRC} 2017 challenge for which we developed a hybrid \ac{MPC} tracking controller \cite{baca2018model}.
An \ac{MPC} feedforward tracker is coupled with the geometric tracking controller \cite{lee2010geometric} to minimize a control error around the pre-planned differentially flat dynamics and to provide us with attitude tracking.
The tracking controller is part of the provided open-source \ac{UAV} system \cite{baca2020mrs}.
The \ac{UAV} system allows the use a visual servoing technique to estimate the states of the \ac{UAV} directly using observations of an object, i.e., the brick.
With the visual servoing, the control feedback loop is closed using only the camera-based data and the onboard \ac{IMU}.
The visual servoing removes the inaccurate \ac{GPS} localization from the loop for the duration of the grasping manoeuvre, significantly increasing the accuracy of the grasping manoeuvre to an order of centimeters.
We empirically verified that relying on a traditional \ac{GPS} introduces a significant localization position drift.
The potential \ac{GPS} drift impacts the \ac{UAV} control performance to the extent of making a precise grasping manoeuvre an unfavorable probabilistic event.
Furthermore, we employed a real-time scheduling of controller gains and dynamic constraints to satisfy the varying conditions during the various stages of the mission.
This was especially important during the transitions between the \ac{GPS} and visual servoing stages of the flight where the \ac{UAV} feedback loop exhibited different properties, mainly due to changing noise and delay characteristics of the \ac{UAV} state estimate.

In this paper, we present the overall approach and system that won, by a significant margin, Challenge 2 of the \ac{MBZIRC} 2020 in autonomously placing the most bricks on the wall.
The vision techniques for brick and wall detection that allowed for precise vision-based grasping and placing have been detailed in this paper.
Description of the \ac{UAV} control used in a closed-loop with the visual detection of the bricks and the wall is given.
The multi-UAV cooperative wall building approach is described as well, including the state machine of individual \acp{UAV}, creation of a topological map of the arena, and the deterministic distribution of the multi-robot task.
The whole system is open-sourced\footnote{\url{https://github.com/ctu-mrs/mbzirc_2020_wall_building}} to allow the community to further build on our successful system.
Finally, the results from the competition along the lessons learned are described.

The remainder of this work is organized as follows --- the rest of this section begins with an overview of related literature works and then details the \ac{MBZIRC} Challenge 2.
Section~\ref{sec:hardware} introduces the hardware platform used for the wall building task. The UAV control system is presented in \refsec{sec:uav_control_system}.
The overall approach used for the task is then described in \refsec{sec:approach}.
Sections~\ref{sec:scanning}, \ref{sec:grasping}, and~\ref{sec:placing} describe, in this order, the three most crucial parts of the wall building system: the arena scanning, brick grasping, and brick placing.
Results achieved during the competition are discussed in \refsec{sec:results} and conclusions are drawn in Section~\ref{sec:conclusions}.



\subsection{Related work}
\label{sec:related_work}

\acp{UAV} can be deployed in various scenarios in the field of construction~\cite{Tatum2017UASapplicationsinConstruction}.
Visual inspection of construction sites, area surveying prior to construction, and security and safety monitoring are examples of such tasks~\cite{Howard2018uavinconstruction}.
Inspection of existing structures, such as bridges~\cite{zink2015bridgeinspection}, can also be considered among these scenarios.
Nowadays, each of these tasks can be performed by considerably small \acp{UAV} that are manually piloted or semi-autonomous.
However, \acp{UAV} participating directly in physical construction and operating autonomously are still being considered, mainly in lab-controlled environments.

Authors of~\cite{lindsey2011construction_cubic_structures} proposed a system for building cubic truss-like structures from simple nodes by a team of \acp{UAV}.
The system relies on a motion tracking system.
Additional work on the assembly of truss structures has been explored by the authors in~\cite{lindsey2013distributed}.
The main focus is on a distributed construction algorithm to build a truss according to a given blueprint using a team of \acp{UAV}.
An approach for building tensile structures, such as structures from ropes, using \acp{UAV} is presented in~\cite{augugliaro2013building}.
The paper focuses on creating trajectories for \acp{UAV} with respect to a built structure and on the \ac{UAV} control required for building elements with tension forces.
Building bridges with cooperating \acp{UAV} using the tensile ropes is further described in~\cite{mirjan2016building}.
Trajectory planning for \acp{UAV} for assembly and structure construction is proposed in~\cite{alejo2014collisionfree}.
The authors focus on collision-free planning for multiple \acp{UAV} performing the construction task.
In~\cite{augugliaro2014flight}, a group of four \acp{UAV} are used to build a tower from foam bricks.
The paper describes the indoor application where the positions of bricks for grasping are predefined and \acp{UAV} rely solely on a motion capture system.
The system is thus very informed about its environment and serves as proof of the concept of building structure from bricks by \acp{UAV}.

Research of \acp{UAV} for assembly and construction with a main focus being on multi-robot cooperative aspects was part of the ARCAS project~\cite{ARCAS_project}.
An important capability of the \acp{UAV} for direct participation in construction is the aerial manipulation and physical interaction with structures being built.
We refer to a thorough survey on the aerial manipulation~\cite{ruggiero2018aerial}.
In~\cite{Kondak13_Aerial_arm_manipulation}, control of aerial robots interacting with other objects is examined for cases such as \acp{UAV} equipped with an arm manipulator which could perhaps be used for building more complex structures.
In~\cite{kondak2014aerial}, an autonomous aerial helicopter is also equipped with an industrial manipulator.
A controller with kinematic coupling is proposed to improve operation with the manipulator onboard the \ac{UAV}.
Fully-actuated \acp{UAV}~\cite{ryll20176d} can also be considered for construction tasks due to having higher stability during physical interaction from various tilt angles.
Authors of~\cite{munoz2015assembly} propose a planning approach for structure construction with multiple \acp{UAV} equipped with a robotic arm.
The approach is addressed by consecutive assembly planning, task allocation planning, and action planning.
In contrast with the approach proposed in this paper, none of the state-of-the art publications solve all the sub-problems required for fully-autonomous operation, i.e., visual brick detection and localization, autonomous detection of the pickup and placement locations, mission scheduling for multiple \acp{UAV}, control, state estimation, and motion planning.

The herein presented grasping approach uses the visual servoing technique that was previously mentioned for gasping in~\cite{thomas2014toward}.
The approach in~\cite{thomas2014toward} simplified the task to a one dimensional problem with an external motion capture system controlling other dimensions.
Such simplification is not possible for a real outdoor experiment.
The detection of an object for manipulation with a robotic arm is discussed in~\cite{ramon2017detection}.
In this work, a stereo camera system is used for object detection in an outdoor environment without a motion capture system.
The speed of object detection is slow, taking up to \SI{1}{\second} and unusable for \ac{UAV} control and visual servoing.
The presented work does not deal with object placing and presents only preliminary results.

In~\cite{gawel2017aerial}, an autonomous aerial pickup and delivery is approached by using a magnetic gripper.
The paper focuses on a grasping device employing \ac{EPM} and on visual servoing for precise object grasping.
The work is motivated by \ac{MBZIRC} 2017.
Similarly, the work \cite{feng2020packages} focuses on object pickup and delivery.
However, both the pickup location and the delivery location are known and marked.
Therefore, this task is similar to the gathering of ferrous objects in the \ac{MBZIRC} 2017.
The approach presented in this paper covers full visual servoing in all three dimensions.
Furthermore, we do not rely on \ac{GNSS} \ac{RTK} thanks to our robust picking mechanism that can compensate for real-world phenomenons.

Related to the previously discussed Challenge 2 of the \ac{MBZIRC} 2020 is Challenge 3 (Ch3) of the \ac{MBZIRC} 2017 which featured a treasure hunt scenario where metallic disk-shaped objects were searched for in an arena by three \acp{UAV} and collected to a common box.
In contrast to the treasure hunt scenario, the wall building task requires additional precise placement on a wall and also features a \ac{UGV} within the challenge.
However, grasping with a magnetic gripper~\cite{loianno2018localization} and required cooperation of the \ac{UAV} team are the same for both challenges.
The team lead by \ac{CTU} won the treasure hunt scenario of the \ac{MBZIRC} 2017~\cite{spurny2019cooperative}.
The approach~\cite{spurny2019cooperative} also contains initial scanning of the arena.
However, the grasping in~\cite{spurny2019cooperative} does not use the herein employed visual servoing and instead uses a more precise \ac{RTK} \ac{GPS}.
Most importantly, the collection box in the Ch3 of the \ac{MBZIRC} 2017 was in an a priori known location and of decent size.

The proposed system for the \ac{MBZIRC} 2017 Ch3 by University of Seville~\cite{castano2019robotics} uses a search phase where the arena is divided and cooperatively scanned.
A centralized \ac{GCS} is used for object detection stochastic filtering and further heuristic cooperative planning is used to assign individual UAVs to collect particular detected objects.
The \ac{GCS} also resolves potential conflicts and minimizes probability of collision.
The object detection uses color segmentation and clustering, while the grasping employs a visual‐based controller to precisely hit the target.
The drop is done using the priori known position of the dropping box.
The employed UAV platform uses standard \ac{GPS} and \ac{IMU} localization while the pickup mechanism uses \ac{EPM}~\cite{nicadrone_epmv3}.

Team from ETH Zurich~\cite{bahnemann2019eth} for Ch3 of the \ac{MBZIRC} 2017 used an approach with repeated switching between exploration and greedy pickup of the closest detected object with consequent delivery.
The exploration uses a predefined zig-zag path during the scanning of an assigned arena part and switches to pickup/delivery mode once a valid target is detected.
The system is decentralized with minimal data sharing of odometry for collision avoidance and drop box semaphore for dropping synchronization.
The object detection is based on color thresholding and a blob detector with consequent classification of blob geometrical shape features for filtering.
The detected objects are further tracked and used for pose‐based visual servoing.
The \ac{NMPC}~\cite{kamel2017robust} is used for trajectory control.
The localization is based on a combination of \ac{RTK} \ac{GPS} and visual-inertial odometry.
The grasping employs \ac{EPM}~\cite{nicadrone_epmv3} gripper with Hall effect sensors grasp feedback.

Approach of the University of Bonn~\cite{beul2019team} for the Ch3 of the \ac{MBZIRC} 2017 divided the arena into sectors with one for each drone.
Each \ac{UAV} broadcasts its position, navigation target, flight state, and detected objects outside of its own sector.
Exploration of each sector is done with one \ac{UAV} using a spiral pattern with random start.
Object detection defines the likelihood of pixels belonging to colored object to be used further in the blob detector.
The detected blobs are filtered based on blob shape and color parameters.
The approach uses visual detection of the drop box in contrast to other teams.
A variant of \ac{MPC} based on precise trajectory generation~\cite{beul2017fast} is used for controlling the UAVs.
The UAV platform uses standard \ac{GPS} and \ac{IMU}, and the grasping device uses an electromagnet on a telescopic rod with a ball joint.

The above presented systems addressing the \ac{MBZIRC} 2017 challenge \cite{bahnemann2019eth, beul2019team, castano2019robotics} including the winning solution \cite{spurny2019cooperative} do not provide sufficient mechanisms for solving the 2020 challenge, despite being state-of-the-art in the field.
The aforementioned solutions require delivery of much larger objects that pose more difficult requirements on the precision of grasping and control.
The placement of the objects is a key factor, which did not need to be solved in the previous installment of the challenge.
The 2020 challenge requires precise placement of the bricks in 3D environment, which is even more challenging due to implied higher risk for the \ac{UAV} since the \ac{UAV} is required to fly nearby a complex 3D structure.

Overall, the presented wall building task of \ac{MBZIRC} 2020 featured a very challenging scenario that required both autonomous outdoor grasping and placing using onboard sensors only.
So far, such construction tasks were restricted mostly to controlled lab environments with motion capture systems or were not entirely autonomous.
Furthermore, Challenge 2 of the \ac{MBZIRC} 2020 is more complex than the former Ch3 of the \ac{MBZIRC} 2017.
The wall in \ac{MBZIRC} 2020 has to be localized automatically due to its arbitrary position in each round and the brick placement has to be very precise for a brick to stay on the wall after placement.



\subsection{Problem overview}

\begin{figure}
  \centering
  \subfloat[Brick sizes, weights, and score points.] {
    \renewcommand{\arraystretch}{1.2}
    \begin{tabular}{c c c c}
      \noalign{\hrule}\noalign{\smallskip}
      \textbf{Brick color} & \textbf{Length}/\si{\meter} & \textbf{Weight}/\si{\kilogram} & \textbf{Score} \\
      \hline
      {\color{red} RED} & 0.3 & 1.0 & 6 \\
      {\color{green} GREEN} & 0.6 & 1.0 & 8 \\
      {\color{blue} BLUE} & 1.2 & 1.5 & 10 \\
      {\color{orange} ORANGE} & 1.8 & 2.0 & 20\\
      \hline
    \end{tabular}%
    \vphantom{\includegraphics[width=0.28\textwidth,valign=c]{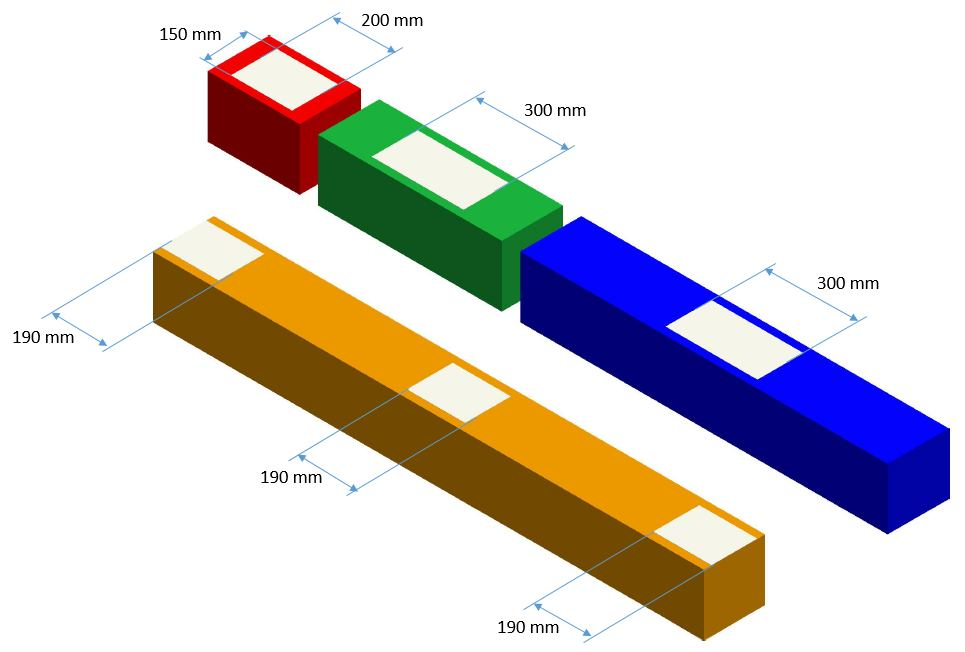}}
  }%
  \hspace{0.05\textwidth}
  \subfloat[Illustration of the bricks' visuals.] {
    \includegraphics[width=0.28\textwidth,valign=c]{./fig/bricks.png}
  }
  \caption{Parameters of the bricks present in Challenge 2 of the \ac{MBZIRC} competition~\cite{mbzirc_description_page}.}
  \label{tab:brick_parameters}
\end{figure}

Challenge 2 of the \ac{MBZIRC} 2020 featured a wall building task carried out by three \acp{UAV} and one \ac{UGV}.
Different wall placement areas and stacks of bricks with which to build were assigned for the \ac{UGV} and for the \acp{UAV}.
The layout of the wall building arena, with size of $40 \times 50 \times 20$\,\si{\meter}, can be seen in \reffig{fig:arena_layout}.
This paper concerns the \ac{UAV} part of the Challenge 2, therefore we will further focus on the challenge details concerning this part.
A total of four colored brick types --- RED, GREEN, BLUE, and ORANGE --- were available for possible placement on a wall, each with different weight, shape, and points for placing.
Table~\ref{tab:brick_parameters} summarizes length, weight, and scoring of each brick type.


The ORANGE brick could be carried and place by a single UAV or by a group of UAVs.
However, the UAV size was penalized if it exceeded a dimension limit, so a collaborative approach was encouraged.
Each brick was also equipped with a ferromagnetic white plate in the middle (and additionally to the sides of the orange brick) to be grasped by a magnetic gripper allowing for multi-robot grasping.
Initial layout of the brick stack for the \acp{UAV} was in $8 \times 4$\,\si{\meter} area with six rows of bricks where two were reserved for 24 RED bricks, two for 12 GREEN bricks, one for six BLUE bricks, and one for four ORANGE bricks.
The \ac{UGV} had a different brick stack area that was distinguishable from the \ac{UAV} stack by its properties, as later discussed in \refsec{sec:scanning}.

The wall for \acp{UAV} had a shape of the letter `W' and consisted of four segments.
Each segment was \SI{4}{\meter} long and placed on a \SI{1.7}{\meter} high base.
Convex U-shaped channels with transparent sides were attached to the top of the segments to simplify placement and to support the already placed bricks in case of wind.
The order in which the bricks were supposed to be placed on the wall was given just before the trials in order to build a wall in a given pattern.
The given wall pattern consisted of randomly ordered 4 RED, 2 GREEN and 1 BLUE brick for each layer of the first three segments.
The last channel was reserved for ORANGE bricks and can fit two such bricks per layer.
Each \ac{UAV} channel could contain two layers.
The final score was based on reward of placed bricks and was further decreased based on number of mistakes in the given wall pattern using a rather complicated formula not relevant to the approach description.
Therefore, the goal of Challenge 2 was to build as many bricks as possible according to the wall pattern within 25 minutes of the challenge trial.




\section{Hardware platform\label{sec:hardware}}



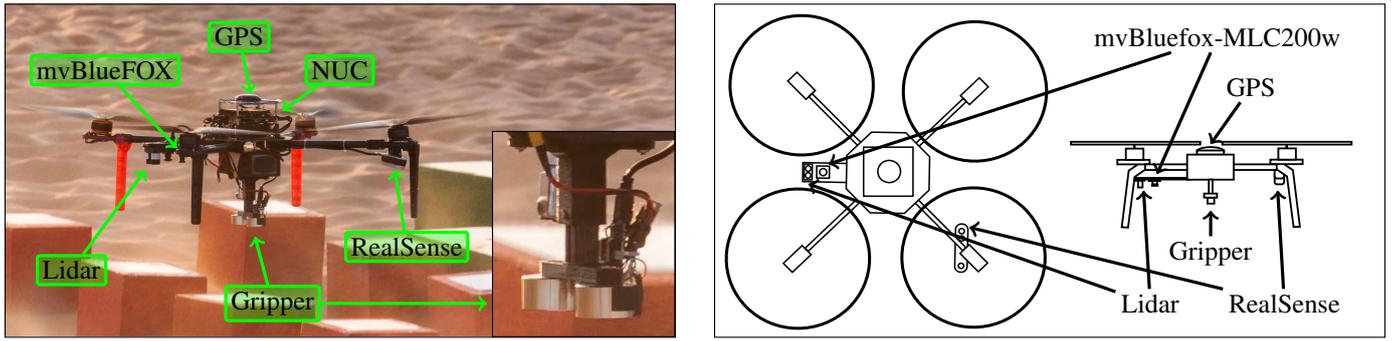
\begin{figure*}
  \centering

  \tikzset{
    partlabel/.style={
      rectangle,
      inner sep=2pt,
      rounded corners=.1em,
      draw=green, very thick,
      text=black,
      minimum height=1em,
      text centered,
      fill=green,
      fill opacity=.3,
      text opacity=1,
    },
    partarrow/.style={
      draw=green, very thick,
    },
    partpos/.style={
      color=green,
      opacity=0.0,
    },
    partlabel_black/.style={
      rectangle,
      inner sep=2pt,
      text=black,
      minimum height=1em,
      text centered,
      text opacity=1,
    },
    partarrow_black/.style={
      draw=black, very thick,
    },
    partpos_black/.style={
      color=black,
      opacity=0.0,
    },
  }

  \centering
  \subfloat{
    \begin{tikzpicture}
      \node[anchor=south west,inner sep=0] (a) at (0,0) {\includegraphics[width=0.48\textwidth]{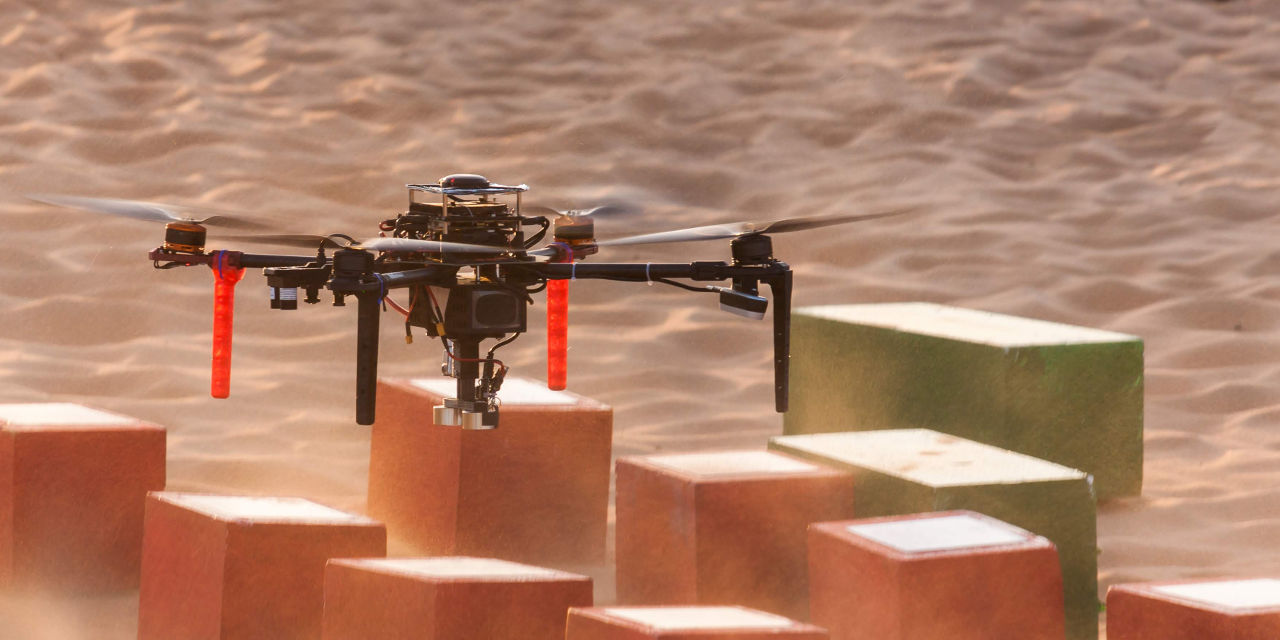}};

      \begin{scope}[x={(a.south east)},y={(a.north west)}]

        \node[anchor=south east,inner sep=0] (b) at (1.0, 0.0) {\includegraphics[width=0.13\textwidth]{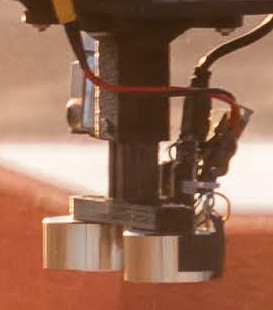}};

        \draw (b.north east) rectangle (b.south west);


        \node[partlabel] (gpslabel) at (0.35, 0.9) {GPS};
        \node[partlabel] (gripperlabel) at (0.4, 0.1) {Gripper};
        \node[partlabel] (garminlabel) at (0.1, 0.2) {Lidar};
        \node[partlabel] (bluefoxlabel) at (0.15, 0.8) {mvBlueFOX};
        \node[partlabel] (realsenselabel) at (0.6, 0.27) {RealSense};
        \node[partlabel] (nuclabel) at (0.5, 0.8) {NUC};

        \filldraw[partpos] (0.37, 0.73) circle (2pt) node[anchor=north] (gpspos) {};
        \filldraw[partpos] (0.37, 0.3) circle (2pt) node[anchor=south] (gripperpos) {};
        \filldraw[partpos] (0.22, 0.5) circle (2pt) node[anchor=south] (garminpos) {};
        \filldraw[partpos] (0.27, 0.51) circle (2pt) node[anchor=south] (bluefoxpos) {};
        \filldraw[partpos] (0.58, 0.5) circle (2pt) node[anchor=south] (realsensepos) {};
        \filldraw[partpos] (0.4, 0.63) circle (2pt) node[anchor=south] (nucpos) {};

        \draw[->,partarrow] (gpslabel) -- (gpspos) ;
        \draw[->,partarrow] (gripperlabel) -- (gripperpos) ;
        \draw[->,partarrow] (garminlabel) -- (garminpos) ;
        \draw[->,partarrow] (bluefoxlabel) -- (bluefoxpos) ;
        \draw[->,partarrow] (realsenselabel) -- (realsensepos) ;
        \draw[->,partarrow] (nuclabel) -- (nucpos) ;

        \draw[->,partarrow] (gripperlabel) -- (0.72, 0.1) ;

        \draw (current bounding box.north east) rectangle (current bounding box.south west);

      \end{scope}

    \end{tikzpicture}
  }%
  \hfill%
  \subfloat{
    \begin{tikzpicture}
      \node[anchor=south west,inner sep=0] (a) at (0,0) {\includegraphics[width=0.48\textwidth]{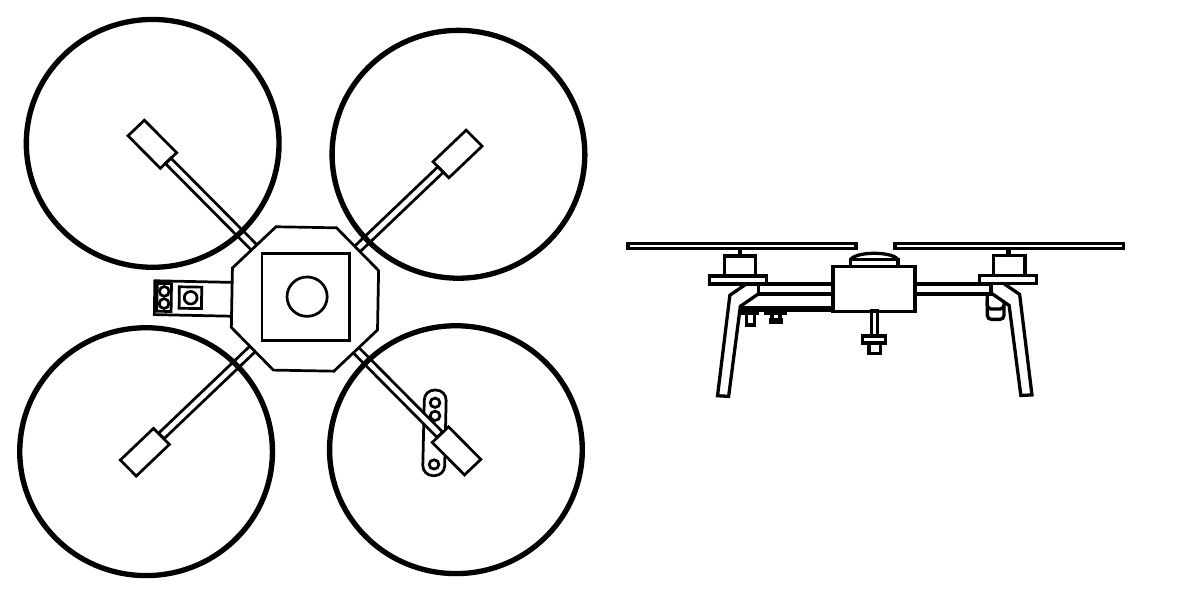}};

      \begin{scope}[x={(a.south east)},y={(a.north west)}]

        \draw (a.north east) rectangle (a.south west);

        \node[partlabel_black] (bluefoxlabel) at (0.75, 0.9) {mvBluefox-MLC200w};
        \node[partlabel_black] (gpslabel) at (0.8, 0.75) {GPS};
        \node[partlabel_black] (gripperlabel) at (0.74, 0.25) {Gripper};
        \node[partlabel_black] (lidarlabel) at (0.65, 0.10) {Lidar};
        \node[partlabel_black] (realsenselabel) at (0.85, 0.10) {RealSense};

        \draw[->,partarrow_black] (bluefoxlabel) -- (0.17, 0.52);
        \draw[->,partarrow_black] (bluefoxlabel) -- (0.66, 0.49);
        \draw[->,partarrow_black] (gpslabel) -- (0.74, 0.58);
        \draw[->,partarrow_black] (gripperlabel) -- (0.74, 0.38);
        \draw[->,partarrow_black] (lidarlabel) -- (0.64, 0.44);
        \draw[->,partarrow_black] (lidarlabel) -- (0.14, 0.46);
        \draw[->,partarrow_black] (realsenselabel) -- (0.84, 0.44);
        \draw[->,partarrow_black] (realsenselabel) -- (0.38, 0.33);

      \end{scope}

    \end{tikzpicture}
  }
  \caption{UAV platform for the brick challenge.\label{fig:platform}}
\end{figure*}


This section describes the \ac{UAV} platform shown in \reffig{fig:platform} which was used for all \acp{UAV} deployed by the CTU-UPenn-NYU team in Challenge 2 of the \ac{MBZIRC} 2020.

The utilized \ac{UAV} quadrotor platform is composed of only \ac{COTS} parts and rather inexpensive components and sensors.
The brick and wall detection relies on one fish-eye color camera and one \ac{RGBD} camera.
The global localization of the \acp{UAV} in the arena is based on a standard \ac{GPS} receiver accompanied by \ac{LiDAR} sensor for measuring altitude.
The grasping was done using an in-house designed electromagnetic gripper with grasp feedback sensors.
Finally, the basic stability of the platform was controlled by \ac{COTS} flight controller governed by an onboard miniature computer that was used for all computations and autonomy during the wall building task.

The platform is based on the \emph{Tarot 650 Sport} quadrotor frame with four \emph{Tarot 4114 320Kv} motors, each connected to \emph{BLheli32 51A} electronic speed controller and equipped with a 15-inch carbon fiber propeller.
The thrust of individual motors, and thus the lowest-level control of the platform, is governed by the \emph{PixHawk 4} flight controller which receives angular rate and total thrust commands from the control pipeline running on the onboard computer.
The primary localization system is based on the \emph{ublox Neo-M8N} \ac{GPS} receiver connected to the flight controller.
\emph{Intel NUC Kit NUC8i7BEH} with \emph{Intel i7-8559U} processor and \SI{8}{\giga\byte} of RAM are used for onboard high-level computations including calculation of control commands, high-level planning, brick and wall detection, and others.
\emph{Ubuntu 18.04 LTS} operating system is installed along with \ac{ROS}~\cite{ROSMelodic} Melodic flavor which integrates the whole UAV software system.

Apart from the \ac{GPS}-based localization, the \emph{Garmin LIDAR Lite v3} distance sensor is used to measure the \ac{UAV} altitude above ground.
Brick detection, primarily during grasping, uses the RGB \emph{mvBlueFOX-200w} camera with global shutter, $752 \times 480$\,\si{\pixel} resolution, and up to 93 \ac{fps}.
The fish-eye camera lens \emph{Sunex DSL215} is used to significantly enlarge the footprint of the camera on the ground.
The camera is set to 20 \ac{fps}, which is a sufficient value for visual servoing during grasping.
To avoid obstruction of both the \ac{LiDAR} and the mvBlueFOX camera by the grasped brick, both sensors are placed on the left side of the platform using a custom holder.
Figure~\ref{fig:platform} shows the placement of the individual sensors on the platform.
Down-facing \emph{Intel RealSense D435} \ac{RGBD} camera, with depth \ac{FOV} of $\approx 90^{\circ} \times 58^{\circ}$ and range of up to \SI{10}{\meter}, is primarily used for the wall detection.
Depth resolution of the RealSense is up to $1280 \times 720$\,\si{\pixel} with a frame rate up to 90 \ac{fps}.
For this challenge, the resolution $848 \times 480$\,\si{\pixel} is used with 30 \ac{fps}.
The RealSense camera is mounted under one of the motors and rotated towards the geometric center of the \ac{UAV}.
The mounting points of both RealSense and mvBlueFOX cameras enable navigation close above the walls and bricks.

Grasping of the bricks is done using two \emph{YJ-40/20} electromagnets, each with up to a \SI{25}{\kilogram} equivalent of holding force.
The magnets are connected to a common rod equipped with a spring mechanism along the z-axis for dampening the shocks when landing on a brick gripper-first.
Each magnet is equipped with an integrated Hall effect sensor to verify proper attachment of the ferromagnetic part of the brick to the magnet.
The electromagnets are rated to operate at \SI{12}{\volt}, however, operation at \SI{24}{\volt} is selected instead to further increase the grasping force at the cost of higher power consumption and heating.



\section{Preliminaries}
\label{sec:preliminaries}

\subsection{UAV control system\label{sec:uav_control_system}}

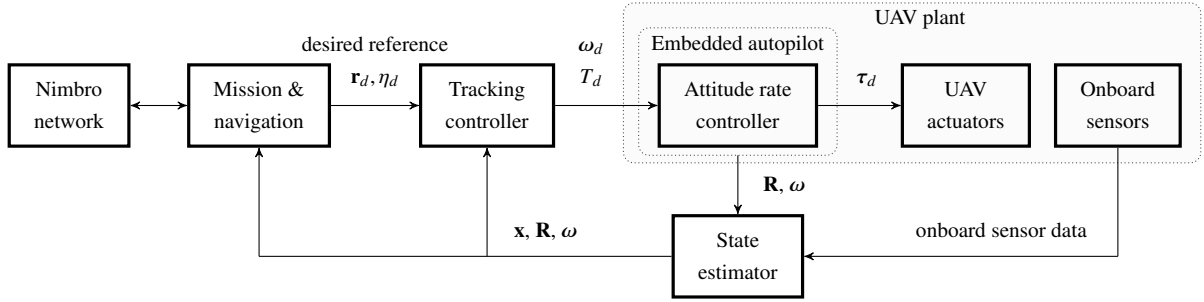
\begin{figure*}
  \centering
  \pgfdeclarelayer{foreground}
\pgfsetlayers{background,main,foreground}

\tikzset{radiation/.style={{decorate,decoration={expanding waves,angle=90,segment length=4pt}}}}


\tikzset{
  state/.style={
    rectangle,
    draw=black, very thick,
    minimum height=1.0em,
    text centered,
  },
  final_state/.style={
    rectangle,
    rounded corners,
    draw=black, very thick,
    minimum height=2em,
    text centered,
  },
  initial_state/.style={
    rectangle,
    double=white,
    double distance=1pt,
    inner sep=2pt,
    draw=black, very thick,
    minimum height=2em,
    text centered,
  },
  point/.style={
    circle,
    inner sep=0pt,
    minimum size=3pt,
    fill=red
  },
  adder/.style={
    circle,
    inner sep=2pt,
    minimum size=0.3in,
    draw=black, very thick,
    text centered
  },
  state_gray/.style={
    rectangle,
    draw=black, very thick,
    fill=gray!40,
    minimum height=1.0em,
    text centered,
    inner sep=0,
  },
  state_white/.style={
    rectangle,
    draw=black, very thick,
    fill=white,
    minimum height=1.0em,
    text centered,
    text=black,
    inner sep=0,
  },
  state_green/.style={
    rectangle,
    draw=black, very thick,
    fill=green!50,
    minimum height=1.0em,
    text centered,
    text=black,
    inner sep=0,
  },
  state_red/.style={
    rectangle,
    draw=black, very thick,
    fill=red!70,
    minimum height=1.0em,
    text centered,
    text=black,
    inner sep=0,
  },
  state_blue/.style={
    rectangle,
    draw=black, very thick,
    fill=blue!40,
    minimum height=1.0em,
    text centered,
    text=black,
    inner sep=0,
  },
  final_state/.style={
    rectangle,
    rounded corners,
    draw=black, very thick,
    minimum height=2em,
    text centered,
  },
  initial_state/.style={
    rectangle,
    double=white,
    double distance=1pt,
    inner sep=2pt,
    draw=black, very thick,
    minimum height=2em,
    text centered,
  },
  point/.style={
    circle,
    inner sep=0pt,
    minimum size=3pt,
    fill=red
  },
}


\begin{tikzpicture}[->,>=stealth', node distance=3.0cm,scale=1.0, every node/.style={scale=1.0}]


  \node[state, shift = {(0.0, 0.0)}] (navigation) {
      \begin{tabular}{c}
        \footnotesize Mission \&\\
        \footnotesize navigation
      \end{tabular}
    };

  \node[state, left of = navigation, shift = {(0.5, 0.0)}] (nimbro) {
      \begin{tabular}{c}
        \footnotesize Nimbro\\
        \footnotesize network
      \end{tabular}
    };

  \node[state, right of = navigation, shift = {(0.0, 0)}] (controller) {
      \begin{tabular}{c}
        \footnotesize Tracking\\
        \footnotesize controller
      \end{tabular}
    };

  \node[state, right of = controller, shift = {(0.3, -0)}] (attitude) {
      \begin{tabular}{c}
        \footnotesize Attitude rate\\
        \footnotesize controller
      \end{tabular}
    };

  \node[state, right of = attitude, shift = {(0.0, -0)}] (actuators) {
      \begin{tabular}{c}
        \footnotesize UAV\\
        \footnotesize actuators
      \end{tabular}
    };

  \node[state, right of = actuators, shift = {(-1.0, -0)}] (sensors) {
      \begin{tabular}{c}
        \footnotesize Onboard\\
        \footnotesize sensors
      \end{tabular}
    };

  \node[state, below of = attitude, shift = {(0, 1.0)}] (estimator) {
      \begin{tabular}{c}
        \footnotesize State\\
        \footnotesize estimator
      \end{tabular}
    };



  \path[->] ($(navigation.east) + (0.0, 0)$) edge [] node[above, midway, shift = {(0.0, 0.05)}] {
      \begin{tabular}{c}
        \footnotesize desired reference\\
        \footnotesize $\mathbf{r}_d, \eta_d$
    \end{tabular}} ($(controller.west) + (0.0, 0.00)$);

  \path[<->] ($(nimbro.east) + (0.0, 0)$) edge [] node[above, midway, shift = {(0.0, 0.05)}] {
      \begin{tabular}{c}
    \end{tabular}} ($(navigation.west) + (0.0, 0.00)$);

  \path[->] ($(controller.east) + (0.0, 0)$) edge [] node[above, midway, shift = {(-0.2, 0.05)}] {
      \begin{tabular}{c}
        \footnotesize $\bm{\omega}_d$\\
        \footnotesize $T_d$
    \end{tabular}} ($(attitude.west) + (0.0, 0.00)$);

  \path[->] ($(attitude.east) + (0.0, 0)$) edge [] node[above, midway, shift = {(0.1, 0.05)}] {
      \begin{tabular}{c}
        \footnotesize $\bm{\tau}_d$
    \end{tabular}} ($(actuators.west) + (0.0, 0.00)$);

  \path[-] ($(estimator.west)+(0, 0)$) edge [] node[above, near start, shift = {(-0.3, 0.0)}] {
      \begin{tabular}{c}
        \footnotesize $\mathbf{x}$, $\mathbf{R}$, $\bm{\omega}$
    \end{tabular}} ($(navigation.south |- estimator.west)$) -- ($(navigation.south |- estimator.west)$) edge [->,] ($(navigation.south)+(0, 0)$);


  \path[->] ($(controller.south |- estimator.west)+(0, 0)$) edge [] ($(controller.south)$);

  \path[->] ($(attitude.south) + (0.0, -0.1)$) edge [] node[right, midway, shift = {(0.0, 0.00)}] {
      \begin{tabular}{c}
        \footnotesize $\mathbf{R}$, $\bm{\omega}$
    \end{tabular}} ($(estimator.north) + (0.0, 0.00)$);

  \path[-] ($(sensors.south)+(0, 0)$) edge [] ($(sensors.south |- estimator.east)$) -- ($(sensors.south |- estimator.east)$) edge [->,] node[above, near start, shift = {(-0.5, 0.0)}] {
      \begin{tabular}{c}
        \footnotesize onboard sensor data
    \end{tabular}}($(estimator.east)$);




  \begin{pgfonlayer}{background}
    \path (attitude.west |- attitude.north)+(-0.45,0.8) node (a) {};
    \path (sensors.south -| sensors.east)+(+0.25,-0.20) node (b) {};
    \path[fill=gray!3,rounded corners, draw=black!70, densely dotted]
      (a) rectangle (b);
  \end{pgfonlayer}
  \node [rectangle, above of=actuators, shift={(-0.6,0.55)}, node distance=1.7em] (autopilot) {\footnotesize UAV plant};

  \begin{pgfonlayer}{background}
    \path (attitude.west |- attitude.north)+(-0.25,0.47) node (a) {};
    \path (attitude.south -| attitude.east)+(+0.25,-0.10) node (b) {};
    \path[fill=gray!3,rounded corners, draw=black!70, densely dotted]
      (a) rectangle (b);
  \end{pgfonlayer}
  \node [rectangle, above of=attitude, shift={(0,0.2)}, node distance=1.7em] (autopilot) {\footnotesize Embedded autopilot};


\end{tikzpicture}
  \caption{
    A diagram of the system architecture. The \emph{Mission \& navigation} part supplies position and heading reference ($\mathbf{r}_d$, $\eta_d$) to a tracking controller.
    The \emph{Tracking controller} encapsulates feed-forward tracking and feedback control techniques to produce desired thrust and angular velocities ($T_d$, $\bm{\omega}_d$) for the Pixhawk embedded flight controller.
    The \emph{State estimator} fuses data from onboard sensors to create an estimate of the \ac{UAV} translation and rotation ($\mathbf{x}$, $\mathbf{R}$).
    The \emph{Nimbro network} manages communication between the \acp{UAV} and allows execution of a coordinated multi-robot scenario.
    \label{fig:control_pipeline}
  }
\end{figure*}

Multirotor \acp{UAV} are notable for their inherently unstable dynamics.
Continual corrections to their flight need to be supplied by a feedback controller at a rate of approximately \SI{100}{\hertz} to maintain stable flight.
Moreover, automatic feedback control requires an accurate estimate of the \ac{UAV} dynamical system states.
The tasks of state estimation and feedback control are complemented by several others, such as automatic feedback reference generation, trajectory following, take-off, landing, and more.
All these vital subsystems are encapsulated in the \emph{MRS UAV System} \cite{baca2020mrs}, an open-source\footnote{\url{http://github.com/ctu-mrs/mrs_uav_system}} standalone and general control pipeline (see \reffig{fig:control_pipeline}).
The \emph{MRS UAV System} was used by the CTU-UPenn-NYU team in all the challenges of the \ac{MBZIRC} 2020 competition.
The provided framework aids deployment of autonomous \acp{UAV}, allowing focus mainly on the diverse scenarios of the competition.
It relies on the PixHawk embedded flight controller to control the \ac{UAV} attitude rate $\bm{\omega}$ and thrust $T$, while the rest of the pipeline is executed on an onboard high-level computer.
The \emph{Mission \& navigation} block, which is the core topic of this manuscript, provides the \emph{MRS UAV System} with desired trajectory references to fulfill the objectives of the challenge.

\subsection{UAV state estimation}

\begin{figure}
  \centering
  \includegraphics[width=0.40\textwidth]{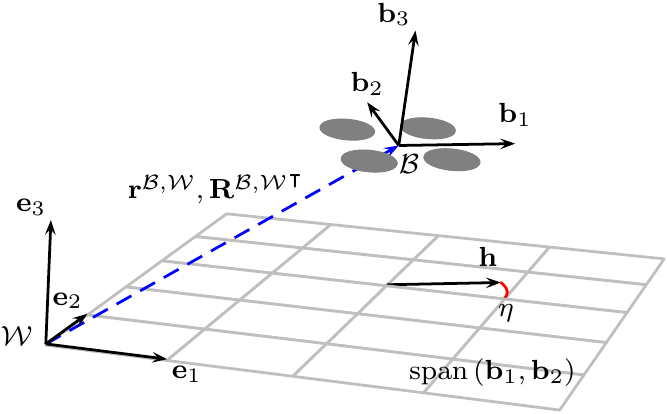}
  \caption{
    The image depicts the world frame $\mathcal{W}$ = \{$\mathbf{\hat{e}}_1$, $\mathbf{\hat{e}}_2$, $\mathbf{\hat{e}}_3$\} in which the 3D position and orientation of the \ac{UAV} body is expressed.
    The body frame $\mathcal{B}$ = \{$\mathbf{\hat{b}}_1$, $\mathbf{\hat{b}}_2$, $\mathbf{\hat{b}}_3$\} relates to $\mathcal{W}$ by translation $\mathbf{r}^{\mathcal{B},\mathcal{W}}\,=\,\left[x, y, z\right]^\intercal$ and rotation ${\mathbf{R}^{\mathcal{B},\mathcal{W}}}^{\intercal}$, respectively.
    The \ac{UAV} heading vector $\mathbf{h}$, which is a projection of $\hat{\mathbf{b}}_1$ to the plane span$\left(\mathbf{\hat{e}}_1, \mathbf{\hat{e}}_2\right)$, forms the heading angle $\eta = \mathrm{atan2}\left(\mathbf{\hat{b}}_1^\intercal\mathbf{\hat{e}}_2, \mathbf{\hat{b}}_1^\intercal\mathbf{\hat{e}}_1\right) = \mathrm{atan2}\left(\mathbf{h}_{(2)}, \mathbf{h}_{(1)}\right)$.
    }
    \label{fig:coordinate_frames}
\end{figure}

The state estimation part of the \emph{MRS UAV System} fuses data from onboard sensors into multiple independent hypotheses of the \ac{UAV} state.
In context of this particular challenge, the \ac{UAV} state is estimated using three individual estimators: a \ac{GPS}-based localization, an optic-flow odometry, and visual servoing relative to an observed brick.
These three sources of localization can be used independently depending on the particular situation.
Transitions between the stack of bricks and the wall area is made using the \ac{GPS}-based estimation and the grasping of a brick is achieved via the visual servoing.
The optic-flow estimator is used as a backup in case the visual servoing fails.
The \emph{MRS UAV System} provides a state estimate consisting of the \ac{UAV} \emph{body frame} ($\mathcal{B}$) position $\mathbf{r}^{\mathcal{B},\mathcal{W}}$ and orientation $\mathbf{R}^{\mathcal{B},\mathcal{W}}$ within a \emph{world frame} $(\mathcal{W})$.
Figure~\ref{fig:coordinate_frames} depicts the coordinate frames used within the control pipeline.
The absolute position of the \emph{world frame} depends on the actively used state estimator.
When the state estimator is changed, the control pipeline synchronizes a virtual jump between the old and new coordinate frame, such that it is not noticeable to an outside observer.

\subsection{UAV feedback control and tracking}

The \emph{tracking controller}, as depicted in \reffig{fig:control_pipeline}, encapsulates a \ac{MPC} feed-forward tracking approach \cite{baca2018model} for generating a smooth control reference and a \emph{geometric tracking controller on SE(3)} for tracking the control reference \cite{lee2010geometric}.
We also utilize an alternative \ac{MPC}-based feedback controller \cite{baca2020mrs}, when the state estimate provided by the onboard estimator might be unreliable (e.g., while grasping a brick).
The input to the control pipeline, supplied from the \emph{Mission \& navigation} block, can be a 3D position and heading reference $\left(\mathbf{r}_d, \eta_d\right)$ or a time-parametrized reference trajectory
\begin{equation}
  \left\{\left(\mathbf{r}_d, \eta_d\right)_1, \left(\mathbf{r}_d, \eta_d\right)_2, \hdots ,\left(\mathbf{r}_d, \eta_d\right)_k\right\}.
\end{equation}



\section{Autonomous multi-UAV wall building\label{sec:approach}}

This section describes the proposed high level approach for the wall building task, the state machine of individual \acp{UAV}, and the approaches used for the multi-robot coordination of the task.
The proposed approach for wall building is designed to distribute the task among the three \acp{UAV} as much as possible while mitigating possible mutual collisions.
Furthermore, we refer to the individual \acp{UAV} as UAV1, UAV2, and UAV3.
The task starts with UAV1 scanning the arena in order to find the positions of the wall and the stack of bricks for the \acp{UAV}.
The scanning process is described in detail \refsec{sec:scanning}.
After scanning, the positions of mapped wall channels and the individual brick type stacks are shared to the other \acp{UAV} (UAV2 and UAV3).
Each \ac{UAV} is then assigned to one of the three first (non-ORANGE) channels to build the bricks sequentially from one side according to the given pattern for individual channels.
The brick stack area is also divided along the longer side, such that UAV1 and UAV3 are grasping from the sides and gradually progressing to the middle of the area with each grasped brick, while UAV2 is grasping from the middle part of the area.
However, due to the \ac{UAV} stack area being a size of $8 \times 4$\,\si{\meter}, only UAV1 and UAV3 were flown simultaneously in the competition while UAV2 waited till the others have finished their mission to increase safeness.
The task then proceeds with grasping and placing according to the wall pattern assigned to each individual \acp{UAV}.
The sensor connection and battery state are checked before each grasping begins and the \ac{UAV} lands after task completion or in the case of battery depletion.
The used state machine is further described in the next section while the multi-robot aspects are detailed in \refsec{sec:multirobot_aspects}.

\subsection{UAV state machine\label{sec:state_machine}}

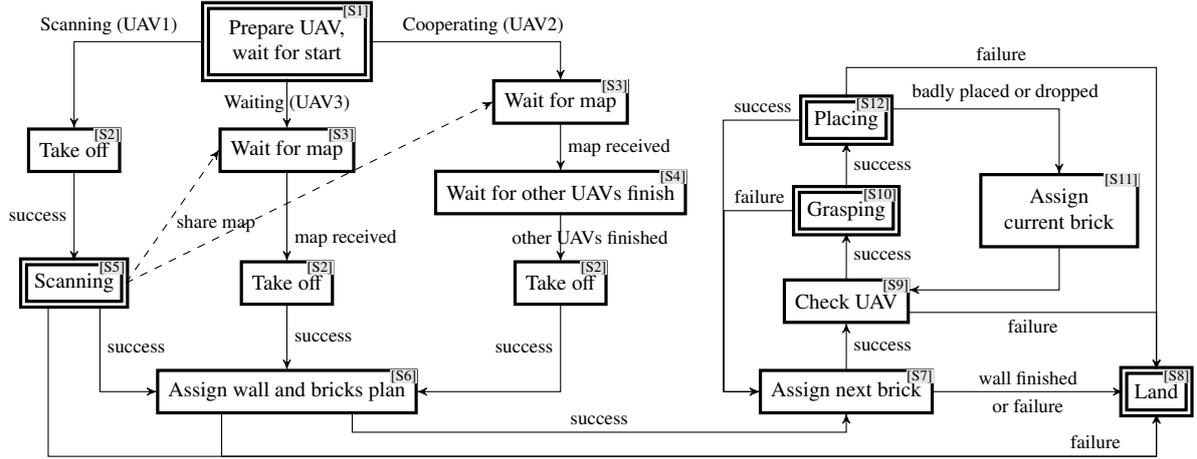
\begin{figure*}
  \centering

  \tikzset{
    state/.style={
      rectangle,
      inner sep=5pt,
      draw=black, very thick,
      minimum height=2em,
      text centered,
    },
    final_state/.style={
      rectangle,
      rounded corners,
      inner sep=5pt,
      draw=black, very thick,
      minimum height=2em,
      text centered,
      dashed,
    },
    sm_state/.style={
      rectangle,
      double=white,
      double distance=1pt,
      inner sep=5pt,
      draw=black, very thick,
      minimum height=2em,
      text centered,
    },
    initial_state/.style={
      rectangle,
      double=white,
      double distance=1pt,
      inner sep=5pt,
      draw=black, very thick,
      minimum height=2em,
      text centered,
    },
    between/.style args={#1 and #2}{
         at = ($(#1)!0.5!(#2)$)
    },
    between base/.style args={#1 and #2}{
        between=#1.base and #2.base
    },
    state_label/.style={
      fill=black!10,
      anchor=north east,
      inner sep=0.1mm,
      outer sep=0.05mm,
      at=(#1.north east)
      text centered,
    },
    new_state_label/.style={
      /utils/exec={\stepcounter{node}},
    }
  }

\begin{tikzpicture}[->,>=stealth', node distance=1.5cm,scale=0.8, every node/.style={scale=0.8}]


  \node[initial_state, shift = {(0, -0.0)}] (prepare) {\begin{tabular}{c}Prepare UAV, \\ wait for start\end{tabular}};
  \node[new_state_label,state_label=prepare] {\nodelabel{state:prepare_uav_wait}};

  \node[state,  below of = prepare, shift = {(0.0, -0.3)}] (wait_for_map) {Wait for map};
  \node[state, right of = wait_for_map, shift = {(3.0, 0.8)}] (wait_for_map_clemp) {Wait for map};
  \node[state,  left of = wait_for_map, shift = {(-2.0, -0.0)}] (takeoff_scanning) {Take off};
  \node[new_state_label,state_label=takeoff_scanning]{\nodelabel{state:takeoff}};
  \node[new_state_label,state_label=wait_for_map]{\nodelabel{state:wait_for_map}};
  \node[state_label=wait_for_map_clemp]{\ref*{state:wait_for_map}};

  \node[sm_state, below of=takeoff_scanning, shift = {(0, -0.7)}] (scanning) {Scanning};
  \node[state,  below of= wait_for_map, shift = {(0, -0.7)}] (takeoff_other) {Take off};
  \node[state,  below of = wait_for_map_clemp, shift = {(0.0, 0.0)}] (wait_for_finish) {Wait for other UAVs finish};

  \node[new_state_label,state_label=wait_for_finish]{\nodelabel{state:wait_for_finish}};
  \node[new_state_label,state_label=scanning]{\nodelabel{state:scanning}};
  \node[state_label=takeoff_other]{\ref*{state:takeoff}};

  \node[state,  below of= wait_for_finish, shift = {(0, -0)}] (takeoff_clemp) {Take off};
  \node[state_label=takeoff_clemp]{\ref*{state:takeoff}};

  \node[state, below of = takeoff_other, shift = {(0, -0.3)}] (assign_wall) {Assign wall and bricks plan};
  \node[new_state_label,state_label=assign_wall]{\nodelabel{state:assign_wall}};

  \node[state, right of = assign_wall, shift = {(7.7, -0)}] (get_next_brick) { Assign next brick};
  \node[new_state_label,state_label=get_next_brick]{\nodelabel{state:get_next_brick}};

  \node[sm_state, right of = get_next_brick ,shift = {(3.6, 0)}] (land) { Land};
  \node[new_state_label,state_label=land]{\nodelabel{state:land}};

  \node[state, above of = get_next_brick, shift = {(0, 0)}] (check_uav) { Check UAV };
  \node[new_state_label,state_label=check_uav]{\nodelabel{state:check_uav}};

  \node[sm_state, above of = check_uav, shift = {(0, 0)}] (grasping) { Grasping };
  \node[state, right of = grasping, shift = {(2., -0)}] (assign_current_brick) { \begin{tabular}{c} Assign\\ current brick \end{tabular}};
  \node[new_state_label,state_label=grasping]{\nodelabel{state:grasping}};
  \node[new_state_label,state_label=assign_current_brick]{\nodelabel{state:assign_current_brick}};

  \node[sm_state, above of = grasping, shift = {(0, 0)}] (placing) { Placing };
  \node[new_state_label,state_label=placing]{\nodelabel{state:placing}};


    \draw[->] (prepare.west) -| (takeoff_scanning.north) node[near start, above,shift = {(-0.5, 0.0)}] {\small Scanning (UAV1)};
    \draw[->] (prepare.south) -- (wait_for_map.north) node[midway,shift = {(0.0, 0.0)}] {\small Waiting (UAV3)};
    \draw[->] (prepare.east) -| (wait_for_map_clemp.north) node[near start, above,shift = {(0.3, 0.0)}] {\small  Cooperating (UAV2)};

    \draw[->] (takeoff_scanning.south) -- (scanning.north) node[midway, left] {\small success};
    \draw[->] (wait_for_map.south) -- (takeoff_other.north) node[near end, right] {\small map received};
    \draw[->] (wait_for_map_clemp.south) -- (wait_for_finish.north) node[midway, right] {\small map received};

    \draw[->,dashed] (scanning.east) -- (wait_for_map.west) node[near start, below,shift = {(1.1, 0.7)}] {\small share map};
    \draw[->,dashed] (scanning.east) -- (wait_for_map_clemp.west);
    \draw[->] (wait_for_finish.south) -- (takeoff_clemp.north) node[midway, right, shift = {(-0.92, 0.0)}] {\small other UAVs finished};

    \draw[->] ($(scanning.south)!0.5!(scanning.south east)$) |- (assign_wall.west) node[near start, right] {\small success};
    \draw[->] ($(scanning.south)!0.5!(scanning.south west)$) |- +(0,-2.5) -| (land.south) node[midway, above,shift = {(-1, 0)}] {\small failure};
    \draw[->] (takeoff_other.south) -- (assign_wall.north) node[midway, right] {\small success};
    \draw[->] (takeoff_clemp.south) |- (assign_wall.east) node[near start, left] {\small  success};

    \draw[->] ($(assign_wall.south)!0.5!(assign_wall.south east)$) |-  +(0,-0.3) -| (get_next_brick.south) node[near start, above] {\small success};
    \draw[->] ($(assign_wall.south)!0.5!(assign_wall.south west)$) |- +(0,-0.7) -| (land.south);

    \draw[->] (get_next_brick.north) -- (check_uav.south) node[midway, right] {\small success};
    \draw[->] (get_next_brick.east) -- (land.west) node[midway, above] {\small wall finished} node[midway, below] {\small or failure} ;

    \draw[->] (check_uav.north) -- (grasping.south) node[midway, right] {\small success};
    \draw[->] ($(check_uav.east)!0.5!(check_uav.south east)$) -| (land.north) node[near start, below] {\small failure} ;

    \draw[->] (grasping.north) -- (placing.south) node[midway, right] {\small success};
    \draw[->] (grasping.west) -- node[midway, above] {\small failure} +(-1.172,0) |-  (get_next_brick);
    \draw[->] (assign_current_brick.south) |- ($(check_uav.north east)!0.5!(check_uav.east)$) node[midway, above] {} ;

    \draw[->] ($(placing.north east)!0.5!(placing.east)$) -| (assign_current_brick.north) node[near start, above, shift = {(0.5, 0)}] {\small badly placed or dropped} ;
    \draw[->] (placing.north) |- +(0,0.5) -| (land) node[near start, above] {\small failure} ;
    \draw[->] (placing.west) -- node[midway, above] {\small success} +(-1.3,0) |-  (get_next_brick);

\end{tikzpicture}
  \caption{\ac{UAV} state machine for the wall building task.\label{fig:wall_sm}}
\end{figure*}

The state machine used onboard each UAV to solve the wall building task is depicted in \reffig{fig:wall_sm}.
It is implemented in FlexBE Behavior Engine~\cite{schillinger2016human} based on the state machine framework SMACH~\cite{Bohren2010SMACH}.
The whole system is integrated in the \ac{ROS}.


The state machine starts by the \emph{Prepare UAV and wait for start} procedure~\ref{state:prepare_uav_wait} that initializes all \ac{UAV} system parts, arms the \acp{UAV}, and awaits trigger from remote control to start wall building task.
The scanning UAV1 then preforms \emph{Take-off}~\ref{state:takeoff} immediately after the task starts.
Meanwhile, both UAV2 and the UAV3 are in the \emph{Wait for map} state~\ref{state:wait_for_map} where they wait for arena map shared from UAV1.
UAV1 scans the arena (\emph{Scanning} procedure~\ref{state:scanning}) detailed in \refsec{sec:scanning} then shares the four mapped wall channels and brick stack designed for the \acp{UAV}.
When the map is received by UAV3, it continues with \emph{Take-off}~\ref{state:takeoff} as it is used with UAV1 for simultaneous wall building.
UAV2 waits for the other two drones to land (\emph{Wait for finish} state~\ref{state:wait_for_finish}) before continuing with the~\emph{Take-off} and further building of the wall.
The wall building then continues with \emph{Assign wall and bricks plan} state~\ref{state:assign_wall} that contains the deterministic method that assigns different wall channels, flight altitude, and grasping positions above the brick stack to each \ac{UAV}.
This state creates a plan of individual grasping and placement attempts according to assigned channel pattern.
It is discussed more in the following~\nameref{sec:multirobot_aspects} section.
The \acp{UAV} then proceed with \emph{Assign next brick} state~\ref{state:get_next_brick} that selects the next brick according to the plan (assuming there are still bricks in the plan to be placed).
If not, the UAV switches to \emph{Land} procedure~\ref{state:land} and lands at the \ac{UAV} take off position.
Before attempting the actual grasping, the \emph{Check UAV} state~\ref{state:check_uav} makes sure that the sensors necessary for the grasping and placing tasks are connected.
Furthermore, the battery state is checked and if both the sensors and the battery are in ready-to-fly conditions, the \ac{UAV} proceeds with grasping.
Otherwise, the \ac{UAV} switches to the \emph{Land} procedure.
In the \emph{Grasping} procedure~\ref{state:grasping}, the \ac{UAV} initially flies to the mapped grasping position for the current brick type in the stack that is assigned to the particular \ac{UAV}.
Afterwards, the procedure continues to the lower-level grasping state machine that includes, e.g., visual brick servoing for precise relative positioning of the \ac{UAV} above the brick, as is described in \refsec{sec:grasping}.
In case of grasping failure, the \ac{UAV} continues with assigning the next brick from the plan in order to try, e.g., a different brick type or brick closer to the middle stack area in case of side \acp{UAV}.
When the grasping is successful, the \emph{Placing} procedure~\ref{state:placing} starts with flying (with a heading minimizing the brick air resistance) to a designated wait position next to wall using the assigned flight altitude.
It then flies to the fixed altitude above the mapped position of the assigned wall channel where a lower-level placing state machine begins.
During the placing procedure, the brick is checked for having been dropped or being badly placed. In both cases, the same brick is assigned in \emph{Assign current brick} state~\ref{state:assign_current_brick} and the \ac{UAV} proceed with check and grasping.
In case of successful placement, the wall building continues to the next brick in the plan~\ref{state:get_next_brick}.
Notice that all states~\ref{state:scanning}, \ref{state:assign_wall}, \ref{state:get_next_brick}, \ref{state:check_uav}, and \ref{state:placing} can result in failure that would switch the \ac{UAV} to landing at take-off position.
However, almost all states contain an additional hard failure mode in which the \ac{UAV} performs an emergency landing at its current place.

\subsection{Multi-robot coordination\label{sec:multirobot_aspects}}

The multi-robot coordination with three \acp{UAV} for the wall building task was proposed on two levels: the first being based on communication and the second utilizing known arena properties and thus mitigating possible \ac{UAV} collisions.

The communication between \acp{UAV} is based on \SI{5}{\giga\hertz} Wi-Fi network together with the \emph{NimbroNetwork}~\cite{nimbro} \ac{ROS} package that handles sharing certain messages over Wi-Fi.
The continuously shared messages are the predicted trajectories and diagnostics of the used onboard \ac{MPC}~\cite{baca2018model}.
The predicted trajectories are mainly used for collision avoidance purposes while the diagnostics is used as a ``heartbeat'' of the flying \acp{UAV} to be used, e.g., for triggering UAV2 take-off after other \acp{UAV} finish or stop responding.
Furthermore, the current drone positions and the arena map are shared among the \ac{UAV} team.
The arena map is shared from the scanning UAV1 once the scanning is finished and is used to proceed from the \textit{Wait for map state}~\ref{state:wait_for_map}.
The map itself contains position and rotation $(x, y, heading)$ of all four wall channels and line segments along the individual brick types.


\begin{figure*}
  \centering
  \renewcommand{\tabcolsep}{-4pt}
  \noindent\begin{tabular}{lr}
    \subfloat[Illustration of the wall building challenge arena layout\label{fig:arena_layout_tikz}]{

\tikzset{
    wall/.style={
      color = yellow,
      line width=0.3cm,
    },
    orange/.style={
      color = orange,
      line width=0.2cm,
    },
    blue/.style={
      color = blue,
      line width = 0.2cm,
    },
    green/.style={
      color = green,
      line width = 0.2cm,
    },
    red/.style={
      color = red,
      line width = 0.2cm,
    },
    redline/.style={
      color = red,
      line width = 0.02cm,
      dashed,
    },
    greenline/.style={
      color = green,
      line width = 0.02cm,
      dashed,
    },
    blueline/.style={
      color = blue,
      line width = 0.02cm,
      dashed,
    },
    orangeline/.style={
      color = orange,
      line width = 0.02cm,
      dashed,
    },
    description/.style={
      rectangle,
      double distance=1pt,
      inner sep=5pt,
      text centered,
    }
}

\trimbox{10.0 0 5.0 0}{
\scalebox{0.5}{
\begin{tikzpicture}

\tikzmath{\mvby = 9; 
          \mvugvbx = -10; 
          \mvugvby = 7.5; 
          \mvugvx = -130;  
          \mvugvxend = \mvugvx + 40;
          \mvugvxenddva = \mvugvx + 4;
          \mvugvyy=30;  
          \mvugvyyend = \mvugvyy + 4;
          \mvugvyydva = \mvugvyy - 1;
          \mvugvyyenddva = \mvugvyydva - 40;
          \mvugvxcm = 0.1 * \mvugvx;
          \mvugvyycm = 0.1 * \mvugvyy;
          }

    \node[description] at (0,1) {UAV wall};
    \draw[wall] (-4.242640687, 2.2121) -- +(-45:2cm);
    \draw[wall] (-4.242640687, 2.2121) -- +(135:2cm);
    \node[description] at (-4.942640687, 1.5121) {UAV1 channel};
    
    \draw[wall] (-1.41 , 2.0) -- +(45:2cm);
    \draw[wall] (-1.41 , 2.0) -- +(-135:2cm);
    \node[description] at (-2.11 , 2.8) {UAV2 channel};

    \draw[wall] (1.41 , 2.2121) -- +(-45:2cm);
    \draw[wall] (1.41 , 2.2121) -- +(135:2cm);
    \node[description] at (2.11 , 2.9121) {UAV3 channel};

    \draw[wall] (4.242640687,2.0) -- +(45:2cm);
    \draw[wall] (4.242640687,2.0) -- +(-135:2cm);

    \node[description] at ($(0.0, -9.5) + (0,\mvby)$) {UAV brick stack};
    \draw[orange] ($(-3.9, -10) + (0,\mvby)$) -- +(0:1.8cm);
    \draw[orange] ($(-1.9, -10) + (0,\mvby)$) -- +(0:1.8cm);
    \draw[orange] ($(0.1, -10) + (0,\mvby)$) -- +(0:1.8cm);
    \draw[orange] ($(2.1, -10) + (0,\mvby)$) -- +(0:1.8cm);
    \draw[orangeline] ($(-4.0 , -10.00) + (0,\mvby)$) -- +(8 , 0) node[right] (orangelinelabe) {};

    \draw[blue] ($(-3.9333, -10.75 ) + (0,\mvby)$) -- +(0:1.2cm);
    \draw[blue] ($( -2.5999, -10.75  ) + (0,\mvby)$) -- +(0:1.2cm);
    \draw[blue] ($(-1.2666, -10.75 ) + (0,\mvby)$) -- +(0:1.2cm);
    \draw[blue] ($(0.066, -10.75  ) + (0,\mvby)$) -- +(0:1.2cm);
    \draw[blue] ($(1.3999, -10.75  ) + (0,\mvby)$) -- +(0:1.2cm);
    \draw[blue] ($(2.7333, -10.75  ) + (0,\mvby)$) -- +(0:1.2cm);

    \draw[blueline] ($(-4.0 , -10.75) + (0,\mvby)$) -- +(8 , 0) node[right] (bluelinelabe) {};

    \draw[green] ($(-3.6333, -11.5) + (0,\mvby)$) -- +(0:0.6cm);
    \draw[green] ($(-2.3000, -11.5) + (0,\mvby)$) -- +(0:0.6cm);
    \draw[green] ($(-0.9666, -11.5) + (0,\mvby)$) -- +(0:0.6cm);
    \draw[green] ($(0.3666, -11.5) + (0,\mvby)$) -- +(0:0.6cm);
    \draw[green] ($(1.7000, -11.5) + (0,\mvby)$) -- +(0:0.6cm);
    \draw[green] ($(3.0333, -11.5) + (0,\mvby)$) -- +(0:0.6cm);
    
    \draw[green] ($(-3.6333, -12.25) + (0,\mvby)$) -- +(0:0.6cm);
    \draw[green] ($(-2.3000, -12.25) + (0,\mvby)$) -- +(0:0.6cm);
    \draw[green] ($(-0.9666, -12.25) + (0,\mvby)$) -- +(0:0.6cm);
    \draw[green] ($(0.3666, -12.25) + (0,\mvby)$) -- +(0:0.6cm);
    \draw[green] ($(1.7000, -12.25) + (0,\mvby)$) -- +(0:0.6cm);
    \draw[green] ($(3.0333, -12.25) + (0,\mvby)$) -- +(0:0.6cm);

    \draw[greenline] ($(-4.0 , -11.875) + (0,\mvby)$) -- +(8 , 0) node[right] (greenlinelabe) {};

    \draw[red] ($(-3.8166 , -13) + (0,\mvby)$) -- +(0:0.3cm);
    \draw[red] ($(-3.14999 , -13) + (0,\mvby)$) -- +(0:0.3cm);
    \draw[red] ($(-2.4833 , -13) + (0,\mvby)$) -- +(0:0.3cm);
    \draw[red] ($(-1.8166 , -13) + (0,\mvby)$) -- +(0:0.3cm);
    \draw[red] ($(-1.1499 , -13) + (0,\mvby)$) -- +(0:0.3cm);
    \draw[red] ($(-0.4833 , -13) + (0,\mvby)$) -- +(0:0.3cm);
    \draw[red] ($(0.1833 , -13) + (0,\mvby)$) -- +(0:0.3cm);
    \draw[red] ($(0.8499 , -13) + (0,\mvby)$) -- +(0:0.3cm);
    \draw[red] ($(1.5166 , -13) + (0,\mvby)$) -- +(0:0.3cm);
    \draw[red] ($(2.1833 , -13) + (0,\mvby)$) -- +(0:0.3cm);
    \draw[red] ($(2.8499 , -13) + (0,\mvby)$) -- +(0:0.3cm);
    \draw[red] ($(3.5166 , -13) + (0,\mvby)$) -- +(0:0.3cm);

    \draw[red] ($(-3.8166 , -13.75) + (0,\mvby)$) -- +(0:0.3cm);
    \draw[red] ($(-3.14999 , -13.75) + (0,\mvby)$) -- +(0:0.3cm);
    \draw[red] ($(-2.4833 , -13.75) + (0,\mvby)$) -- +(0:0.3cm);
    \draw[red] ($(-1.8166 , -13.75) + (0,\mvby)$) -- +(0:0.3cm);
    \draw[red] ($(-1.1499 , -13.75) + (0,\mvby)$) -- +(0:0.3cm);
    \draw[red] ($(-0.4833 , -13.75) + (0,\mvby)$) -- +(0:0.3cm);
    \draw[red] ($(0.1833 , -13.75) + (0,\mvby)$) -- +(0:0.3cm);
    \draw[red] ($(0.8499 , -13.75) + (0,\mvby)$) -- +(0:0.3cm);
    \draw[red] ($(1.5166 , -13.75) + (0,\mvby)$) -- +(0:0.3cm);
    \draw[red] ($(2.1833 , -13.75) + (0,\mvby)$) -- +(0:0.3cm);
    \draw[red] ($(2.8499 , -13.75) + (0,\mvby)$) -- +(0:0.3cm);
    \draw[red] ($(3.5166 , -13.75) + (0,\mvby)$) -- +(0:0.3cm);

    \draw[redline] ($(-4.0 , -13.375) + (0,\mvby)$) -- +(8 , 0) node[right] (redlinelabe) {};

    \node[description] at ($(5, -11.85) + (0,\mvby)$) (desc_mapped_bricks) {\begin{tabular}{c}Mapped\\lines of\\brick\\types\end{tabular} };

    \draw[->,dotted] (desc_mapped_bricks.west) -- (redlinelabe.west);
    \draw[->,dotted] (desc_mapped_bricks.west) -- (greenlinelabe.west);
    \draw[->,dotted] (desc_mapped_bricks.west) -- (bluelinelabe.west);
    \draw[->,dotted] (desc_mapped_bricks.west) -- (orangelinelabe.west);

    \draw[thick,dash dot] ($(-1.3333333, -9.75) + (0,\mvby)$) -- ($(-1.3333333 , -14.20) + (0,\mvby)$);
    \node[description] at ($(-2.6666666, -14.10) + (0,\mvby)$) {UAV1 part};
    \draw[thick,dash dot] ($( 1.3333333, -9.75) + (0,\mvby)$) -- ($(1.3333333 , -14.20) + (0,\mvby)$);
    \node[description] at ($(0, -14.10) + (0,\mvby)$) {UAV2 part};
    \node[description] at ($(2.6666666, -14.10) + (0,\mvby)$) {UAV3 part};

    \node[description] at ($(+2, -0.5) + (\mvugvxcm,\mvugvyycm)$) { UGV wall place};
    \foreach \x in {\mvugvx,...,\mvugvxend} \foreach \y in {\mvugvyy,...,\mvugvyyend}
    {
        \pgfmathparse{mod(\x+\y,2) ? "magenta" : "white"}
        \edef\colour{\pgfmathresult}
        \path[fill=\colour] (0.1*\x,0.1*\y) rectangle ++ (0.1,0.1);
    }
    \foreach \x in {\mvugvx,...,\mvugvxenddva} \foreach \y in {\mvugvyydva,...,\mvugvyyenddva}
    {
        \pgfmathparse{mod(\x+\y,2) ? "magenta" : "white"}
        \edef\colour{\pgfmathresult}
        \path[fill=\colour] (0.1*\x,0.1*\y) rectangle ++ (0.1,0.1);
    }

   \node[description] at ($(0.5, -9.5) + (\mvugvbx,\mvugvby)$) { UGV brick stack};
   \draw[orange] ($(2.85, -10.0 ) + (\mvugvbx,\mvugvby)$) -- +(0:1.8cm);
   \draw[orange] ($(2.85, -10.3 ) + (\mvugvbx,\mvugvby)$) -- +(0:1.8cm);
   \draw[orange] ($(2.85, -10.6 ) + (\mvugvbx,\mvugvby)$) -- +(0:1.8cm);

   \draw[blue] ($(0.85, -10.0  ) + (\mvugvbx,\mvugvby)$) -- +(0:1.2cm);
   \draw[blue] ($(0.85, -10.3 ) + (\mvugvbx,\mvugvby)$) -- +(0:1.2cm);
   \draw[blue] ($(0.85, -10.6  ) + (\mvugvbx,\mvugvby)$) -- +(0:1.2cm);

   \draw[green] ($(-1.25, -10) + (\mvugvbx,\mvugvby)$) -- +(0:0.6cm);
   \draw[green] ($(-0.55, -10) + (\mvugvbx,\mvugvby)$) -- +(0:0.6cm);
   \draw[green] ($(-1.25, -10.3) + (\mvugvbx,\mvugvby)$) -- +(0:0.6cm);
   \draw[green] ($(-0.55, -10.3) + (\mvugvbx,\mvugvby)$) -- +(0:0.6cm);
   \draw[green] ($(-1.25, -10.6) + (\mvugvbx,\mvugvby)$) -- +(0:0.6cm);
   \draw[green] ($(-0.55, -10.6) + (\mvugvbx,\mvugvby)$) -- +(0:0.6cm);

   \draw[red] ($(-3.4, -10) + (\mvugvbx,\mvugvby)$) -- +(0:0.3cm);
   \draw[red] ($(-3.0, -10) + (\mvugvbx,\mvugvby)$) -- +(0:0.3cm);
   \draw[red] ($(-2.6, -10 ) + (\mvugvbx,\mvugvby)$) -- +(0:0.3cm);
   \draw[red] ($(-2.2, -10 ) + (\mvugvbx,\mvugvby)$) -- +(0:0.3cm);
   \draw[red] ($(-3.4, -10.3) + (\mvugvbx,\mvugvby)$) -- +(0:0.3cm);
   \draw[red] ($(-3.0, -10.3) + (\mvugvbx,\mvugvby)$) -- +(0:0.3cm);
   \draw[red] ($(-2.6, -10.3 ) + (\mvugvbx,\mvugvby)$) -- +(0:0.3cm);
   \draw[red] ($(-2.2, -10.3 ) + (\mvugvbx,\mvugvby)$) -- +(0:0.3cm);
   \draw[red] ($(-3.0, -10.6 ) + (\mvugvbx,\mvugvby)$) -- +(0:0.3cm);
   \draw[red] ($(-2.6, -10.6 ) + (\mvugvbx,\mvugvby)$) -- +(0:0.3cm);
     
\end{tikzpicture}
}
}
    }
    \hspace{0.5em}
    \subfloat[Photo of the UAV arena \label{fig:arena_layout_photo}]{
      \includegraphics[width=0.41\linewidth]{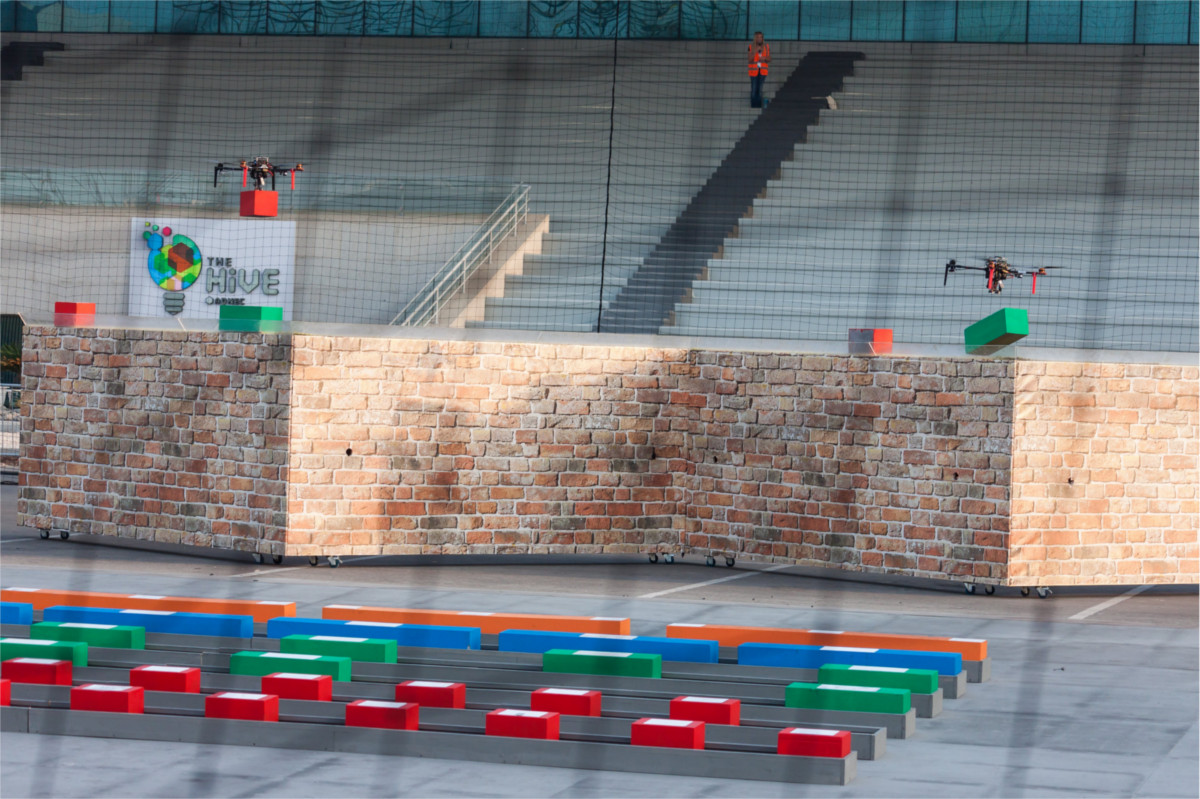}
    }
  \end{tabular}
  \caption{Wall building challenge arena layout~\protect\subref{fig:arena_layout_tikz} and photo from the competition~\protect\subref{fig:arena_layout_photo}.\label{fig:arena_layout}}
\end{figure*}


The arena properties that are used to mitigate collisions are the possible partitioning of the \ac{UAV} wall into four channels and the division of the brick stack area to three parts along the longer side.
Only the first three wall channels (non-ORANGE) are handled, each by different \ac{UAV} and filled sequentially from one side according to the given wall pattern (i.e., from the left in the case of \reffig{fig:arena_layout}).
Each \ac{UAV} has its own brick stack part where the grasping maneuver of selected brick type starts at the mapped line of that particular brick type (see \reffig{fig:arena_layout_tikz}).
UAV1 and UAV3 begin grasping initially on the outside of their brick stack parts and gradually progress to the middle.
UAV2 starts the grasping maneuver in the middle of its part.
Furthermore, the stack part of UAV1 and UAV3 are optionally swapped, minimizing the distance between a particular channel and stack part.
Finally, each \ac{UAV} has its own flight altitude ($z \in \{3.0, 4.0, 5.0\}\,${\si\meter}) that is used between the assigned stack part and wall channel.
See \reffig{fig:arena_layout_tikz} for the arena layout and partitioning of the individual wall channels and brick stack among the \acp{UAV}.



\section{Scanning for bricks and wall placement\label{sec:scanning}}

The proposed approach begins with scanning the arena as the brick stack and wall channel locations are initially unknown.
The scanning task as described throughout this section includes planning of the scanning path, detection of both the bricks and wall channels, filtering and tracking of the detections, and, finally, in creation of the topological map of the brick stack and wall channels.


\subsection{Path planning for scanning the arena\label{sec:arena_scanning_path}}

The path planning for scanning of the arena is a task of coverage path planning~\cite{CPPsurvey2013} where the entire arena has to be be covered with both the RGB BlueFOX camera and the \ac{RGBD} RealSense sensors.
The whole task is handled by one \ac{UAV} as topological map creation from multiple \acp{UAV} would require a high-bandwidth network and synchronization of all detections.
The path is composed of classical zig-zag primitives with smooth curvature constrained by turning radius $\rho$ computed as $\rho = v_{t}^2/a_{max}$.
The used turning velocity $v_{t}$ and maximal acceleration $a_{max}$ are \SI{1.5}{\meter\per\second} and \SI{2}{\meter\per\second\squared}, respectively.
Scanning speed is however \SI{3}{\meter\per\second}, meaning that the scanning \ac{UAV} is accelerating and decelerating after and before turns.
The final zig-zag path, as shown in \reffig{fig:scanning} in the \refsec{sec:results}, is then calculated with respect to the $\approx $\SI{90}{\degree} RealSense horizontal \ac{FOV} (which is smaller than the BlueFOX camera lens), and also with respect to the set scanning altitude of \SI{4.5}{\meter} (due to limitations of the RealSense distance measurements quality).



\subsection{Wall detection\label{sec:wall_detection}}

The wall detection method uses the onboard RealSense D435 sensor.
As mentioned in~\cite {realsense_depth}, the accuracy of this sensor depends on the selected resolution and parameters of the sensor.
The RealSense was dedicated for the improvement of brick detection primarily from short distances. Therefore, the image resolution was set to $848 \times 480$\,\si{\pixel} with a minimal detection distance of \SI{0.175}{\meter}.
The selected resolution has a further influence on depth accuracy. Therefore, while scanning from altitude \SI{4.5}{\meter}, the \ac{RMS} error of the distance measurements is $\approx \SI{0.6}{\meter}$.
Such measurement error is for the worst case scenario in an outdoor environment where it also depends highly on the target's texture.

The first step of the wall detection method is to find the ground plane.
As the \ac{UAV} orientation is known, it only seeks to find the \ac{UAV} height above the ground.
The arena for Challenge 2 had an almost flat surface with the brick height at \SI{0.2}{\meter}, the pillar of bricks for \ac{UGV} with 0.6--0.8\,\si{\meter} height, and a wall height of \SI{1.7}{\meter}.
The \ac{UAV} is equipped with down-pointing Garmin \ac{LiDAR} lite v3 sensor to measure \ac{UAV} height, but the sensor can also point to an obstacle.
The \ac{LiDAR} height measurement was not used during the grasping or placing of a brick.
Instead, we used a ground plane distance estimated from the RealSense stereo camera.
The orientation of the \ac{UAV} is estimated using the onboard \ac{IMU} (tilt) and magnetometer (heading).

To speed up the detection, the input depth data with resolution $848 \times 480$\,\si{\pixel} are reduced to resolution $106 \times 60$\,\si{\pixel} by selecting minimal valid value (greater than zero) from each $8 \times 8$\,\si{\pixel} sub-image.
The minimal value filter is used to reduce the size of the input data and to simultaneously remove outliers with invalid data of zero or measurements that are higher than actually possible.

The measured sensor data are then rotated to the world coordinate system.
The depth measurement $d=d\left(x, y\right)$ represents a 3D point
\begin{equation}
  \mathbf{p}=\left[d \cdot \frac{x-c_x}{f_x}, d \cdot \frac{y-c_y}{f_y}, d\right]^\intercal,
\end{equation}
where $c_x$, $c_y$, $f_x$, $f_y$ are parameters of the RealSense camera received from factory calibration.
Firstly, we transform the measurements $\mathbf{p}^{\mathcal{S}}$ from the sensor frame $\mathcal{S}$ to the UAV body frame $\mathcal{B}$ as
\begin{equation}
  \mathbf{p}^{\mathcal{B}} = \mathbf{R}^{\mathcal{S},\mathcal{B}}\mathbf{p}^{\mathcal{S}} + \mathbf{r}^{\mathcal{S},\mathcal{B}},
\end{equation}
where $\mathbf{R}^{\mathcal{S},\mathcal{B}}$ is the rotation from $\mathcal{S}$ to $\mathcal{B}$ and $\mathbf{r}^{\mathcal{S},\mathcal{B}}$ is the translation from $\mathcal{S}$ to $\mathcal{B}$.
We then similarly transform the measurement from the body frame $\mathcal{B}$ to the world frame $\mathcal{W}$:
\begin{equation}
  \mathbf{p}^{\mathcal{W}} = \mathbf{R}^{\mathcal{B},\mathcal{W}}\mathbf{p}^{\mathcal{B}} + \mathbf{r}^{\mathcal{B},\mathcal{W}}.
\end{equation}
Therefore, the final transformation is written as
\begin{align*}
  \mathbf{p}^{\mathcal{W}} &= \mathbf{R}^{\mathcal{B},\mathcal{W}}\left(\mathbf{R}^{\mathcal{S},\mathcal{B}}\mathbf{p}^{\mathcal{S}} + \mathbf{r}^{\mathcal{S},\mathcal{B}}\right) + \mathbf{r}^{\mathcal{B},\mathcal{W}}\\
  &= \mathbf{R}^{\mathcal{B},\mathcal{W}}\mathbf{R}^{\mathcal{S},\mathcal{B}}\mathbf{p}^{\mathcal{S}} + \mathbf{R}^{\mathcal{B},\mathcal{W}}\mathbf{r}^{\mathcal{S},\mathcal{B}} + \mathbf{r}^{\mathcal{B},\mathcal{W}},
\end{align*}
to obtain $\mathbf{p}^{\mathcal{W}}=\left[p_{x}^{\mathcal{W}}, p_{y}^{\mathcal{W}}, p_{z}^{\mathcal{W}}\right]^\intercal$.
However, for object detection only the $z$ component $p_{z}^{\mathcal{W}}$ is important and $\mathbf{R}^{\mathcal{B},\mathcal{W}}\mathbf{r}^{\mathcal{S},\mathcal{B}} + \mathbf{r}^{\mathcal{B},\mathcal{W}}$ is constant for one measurement.
The altitude of the point representing depth measurement $d=d\left(x, y\right)$ at pixel coordinates $\left[x, y\right]$ is defined as
\begin{equation}
  p_{z}^{\mathcal{W}} = d\cdot\mathbf{R}_{z} \left[\frac{x-c_x}{f_x}, \frac{y-c_y}{f_y}, 1\right]^\intercal+\mathbf{r},
\end{equation}
where $\mathbf{R}_z \in \mathbb{R}^3$ is the last row of the rotation matrix $\mathbf{R}=\mathbf{R}^{\mathcal{B},\mathcal{W}}\mathbf{R}^{\mathcal{S},\mathcal{B}}$ and $\mathbf{r}$ is a constant value for one measurement. This simplification speeds up computation by $3$ times.

The altitude of the \ac{UAV} is then computed from a histogram of altitudes of all points in the reduced $106 \times 60$\,\si{\pixel} image.
Based on the experimental evaluations in desert environment prior to the competition, altitude of the \ac{UAV} is estimated as a value for which more than \SI{1000}{\pixel} have a larger or equal measurement.
If the detected distance is bigger than \SI{3.5}{\meter}, the accuracy of the RealSense sensor decreases and the UAV altitude combines the Garmin \ac{LiDAR} measurement and RealSense measurement.

The next step is to create a binary image $I_{thr}$ by thresholding all pixels with an altitude higher than \SI{1}{\meter} above the ground (the height of the wall is \SI{1.7}{\meter} and the accuracy of the RealSense sensor from an altitude of \SI{5}{\meter} is \SI{0.6}{\meter}).
The following steps in the \refalg{alg:wall_detection} use the OpenCV functions \cite{opencv_library}.


\begin{algorithm}
  \caption{Wall detection}\label{alg:wall_detection}
  \small
  \KwIn{$I_{thr} - thresholded image $}
  \KwOut{$(x, y, \alpha)$ --- center of the wall segment with $\alpha$ wall rotation}
  \algrule
  \DontPrintSemicolon
  $I_{clo} = morphology\_close(I_{thr}) $ \tcp*{OpenCV functions erode, dilate}
  $Contours = findContours(I_{clo})$ \tcp*{OpenCV functions findContours}
  $Contours_{transform} = \emptyset $\;
  \For{$p \in Contours$} {
    \If{ $p$ is not at border} {Add $R \cdot p $ into list $Contours_{transform}$}
  }
  $Lines_{approx} = approxPolyDP(Contours_{transform})$ \tcp*{OpenCV functions approxPolyDP}
  \For{$l_1, l_2 \in Lines_{approx}$} {
    \If(\tcp*[f]{$l_1$ is parralel to $l_2$}){ $|l_1 \times l_2| < thr_1$ } {
      \If(\tcp*[f]{distance $l_1$ to $l_2$ is correct}){ $|l_1-l_2| \approx wall\_width$ } {
        \textbf{output} $(x, y,\alpha)$ --- $(x, y)$ is center between $l_1$ and $l_2$, $\alpha$ --- is $l_1$, $l_2$ orientation\\
      }
    }
  }
\end{algorithm}

\begin{figure}
  \centering
  \includegraphics[width=0.45\textwidth]{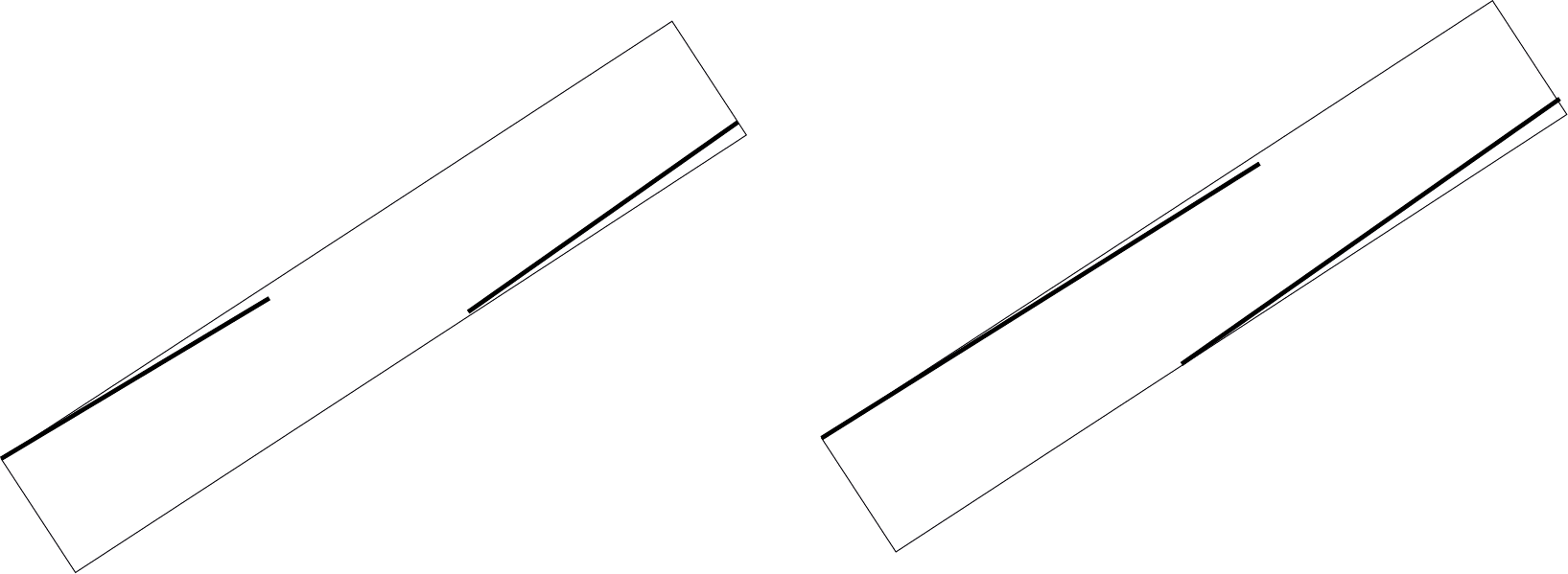}
  \caption{Positions of the wall border lines in a minimal containing rectangle.
  Wrong position of the parallel lines is show on the left and the correct position on the right.}
  \label{fig:line_position}
\end{figure}

The functions $morphology\_close, findContours$, and $approxPolyDP$ are from OpenCV library.
The last step of the \refalg{alg:wall_detection} is testing for whether the distance between two lines is correct.
This test compares the distance of endpoints of one line from the second line and vice versa, but this test does not recognize whether the lines are parallel and opposite to each other.
To test this feature, the minimal rectangle that contains all endpoints of both lines is created.
The length of this rectangle has to be less than a $0.8$ sum of both lines and the width of this rectangle should be approximately equal to wall width, 0.25--0.40 \si{\meter}.
Positions of the lines with respect to minimal rectangle is shown in \reffig{fig:line_position}.


\begin{figure*}
  \centering
  \subfloat {
    \begin{tikzpicture}
      \node[anchor=south west,inner sep=0] (img) at (0,0) {\includegraphics[height=6.2em]{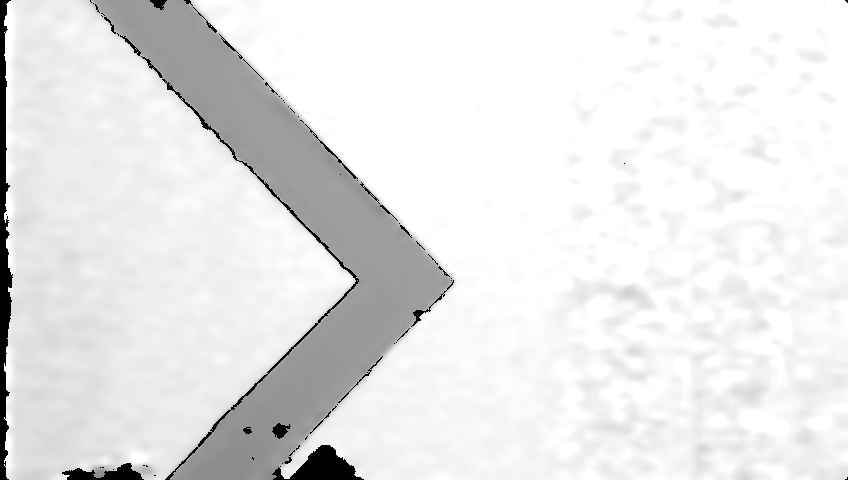}};
      \begin{scope}[x={(img.south east)},y={(img.north west)}]
        \node[imgletter,text=black] (label) at (img.south west) {(a)};
      \end{scope}
    \end{tikzpicture}
  } \subfloat {
    \begin{tikzpicture}
      \node[anchor=south west,inner sep=0] (img) at (0,0) {\includegraphics[height=6.2em]{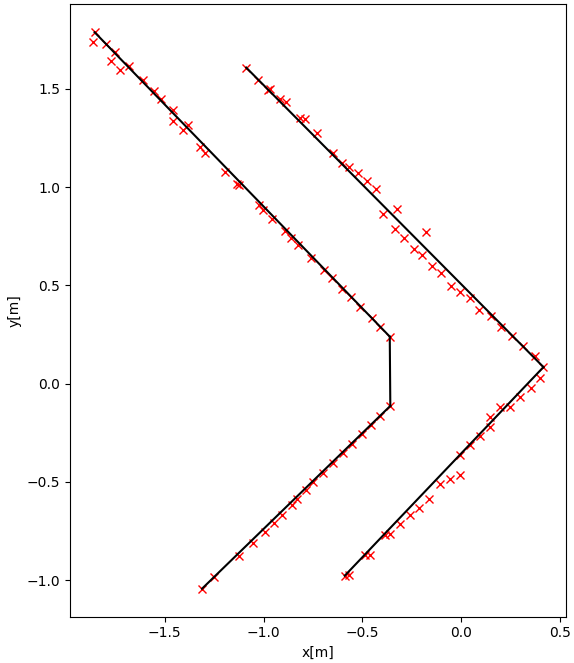}};
      \begin{scope}[x={(img.south east)},y={(img.north west)}]
        \node[imgletter,text=black] (label) at (img.south west) {(b)};
      \end{scope}
    \end{tikzpicture}
  } \subfloat{
    \begin{tikzpicture}
      \node[anchor=south west,inner sep=0] (img) at (0,0) {\includegraphics[height=6.2em]{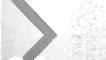}};
      \begin{scope}[x={(img.south east)},y={(img.north west)}]
        \node[imgletter,text=black] (label) at (img.south west) {(c)};
      \end{scope}
    \end{tikzpicture}
  } \subfloat{
    \begin{tikzpicture}
      \node[anchor=south west,inner sep=0] (img) at (0,0) {\includegraphics[height=6.2em]{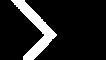}};
      \begin{scope}[x={(img.south east)},y={(img.north west)}]
        \node[imgletter,text=black] (label) at (img.south west) {(d)};
      \end{scope}
    \end{tikzpicture}
  } \subfloat{
    \begin{tikzpicture}
      \node[anchor=south west,inner sep=0] (img) at (0,0) {\includegraphics[height=6.2em]{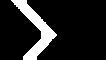}};
      \begin{scope}[x={(img.south east)},y={(img.north west)}]
        \node[imgletter,text=black] (label) at (img.south west) {(e)};
      \end{scope}
    \end{tikzpicture}
  }
  \caption{Original depth data (a) used for computing the result of the wall detection (b).
  The depth data are first filtered to lower resolution (c), then thresholded using wall height (d), and applied with the morphological \emph{closing} operation (e).}
  \label{fig:wall_detection}
\end{figure*}
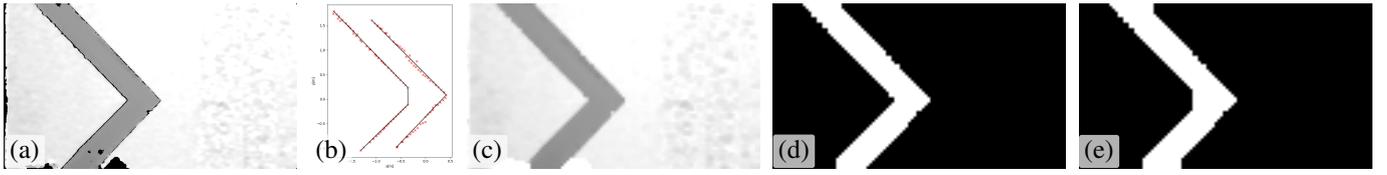

The results and various stages of the wall detection method are depicted in \reffig{fig:wall_detection}.


\subsection{Brick detection}

\label{sec:color_detection}
Detection of the bricks using the mvBlueFOX color camera is based on the white plate detection in the center of each brick.
The position of a brick in global world is based on known altitude and orientation of \ac{UAV} and the known brick height.
The white detection is based on simple color segmentation using OpenCV function \emph{inRange()} applied to a \ac{HSV} image.
The \ac{HSV} image is created by function \emph{cvtColor} (for color space conversion) from the original color camera data.
The \ac{HSV} image is further used for red, green, and blue detection.
The parameters for white, red, green, and blue segmentation are listed in \reftab{tbl:ranges}, where hue is from interval $\langle0,180\rangle$, with saturation and value from interval $\langle0,255\rangle$.


\begin{table}
  \centering
  \caption{HSV ranges for color segmentation.\label{tbl:ranges}}
  \setlength{\tabcolsep}{7pt}
  \begin{tabular}{lcccccc}
    \noalign{\hrule height 1.1pt}\noalign{\smallskip}
    & \multicolumn{2}{c}{Hue}  & \multicolumn{2}{c}{Saturation} & \multicolumn{2}{c}{Value} \\
    \cmidrule(lr){2-3} \cmidrule(lr){4-5} \cmidrule(lr){6-7}
    & min & max & min & max & min & max \\ \hline \\[-1em]
    \textit{White}& 0   & 180 & 0  & 60  & 180 & 255\\
    \textit{Red1} & 0   & 8   & 70 & 255 & 80  & 255\\
    \textit{Red2} & 160 & 180 & 70 & 255 & 80  & 255\\
    \textit{Green}& 44  & 80  & 60 & 255 & 60  & 255\\
    \textit{Blue} & 80  & 130 & 60 & 255 & 60  & 255\\
    \noalign{\smallskip}\noalign{\hrule height 1.1pt}
  \end{tabular}
\end{table}


The method that finds a white plate in the segmented image is described in \refalg{alg:white_detection}.
After segmentation, the image is processed by morphological \emph{closing} operation and the contours are computed by OpenCV \emph{findContours()} function with chain simple approximation.
All points on contours are then undistorted and transformed into a plane parallel to the ground plane with height equal to the brick height of \SI{0.2}{\meter}.
The lenses used for the color camera are very wide with a horizontal \ac{FOV} of \SI{185}{\degree}; therefore we use the \emph{Ocam toolbox}~\cite{Ocam2006} for omnidirectional cameras.
The undistortion operation is done simultaneously with the rotational transformation to the global coordinate system in order to speed up the computation.
Finally, a convex hull of points in the global coordinate system is found and used for brick classification.


\begin{algorithm}[ht]
  \caption{White\_plate\_detection}\label{alg:white_detection}
  \small
  \KwIn{$I_{thr} - thresholded image $}
  \KwOut{$(x, y, \alpha)$ --- center of the white plate with $\alpha$ plate rotation }
  \algrule
  \DontPrintSemicolon
  $I_{clo} = morphology\_close(I_{thr}) $ \tcp*{OpenCV functions erode, dilate}
  $Contours = findContours(I_{clo})$ \tcp*{OpenCV functions findContours}
  \For{$Contour \in Contours$} {
    $Points = \emptyset$\;
    \For{ $p \in Contour$ } {
      $v = undistort(p)$ \tcp*{$v$ is vector pointing in undistorted direction}
      $v' =  R \cdot v $   \tcp*{$R$ is the rotation matrix from sensor to world frame}
      $koef = UAV_{pos}(2)-brick\_height$  \tcp*{$UAV_{pos}(2)$ is altitude of the UAV}
      $p' = UAV_{pos}+v'\cdot koef$\;
      Add $p'$ into list $Points$
    }
    $Convex = convexHull(Points)$\tcp*{OpenCV functions convexHull}
    \eIf{ $Contour$ is not at border} {
      $box = minAreaRect(Convex)$ \tcp*{OpenCV functions minAreaRect}
      \If { $box$ has correct size }
      { \textbf{output} $(x, y,\alpha)$ -- $(x, y)$ is $box$ center, $\alpha$ --- is $box$ orientation\; }
    }
    {
      $Lines = approxPolyDP(Convex)$ \tcp*{only for detection from discance less than \SI{1.5}{\meter}}
      \If { $Lines$ forms U shape and size of U shape is correct}
      { find middle parallel line inside U shape and found expected brick center $(x,y)$\;
      \textbf{output} $(x, y,\alpha)$ --- $\alpha$ --- is middle line orientation\;}
    }
  }
\end{algorithm}


If the entire white segment is inside the camera image, then the correctness of detection depends only on the size of the minimal rectangle area that contains the border of white segment transformed into the world coordinate frame.
If the UAV is close to the brick, the white segment can cross the border of the image, so that the entire white plate is not in the camera image.
If the white segment forms a U shape (i.e. shape from two parallel lines and one perpendicular line) then the center of the brick can be calculated not as the center of the transformed area, but as a point with correct distance from the transformed border.

\begin{figure*}

  \tikzset{
    partlabel/.style={
      rectangle,
      inner sep=2pt,
      rounded corners=.1em,
      text=black,
      minimum height=1em,
      text centered,
      fill=white,
      fill opacity=.7,
      text opacity=1,
      anchor=south west,
    },
    partarrow/.style={
      draw=green, thick,
    },
  }

  \centering
  \subfloat {
    \begin{tikzpicture}
      \node[anchor=south west,inner sep=0] (img) at (0,0) {\includegraphics[height=5.3em]{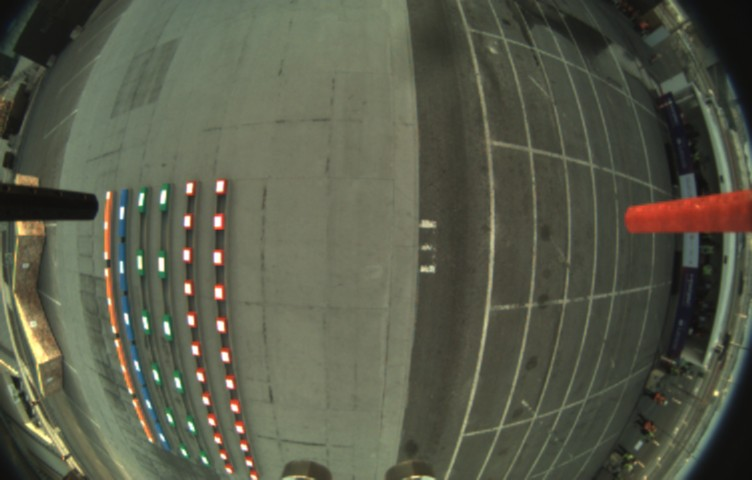}};
      \begin{scope}[x={(img.south east)},y={(img.north west)}]
        \node[imgletter,text=black] (label) at (img.south west) {(a)};
      \end{scope}
    \end{tikzpicture}
  } \subfloat {
    \begin{tikzpicture}
      \node[anchor=south west,inner sep=0] (img) at (0,0) {\includegraphics[height=5.3em]{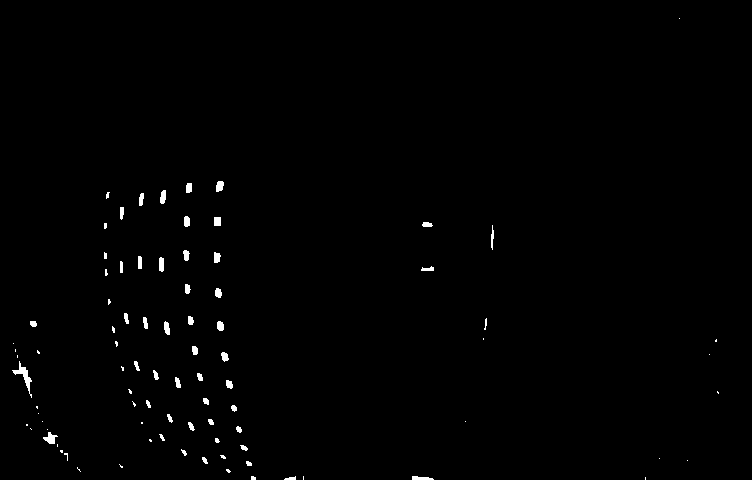}};
      \begin{scope}[x={(img.south east)},y={(img.north west)}]
        \node[imgletter,text=black] (label) at (img.south west) {(b)};
      \end{scope}
    \end{tikzpicture}
  } \subfloat {
    \begin{tikzpicture}
      \node[anchor=south west,inner sep=0] (img) at (0,0) {\includegraphics[height=5.3em]{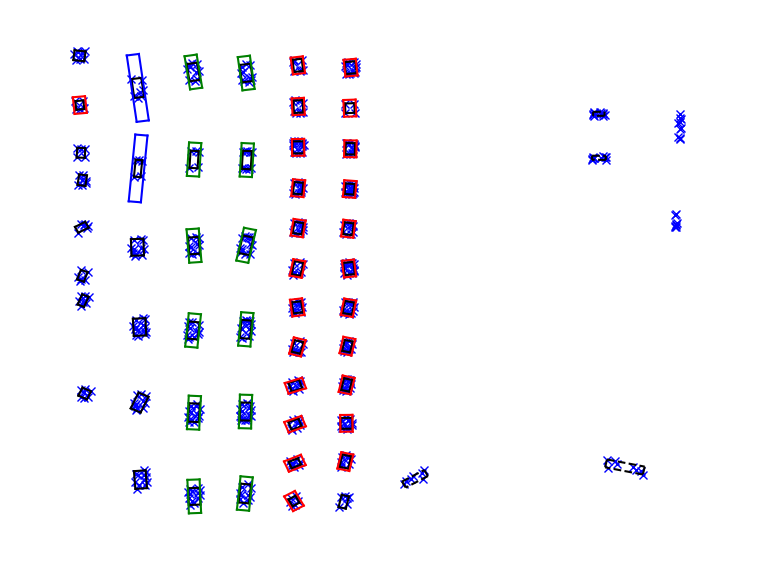}};
      \begin{scope}[x={(img.south east)},y={(img.north west)}]
        \node[imgletter,text=black] (label) at (img.south west) {(c)};
        \draw (0.0, 0.0) rectangle (1.0, 1.0);
      \end{scope}
    \end{tikzpicture}
  } \subfloat {
    \begin{tikzpicture}
      \node[anchor=south west,inner sep=0] (img) at (0,0) {\includegraphics[height=5.3em]{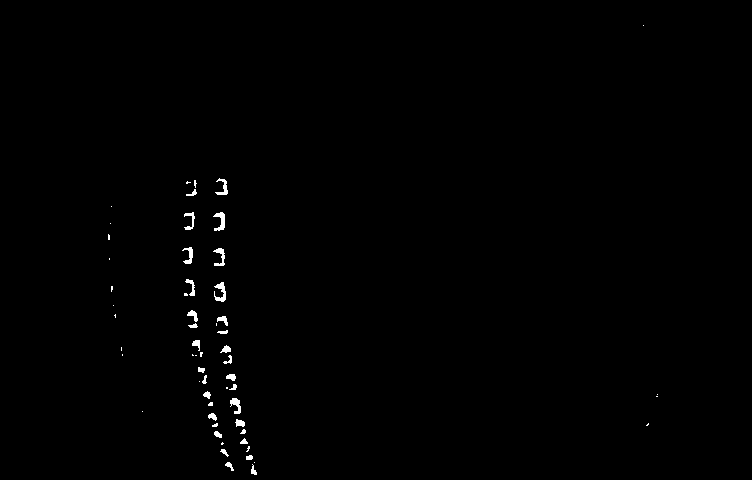}};
      \begin{scope}[x={(img.south east)},y={(img.north west)}]
        \node[imgletter,text=black] (label) at (img.south west) {(d)};
      \end{scope}
    \end{tikzpicture}
  } \subfloat {
    \begin{tikzpicture}
      \node[anchor=south west,inner sep=0] (img) at (0,0) {\includegraphics[height=5.3em]{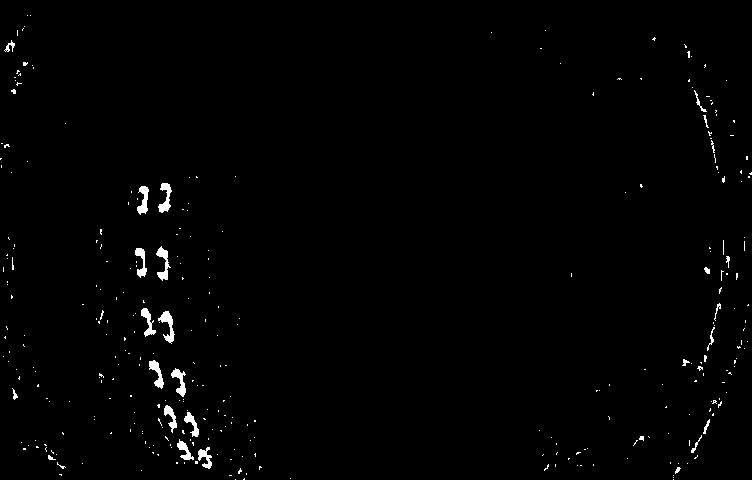}};
      \begin{scope}[x={(img.south east)},y={(img.north west)}]
        \node[imgletter,text=black] (label) at (img.south west) {(e)};
      \end{scope}
    \end{tikzpicture}
  } \subfloat {
    \begin{tikzpicture}
      \node[anchor=south west,inner sep=0] (img) at (0,0) {\includegraphics[height=5.3em]{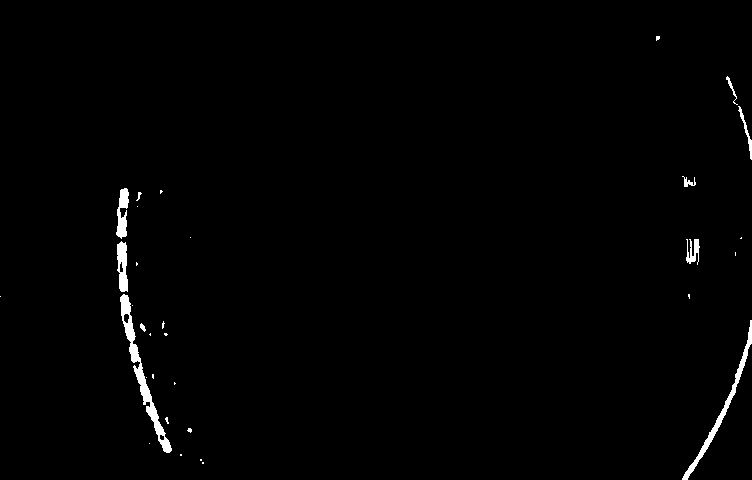}};
      \begin{scope}[x={(img.south east)},y={(img.north west)}]
        \node[imgletter,text=black] (label) at (img.south west) {(f)};
      \end{scope}
    \end{tikzpicture}
  }
  \caption{Brick detection based on an original color camera image (a) with consequent white color segmentation (b) and contour detection (c).
  The segmentation results are shown for red (d), green (e), and blue (f) colors.}
  \label{fig:small_detection}
\end{figure*}
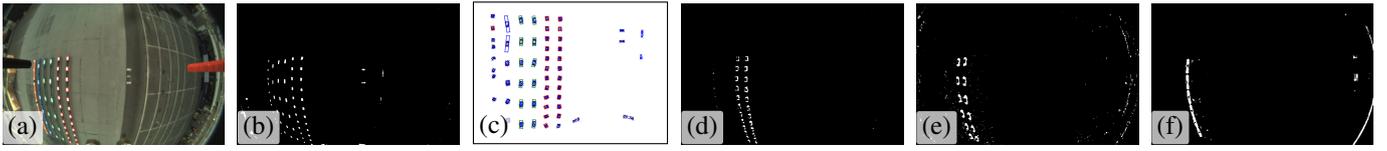

\begin{figure}
  \centering
  \includegraphics[width=0.48\textwidth]{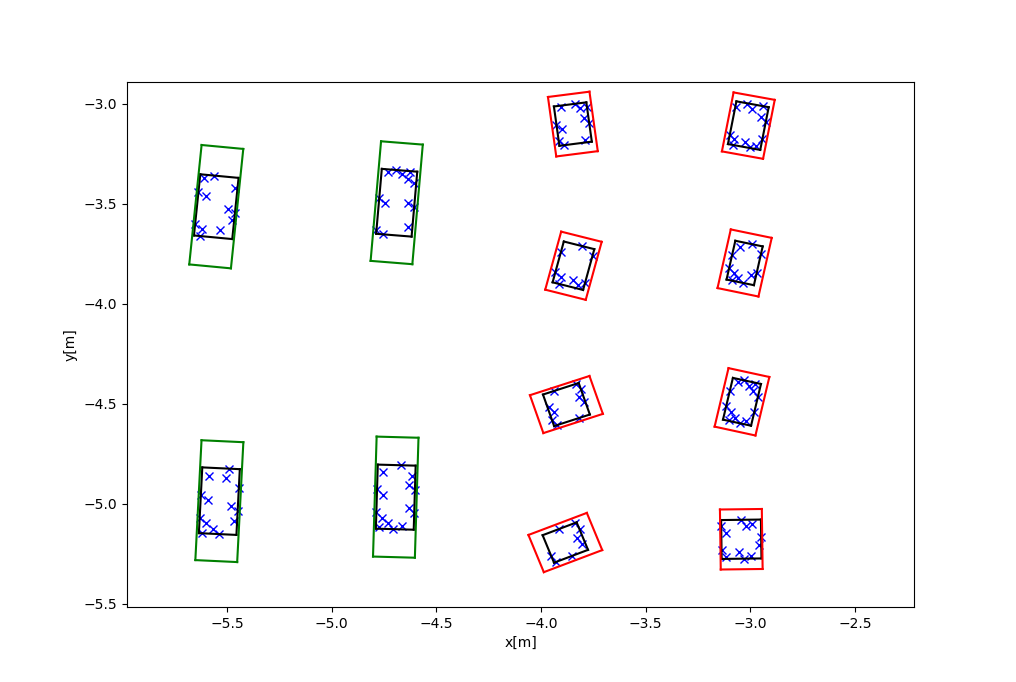}
  \caption{Detail of contours and brick detection results from \reffig{fig:small_detection} including the incorrect orientation of several bricks from scanning altitude.}
  \label{fig:small_detection_detail}
\end{figure}

\begin{figure}
  \centering
  \subfloat {
    \begin{tikzpicture}
      \node[anchor=south west,inner sep=0] (img) at (0,0) {\includegraphics[width=0.23\textwidth]{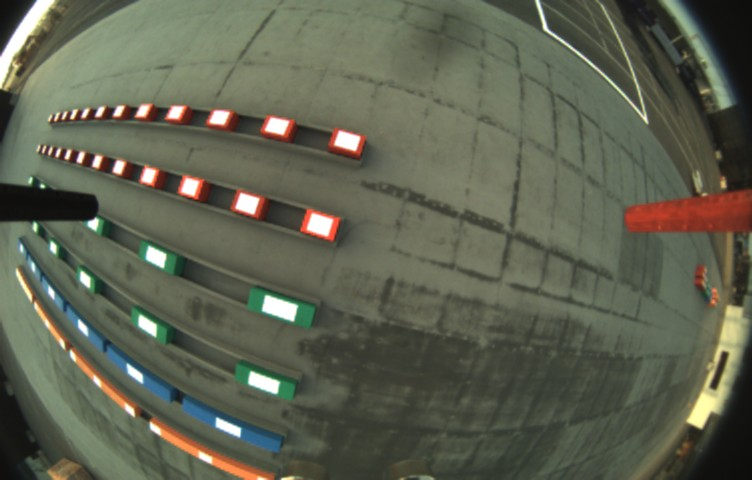}};
      \begin{scope}[x={(img.south east)},y={(img.north west)}]
        \node[imgletter,text=black] (label) at (img.south west) {(a)};
      \end{scope}
    \end{tikzpicture}
  } \subfloat {
    \begin{tikzpicture}
      \node[anchor=south west,inner sep=0] (img) at (0,0) {\includegraphics[width=0.23\textwidth]{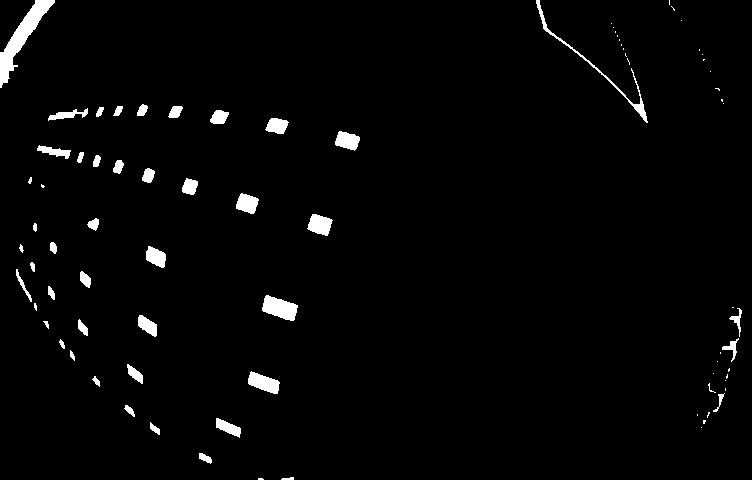}};
      \begin{scope}[x={(img.south east)},y={(img.north west)}]
        \node[imgletter,text=black] (label) at (img.south west) {(b)};
      \end{scope}
    \end{tikzpicture}
  } \\
  \subfloat {
    \begin{tikzpicture}
      \node[anchor=south west,inner sep=0] (img) at (0,0) {\includegraphics[width=0.47\textwidth]{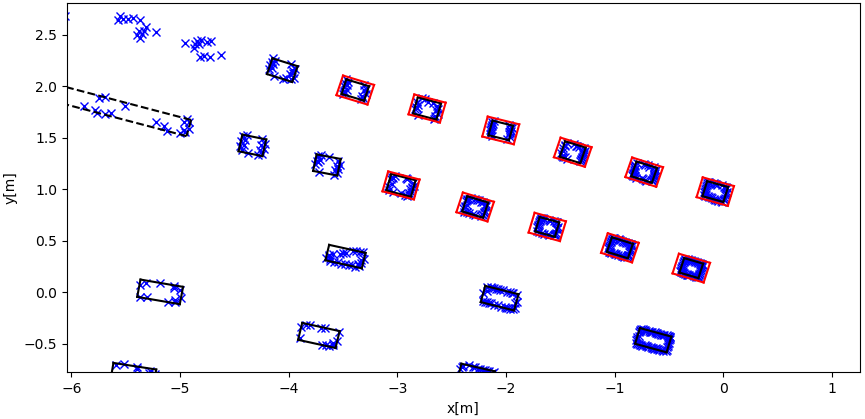}};
      \begin{scope}[x={(img.south east)},y={(img.north west)}]
        \node[imgletter,text=black] (label) at (img.south west) {(c)};
      \end{scope}
    \end{tikzpicture}
  }
  \caption{An original image from the color camera, white color segmentation result, and the contours and brick detection results.}
  \label{fig:middle_detection}
\end{figure}

\begin{figure*}
  \centering
  \renewcommand{\tabcolsep}{2pt}
  \noindent\begin{tabular}{lcr}
    \includegraphics[width=0.25\textwidth]{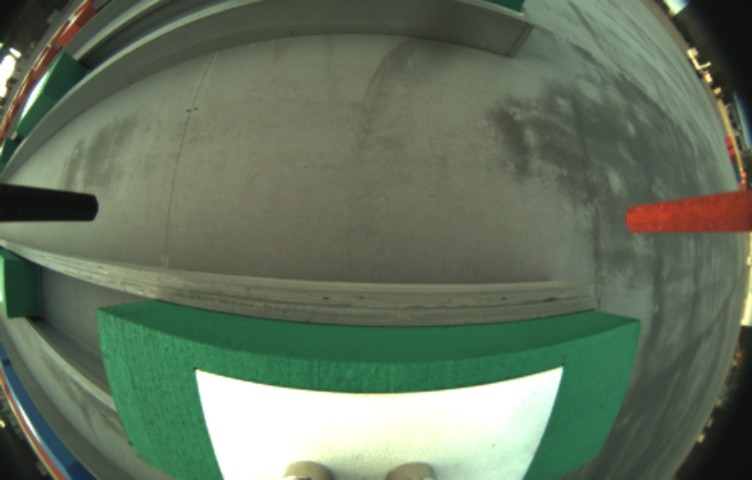}
    &
    \includegraphics[width=0.25\textwidth]{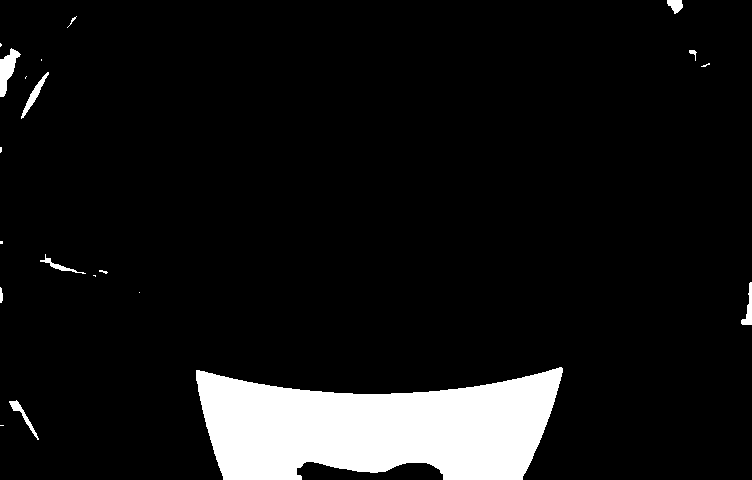}
    &
    \includegraphics[width=0.48\textwidth,trim=2.2cm 1.0cm 3.1cm 1.2cm, clip]{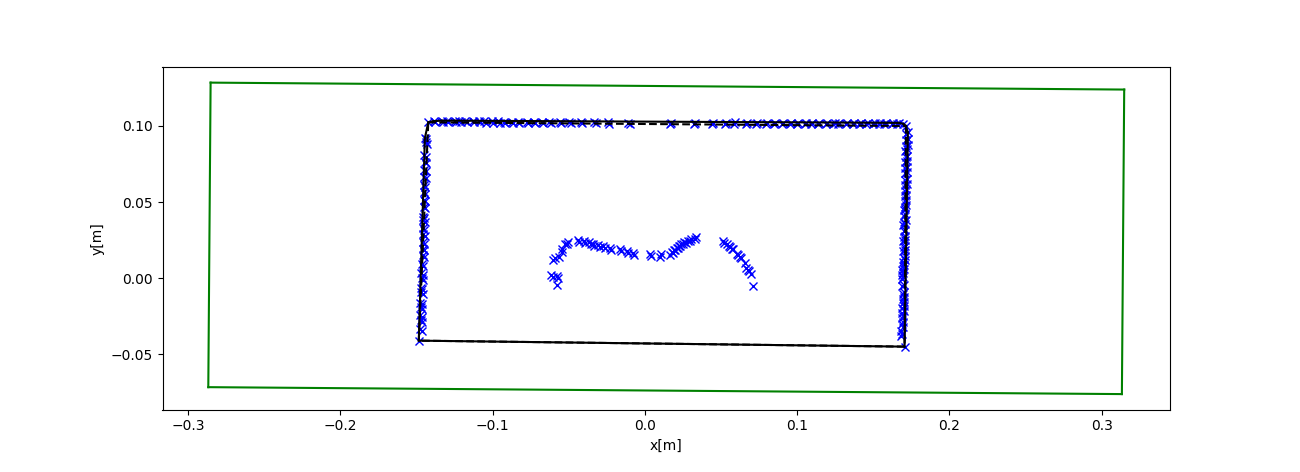}\\
  \end{tabular}
  \caption{An original image from the color camera, white color segmentation result, and the contours and brick detection results.}
  \label{fig:touch_detection}
\end{figure*}

The results of the brick detection algorithm from the scanning altitude \SI{4.5}{\meter} are depicted in \reffig{fig:small_detection}.
Details of transformed points to a global coordinate system with illustrated brick detection is shown in \reffig{fig:small_detection_detail}.
The better and more accurate detection from a lower altitude is depicted in \reffig{fig:middle_detection}.
In results from brick grasping (see \reffig{fig:touch_detection}), it can be noticed that only part of the white plate is visible and the correct brick position has to be calculated from the border shape.


\subsection{Brick and wall filtering}
\label{sec:brick_wall_filtering}

Each detected brick and wall segment is filtered and placed in a map.
The map consists of a bank of \acp{LKF} that maintains a smooth hypotheses of each object and provides stable references during grasping and placing attempts.
Upon detection, each object is first checked for a series of preconditions to be later fused in the map:
\begin{itemize}
  \item the object is excluded, if its coordinates are within \SI{5}{\meter} from other \ac{UAV} target,
  \item objects situated outside of the designated challenge area are excluded,
  \item bricks whose attempt to grasp was previously unsuccessful is excluded,
  \item wall segments outside the expected height range $\left[1.0, 2.3\right]$\,\si{\meter} are excluded.
\end{itemize}
The objects which pass the preconditions are matched with their nearest neighbour in the map.
In the case of a brick, a standard correction to the \ac{LKF} is formed, containing the brick's x, y, z world coordinates and heading.
The wall segments also contain their length, which is an important factor for matching the measurements to the map.
The wall segment detections are projected orthogonally to a candidate hypothesis to obtain a measure for evaluating the similarity of the segments.
When no match is found for the detected object, a new instance of \acp{LKF} is created and placed in the map.
Each hypothesis in the map maintains a counter for the number of corrections that were applied to the instance of the \ac{LKF}.

Post-processing of the map is applied periodically during flight to merge nearby hypotheses by combining their states in the ratio of the number of corrections in each hypothesis.
This is required due to the drift of the \ac{GPS} localization system which causes the objects to drift even in the time span of a single flight.
The post-processed detection map is later used to obtain a topological estimate of important sectors in the map (e.g. \ac{UAV} wall area, \ac{UGV} wall area, \ac{UAV} brick area, and \ac{UGV} brick area).


\subsection{Topological map creation}

The wall and brick detections are saved and filtered in the detection map during the entire scanning flight in order to create the topological map of the arena.
It is essential to precisely map positions of the wall channels and the individual brick type stacks (as shown in \reffig{fig:arena_layout_tikz}), to determine wall building plans for individual \acp{UAV}.
Map creation had to manage possible wrong detections, filter out the \ac{UGV} bricks present in the arena, and correctly decide the order of wall channels to follow for the prescribed wall pattern.
Figure~\ref{fig:mapping_detections} in the results section shows an example of the mapped wall channels and bricks.

The brick and wall detections received from the detection map are handled separately as they were placed independently in the arena.
Initially, all bricks with a low number of corrections (empirically set to 6 corrections) are filtered out and considered detection noise.
Next, a \ac{GMM}~\cite{Reynolds2009_GMM} for two clusters is estimated using the detections' $(x, y)$ position in order to separate \ac{UAV} and \ac{UGV} brick stacks.
Only the detections which are close (within $\SI{6}{\meter}$) to one of the two cluster means are kept and a \ac{PCA}~\cite{Jolliffe2011_PCA} with two components (due to the data being two-dimensional in $(x, y)$) is applied to both clusters.
The \ac{PCA} returns two variances for each cluster effectively proportional to width and height of the arbitrarily rotated \ac{UAV} and \ac{UGV} brick stacks.
The \ac{UAV} stack is then selected as the one with the larger width.
A median filter together with outlier removal (bricks farther than \SI{8}{\meter} from the median) is then iteratively used for the \ac{UAV} stack until the median converges, or for a limited number of iterations.
The line segments along the brick types (see \reffig{fig:arena_layout_tikz} and \reffig{fig:mapping_detections_uav}) are formed from the brick $(x, y)$ mean position of individual brick types and from the median heading of all remaining bricks.

The walls from the detection map are also first filtered out in case of no corrections of the particular wall.
The iterative median filter with outlier removal (of wall centers farther than \SI{10}{\meter} from the median), similar to the one for the brick detections, is then used to find the most perspective location of the `W' letter-shaped wall segments (see \reffig{fig:arena_layout_tikz}).
After the median filtering, the remaining wall segments are clustered together such that each detection is assigned to a cluster with horizontal distance within \SI{3}{\meter} and heading distance within \SI{0.5}{\radian}.
The average position and rotation of the clusters are then assumed to be the individual wall channels.
Line intersections of such wall channels are used to decide the order of wall channels in the `W' letter shape while only intersections within short distance ($\leq \SI{6}{\meter}$) from channel centers and with rather perpendicular mutual heading angle ($\geq \SI{0.7}{\radian}$) are further used.
A channel with only one intersection is then used as the first in the string of the `W' shape.
Alternatively, an intersection with the highest sum of distances from wall centers is iteratively removed until a wall with one intersection exists.
Afterwards, the other wall channels are added to the selected first channel according to the remaining intersections, and if more options exist, by selecting the intersection with shortest distances from wall centers.
Finally, the channel centers, lengths, and headings are determined (see \reffig{fig:mapping_detections_uav}) based on the intersections of the formed `W' shape wall structure.




\section{Brick grasping\label{sec:grasping}}


Brick grasping is the second primary capability for the wall building task.
The proposed approach for grasping uses a fusion of color and depth camera sensors for brick detection and visual servoing for precise control during grasping attempts.
The grasping state machine is used to govern various stages of grasping with \ac{UAV} mass and attitude being checked during grasping to abort in case of, e.g., a grasp far from the brick center of mass.

\subsection{Brick detection and localization}

\begin{figure*}
  \centering
  \subfloat {
    \begin{tikzpicture}
      \node[anchor=south west,inner sep=0] (img) at (0,0) {\includegraphics[width=0.223\textwidth]{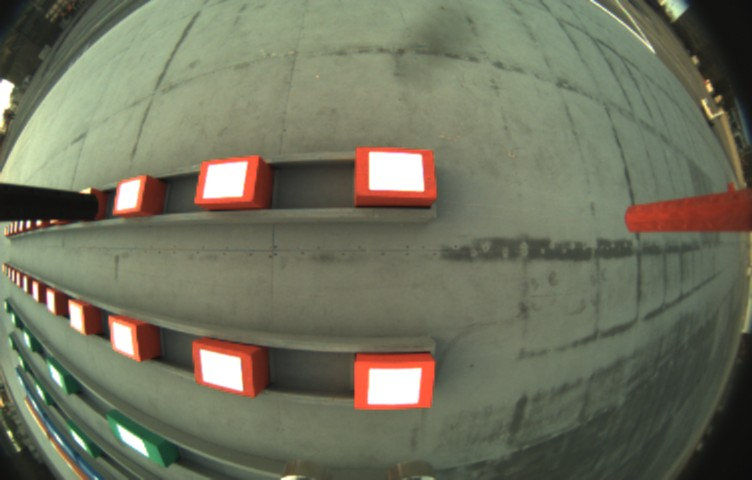}};
      \begin{scope}[x={(img.south east)},y={(img.north west)}]
        \fill[draw=black, draw opacity=0.5, fill opacity=0] (0,0) rectangle (1, 1);
        \node[imgletter,text=black] (label) at (img.south west) {(a)};
      \end{scope}
    \end{tikzpicture}}
  \hfill%
  \subfloat {
    \begin{tikzpicture}
      \node[anchor=south west,inner sep=0] (img) at (0,0) {\includegraphics[width=0.25\textwidth]{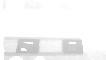}};
      \begin{scope}[x={(img.south east)},y={(img.north west)}]
        \fill[draw=black, draw opacity=0.5, fill opacity=0] (0,0) rectangle (1, 1);
        \node[imgletter,text=black] (label) at (img.south west) {(b)};
      \end{scope}
    \end{tikzpicture}}
  \hfill%
  \subfloat {
    \begin{tikzpicture}
      \node[anchor=south west,inner sep=0] (img) at (0,0) {\includegraphics[width=0.25\textwidth]{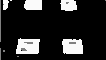}};
      \begin{scope}[x={(img.south east)},y={(img.north west)}]
        \fill[draw=black, draw opacity=0.5, fill opacity=0] (0,0) rectangle (1, 1);
        \node[imgletter,text=black] (label) at (img.south west) {(c)};
      \end{scope}
    \end{tikzpicture}}
  \hfill%
  \subfloat {
    \begin{tikzpicture}
      \node[anchor=south west,inner sep=0] (img) at (0,0) {\includegraphics[width=0.235\textwidth]{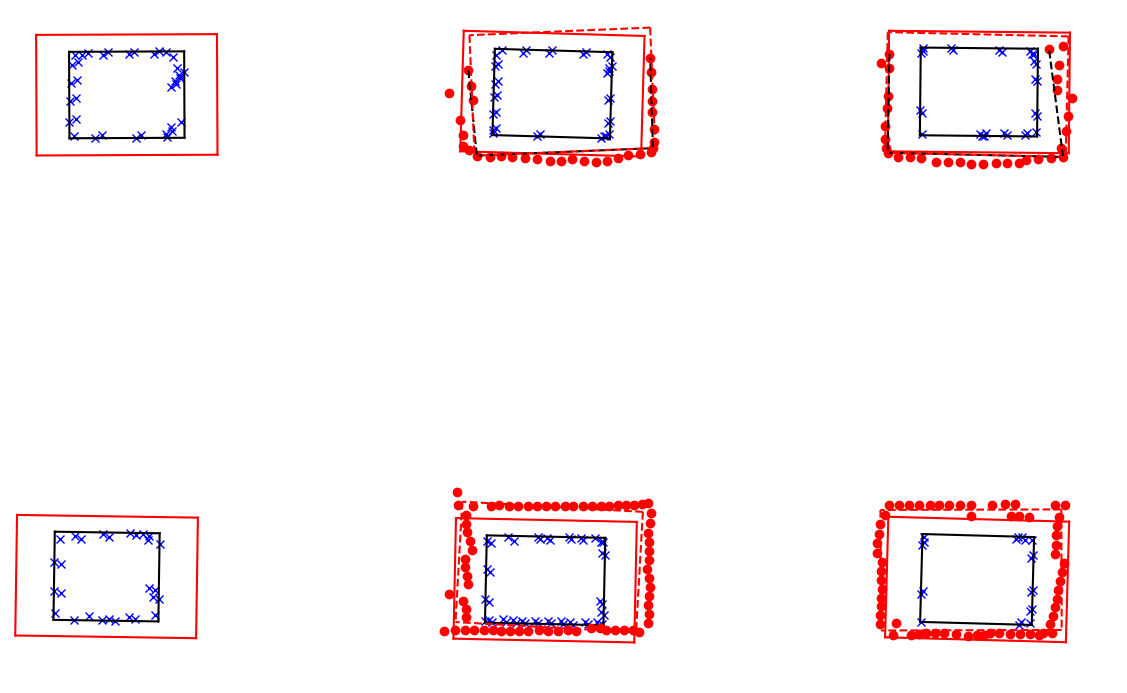}};
      \begin{scope}[x={(img.south east)},y={(img.north west)}]
        \fill[draw=black, draw opacity=0.5, fill opacity=0] (0,0) rectangle (1, 1);
        \node[imgletter,text=black] (label) at (img.south west) {(d)};
      \end{scope}
    \end{tikzpicture}}
  \caption{Fusion of data for brick detection using color-based (a) and depth-based (b) detection.
  Thresholded depth data are shown in (c), while the fused data are in (d) showing detections from the color camera as solid line, and from depth camera as dashed line.}
  \label{fig:data_fusion}
\end{figure*}
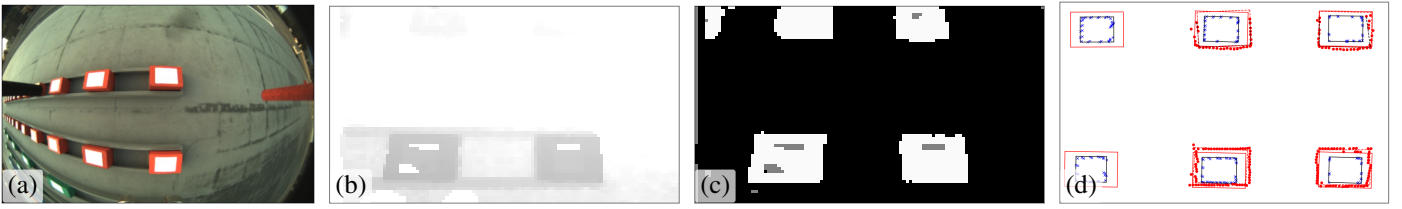

Brick grasping is based on fast and robust brick detection.
A fusion of detections from the color camera and from the depth RealSense sensor is used to improve the robustness.
The method of brick detection from the color camera is the same as is used during the scanning (described in \refsec{sec:color_detection}).
Brick detection from the RealSense sensor is similar to the wall detection from the RealSense as described in \refsec{sec:wall_detection}.
The altitude of measured points is computed for each pixel of the depth reduced image.
Similarly to the wall detection, the brick detection uses altitude thresholding with the threshold value of \SI{0.15}{\meter}.
Figure~\ref{fig:data_fusion} shows the result of the thresholding (\ref{fig:data_fusion}c), boundary pixels transformed to \ac{UAV} coordinate system, and their line approximations (\ref{fig:data_fusion}d - points and dashed lines).
The final brick detection is the same as for the color brick detection from \refsec{sec:color_detection}.

The final data fusion uses a weighted average of the color and depth detections where the weights depend on the quality of the detections.
The best quality detection weight $1$ is in case of the whole brick contour being visible by the sensor.
Supposing three perpendiculars lines represent the contour then the weight $0.5$ is used.
In case of only two perpendicular lines, the weight $0.25$ is used.
Example of data fusion is depicted in \reffig{fig:data_fusion}.


\subsection{UAV visual servoing}

\begin{figure}
  \centering
  \includegraphics[width=0.4\textwidth]{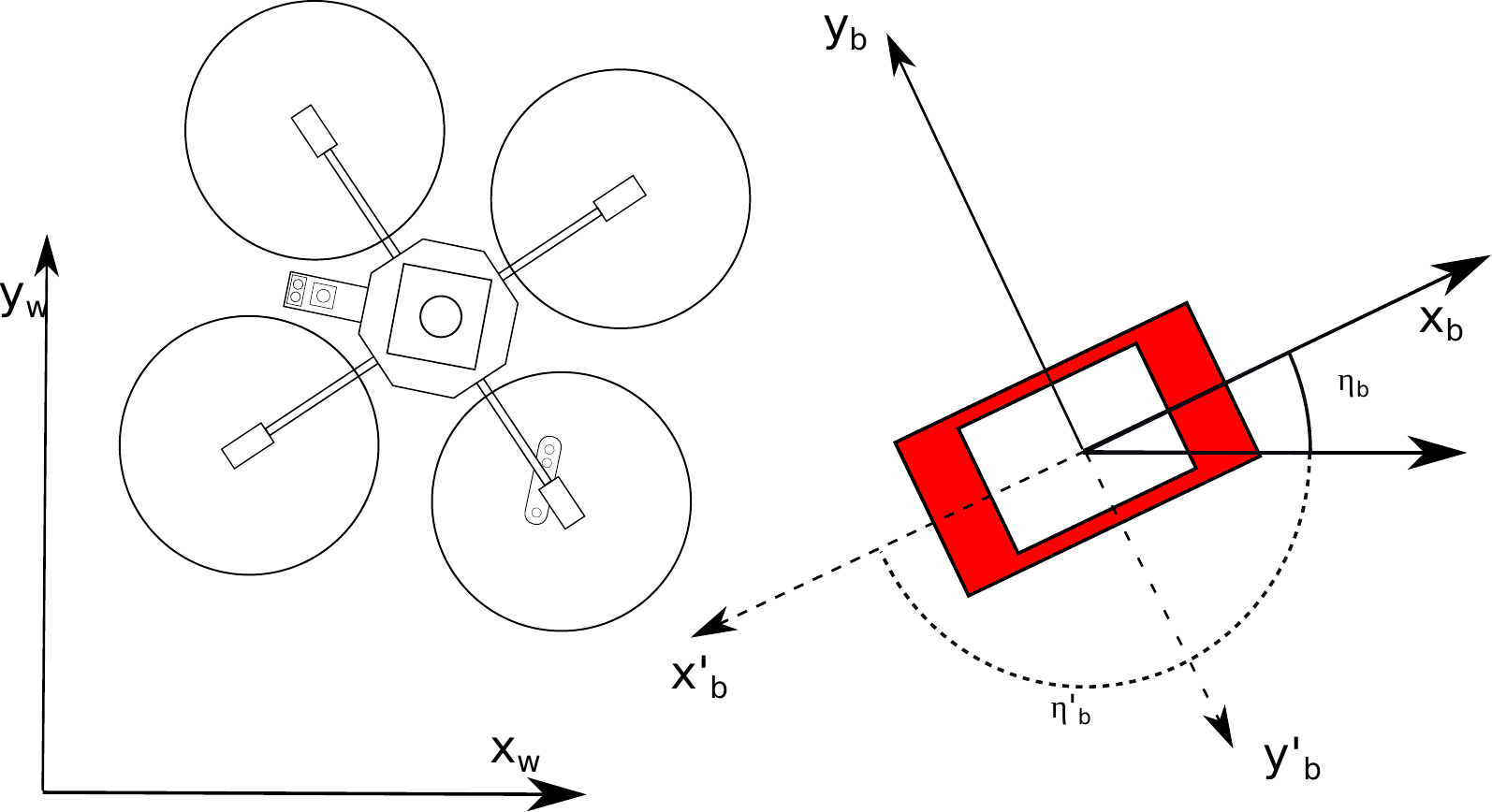}
  \caption{Brick coordinate with two possible axis placements.}
  \label{fig:brick_coord}
\end{figure}

For precise grasping, we employ a visual servoing technique where the position of the \ac{UAV} is computed in the coordinate system of the brick that is being grasped.
The main challenge of such visual servoing is the ambiguity of the brick's coordinate system (see \reffig{fig:brick_coord} with two possible axis placements).
The $z$ axis of the brick frame is parallel to the world frame $z$ axis, and therefore the $z$ coordinate is only a shift of the $z$ coordinate in world frame by the brick's height.
The brick coordinate system is defined by brick position in world frame $\mathbf{b}^{\mathcal{W}}$ and brick orientation $\eta_b$ in $x,y$ the axis plane of world frame as seen in \reffig{fig:brick_coord}.
The computation of \ac{UAV} position within the brick coordinate system must remember the last brick orientation, and thus the choice of initial axis placement. The ambiguity of the brick coordinate system is caused by inaccuracy of brick orientation that can change as the \ac{UAV} moves closer to the brick.


The detection algorithm finds bricks from actual data in a temporarily created map for data fusion.
The last brick orientation is used to select a new brick orientation.
The selection is based on the angle difference between orientation of the last brick $\eta_{b}^{map}$ and the newly detected brick $\eta^{new}_{b}$.
The selection can be expressed as
\begin{equation}
  \eta_b = \begin{cases}
    \eta^{new}_{b}, & \text{if $\langle \eta^{map}_{b} - \eta^{new}_{b} \rangle < \langle \eta^{map}_{b} - \eta^{new}_{b} - \Pi \rangle$} ,\\
    \eta^{new}_{b} + \Pi, & \text{otherwise}.
  \end{cases}
\end{equation}
The equation uses angle difference $\langle a-b \rangle$ which is the absolute value difference between angle $a$ and angle $b$, with result in interval $<0, \Pi>$.

The position of the \ac{UAV} in world frame system is denoted $\mathbf{r}^{\mathcal{W}}$.
The \ac{UAV} position $\mathbf{r}^{\mathcal{O}}$ within the coordinate frame $\mathcal{O}$ of the brick is expressed as
\begin{equation}
  \mathbf{r}^{\mathcal{O}} = \left[\begin{matrix} \cos{\eta_b} & \minus\sin{\eta_b} & 0\\
    \sin {\eta_b} & \cos {\eta_b} & 0\\
    0 & 0 & 1
  \end{matrix}\right] \cdot \left[\begin{matrix} \mathbf{r}^{\mathcal{W}}_x- \mathbf{b}^{\mathcal{W}}_x\\
    \mathbf{r}^{\mathcal{W}}_y - \mathbf{b}^{\mathcal{W}}_y \\
    \mathbf{r}^{\mathcal{W}}_z - 0.2
  \end{matrix}\right],
  \label{eq:uav_pos_calculation}
\end{equation}
where 0.2 represents height of the brick that is used as a shift in the $z$ axis.


\subsection{UAV-brick interaction and control}

Interaction of a multirotor \ac{UAV} with the environment is a complex challenge.
Small objects, such as the ones being collected during the \ac{MBZIRC} 2017 challenge \cite{spurny2019cooperative}, posed little to no challenge for common \acp{UAV} to carry.
However, the much larger and heavier bricks impose torque on the \ac{UAV} if not grasped in line with the center of mass of the object.
Moreover, the grasping event poses a threat to the \ac{UAV} by possibly limiting the controllable \ac{DOFs} of the \ac{UAV} due to mechanical contact.

The first challenge of carrying a sizeable elongated object was solved by designing the underlying \ac{UAV} control architecture.
The control pipeline executes a real-time weight estimator that allows the \ac{UAV} to not only detect an increase of its weight if an object was grasped, but also to detect a decrease when the \ac{UAV} rests upon the brick during the grasping maneuver.
The estimated mass is used throughout the control pipeline to provide adequate feed-forward control terms and scale the control gains of the employed \emph{SE(3) geometric feedback controller} \cite{lee2010geometric}.
With such measures, our \acp{UAV} were able to repeatedly carry all the brick types while performing moderately aggressive maneuvers.
It is worth noting our team's \acp{UAV} were the smallest vehicles of all teams, which conducted the task autonomously with an approximate 3:1 ratio of \ac{UAV} mass to brick mass.

The task of grasping a brick requires automatic safety measures to abort the action when the \ac{UAV} becomes uncontrollable.
Such a situation often occurs if the \ac{UAV} transfers its weight unevenly through its landing gear to the ground during the last moments of the grasping maneuver.
This state needs to be detected automatically by measuring the attitude control error and applying acceleration upwards to mitigate the effect quickly.
On the other hand, a false positive grasping event can occur when the magnetic gripper fails to attach.
This situation is detected as a significant decrease in the estimated mass due to the transfer of the \ac{UAV} weight thought the gripper to the brick.
In both cases, the maneuver is aborted and repeated before a collision can occur.

\subsection{Brick grasping state machine\label{sec:grasping_state_machine}}


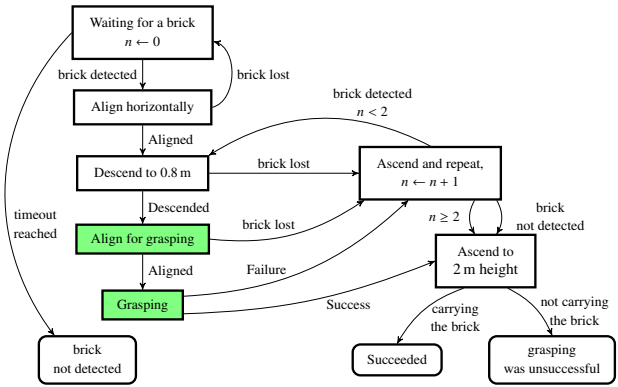
\begin{figure}
  \centering
  \begin{adjustbox}{width=0.47\textwidth}

\tikzset{radiation/.style={{decorate,decoration={expanding waves,angle=90,segment length=4pt}}}}


\tikzset{
  state/.style={
    rectangle,
    inner sep=5pt,
    draw=black, very thick,
    minimum height=2em,
    text centered,
  },
  final_state/.style={
    rectangle,
    rounded corners,
    draw=black, very thick,
    minimum height=2em,
    text centered,
  },
  initial_state/.style={
    rectangle,
    double=white,
    double distance=1pt,
    inner sep=2pt,
    draw=black, very thick,
    minimum height=2em,
    text centered,
  },
  point/.style={
    circle,
    inner sep=0pt,
    minimum size=3pt,
    fill=red
  },
  adder/.style={
    circle,
    inner sep=2pt,
    minimum size=0.3in,
    draw=black, very thick,
    text centered
  },
  state_gray/.style={
    rectangle,
    draw=black, very thick,
    fill=gray!40,
    minimum height=1.0em,
    text centered,
    inner sep=0,
  },
  state_white/.style={
    rectangle,
    draw=black, very thick,
    fill=white,
    minimum height=1.0em,
    text centered,
    text=black,
    inner sep=0,
  },
  state_green/.style={
    rectangle,
    draw=black, very thick,
    fill=green!50,
    minimum height=1.0em,
    text centered,
    text=black,
  },
  state_red/.style={
    rectangle,
    draw=black, very thick,
    fill=red!70,
    minimum height=1.0em,
    text centered,
    text=black,
    inner sep=0,
  },
  state_blue/.style={
    rectangle,
    draw=black, very thick,
    fill=blue!40,
    minimum height=1.0em,
    text centered,
    text=black,
    inner sep=0,
  },
  final_state/.style={
    rectangle,
    rounded corners,
    draw=black, very thick,
    minimum height=2em,
    text centered,
  },
  initial_state/.style={
    rectangle,
    double=white,
    double distance=1pt,
    inner sep=2pt,
    draw=black, very thick,
    minimum height=2em,
    text centered,
  },
  point/.style={
    circle,
    inner sep=0pt,
    minimum size=3pt,
    fill=red
  },
}


\begin{tikzpicture}[->,>=stealth', node distance=1.5cm,scale=0.8, every node/.style={scale=0.8}]

  \node[state, shift = {(0, -0.5)}] (idle) {
      \begin{tabular}{c}
        \small Waiting for a brick \\
        \small $n \leftarrow 0$
      \end{tabular}
    };


  \node[state, below of = idle, shift = {(0, -0.2)}] (align1) {
      \begin{tabular}{c}
        \small Align horizontally
      \end{tabular}
    }; 

  \node[state, below of = align1, shift = {(0, 0)}] (descend) {
      \begin{tabular}{c}
        \small Descend to \SI{0.8}{\meter}
      \end{tabular}
    }; 

  \node[state_green, below of = descend, shift = {(0, 0)}] (align2) {
      \begin{tabular}{c}
        \small Align for grasping
      \end{tabular}
    }; 

  \node[state_green, below of = align2, shift = {(0, 0)}] (grasping) {
      \begin{tabular}{c}
        \small Grasping
      \end{tabular}
    }; 

  \node[state, right of = descend, shift = {(5.0cm, 0)}] (repeat) {
      \begin{tabular}{c}
        \small Ascend and repeat,\\
        \small $n \leftarrow n+1$
      \end{tabular}
    }; 

  \node[state, below of = repeat, shift = {(1.25, -0.5)}] (takeoff) {
      \begin{tabular}{c}
       \small Ascend to\\
       \SI{2}{\meter} height
      \end{tabular}
    }; 

  \node[final_state, below of = takeoff, shift = {(-2.0, -0.75)}] (succes) {
      \begin{tabular}{c}
        \small Succeeded
      \end{tabular}
    }; 

    \node[final_state, left of = succes, shift = {(-5.5, 0)}] (timeout) {
      \begin{tabular}{c}
        \small brick \\ 
        \small not detected
      \end{tabular}
    }; 

    \node[final_state, below of = takeoff, shift = {(1.5, -0.75)}] (failure) {
      \begin{tabular}{c}
        \small grasping \\
        \small was unsuccessful
      \end{tabular}
    }; 


    \path[->] (idle.south) edge node[midway, left, shift = {(0.0, 0.0)}] {\small brick detected} (align1.north);
    \draw[->] (align1.south) -- (descend.north) node[midway, right] {\small Aligned};
    \draw[->] (descend.south) -- (align2.north) node[midway, right] {\small Descended};
    \draw[->] (align2.south) -- (grasping.north) node[midway, right] {\small Aligned};

    \path[->] (grasping.east)+(0, 0.2) edge [bend right=20] node[midway, left, shift = {(-0.1, 0.0)}] {\begin{tabular}{c}
        \small Failure
    \end{tabular}} ($(repeat.south)+(-0.5,0.0)$);

    \path[->] ($(grasping.east)+(0.0,-0.2)$) edge [bend right=10] node[midway, right, shift = {(0.2, -0.1)}] {\small Success} (takeoff.west);


    \path[->] ($(repeat.south)+(1.5,0.0)$) edge [bend left=30] node[midway, right, shift = {(0.0, 0.0)}] {
      \begin{tabular}{c}
        \small brick \\
        \small not detected
      \end{tabular}
    }($(takeoff.north)+(0.25,0.0)$);

    \path[->] ($(repeat.north)+(0,0)$) edge [bend right=30] node[midway, right, shift = {(0.0, 0.35)}] {\begin{tabular}{c}
        \small brick detected \\
        \small $n<2$
    \end{tabular}} ($(descend.north)+(1.5, 0.0)$);

    \path[->] ($(repeat.south)+(1,-0.0)$) edge [bend right=30] node[midway, left, shift = {(-0.1, 0.0)}] {\small $n\geq2$} ($(takeoff.north)+(-0.25,-0.0)$);

    \path[->] ($(idle.west)+(0,-0.0)$) edge [bend right=40] node[midway, right, shift = {(-0.15, -1.0)}] {
      \begin{tabular}{c}
        \small timeout \\
        \small reached
      \end{tabular}
  } ($(timeout.north)+(-0.75,-0.0)$);

    \path[->] ($(takeoff.south)+(0.5,-0.0)$) edge [bend left=20] node[midway, right, shift = {(-0.2, -0.1)}] { 
      \begin{tabular}{c}
        \small not carrying \\
        \small the brick
      \end{tabular}
    } ($(failure.north)+(0.0,-0.0)$);

    \path[->] ($(takeoff.south)+(-0.5,-0.0)$) edge [bend right=20] node[midway, right, shift = {(-0.2, -0.2)}] { 
      \begin{tabular}{c}
        \small carrying \\
        \small the brick
      \end{tabular}
    } ($(succes.north)+(0.0,-0.0)$);


    \path[->] (descend.east) edge [] node[midway, above, shift = {(0.0, 0.0)}] {\small brick lost} ($(repeat.west)+(0.0,0.0)$);
    \path[->] (align2.east) edge [bend right=20] node[midway, above, shift = {(-0.5, 0.0)}] {\small brick lost} ($(repeat.south)+(-1.5,0.0)$);
    \path[->] (align1.east)+(0, 0.0) edge [bend right=75] node[midway, right, shift = {(0.0, 0.0)}] {\small brick lost} ($(idle.east)+(-0.0,-0.2)$);



  \end{tikzpicture}
  \end{adjustbox}
  \caption{\ac{UAV} state machine for grasping a particular brick. This whole state machine corresponds to state \ref{state:grasping} within \reffig{fig:wall_sm}. The white state represents situations when the \ac{UAV} was localized by the \ac{GPS}, while in the green state, the states of the \ac{UAV} were estimated using only the visual detection of the brick for increased precision.}
  \label{fig:grasping_state_machine}
\end{figure}


The action of grasping a brick by the \ac{UAV} was governed by a state machine closely resembling the prize-winning variant from our last success during the \ac{MBZIRC} 2017 challenge \cite{spurny2019cooperative}.
Figure~\ref{fig:grasping_state_machine} depicts the states of the grasping state machine.
The \ac{UAV} is expected to be located at the vicinity of the desired brick (such that the brick is visible in the camera) when the first state is activated.
When a brick is detected, the \ac{UAV} first aligns itself horizontally with the closest brick of the desired color.
After the horizontal alignment distance is lower than \SI{0.2}{\meter}, the \ac{UAV} slowly descends to the height of \SI{0.8}{\meter} while actively maintaining the alignment.
If the alignment is broken, the \ac{UAV} ascends and attempts to realign with the brick to repeat the process.
The process is repeated a maximum of only twice after which the brick is abandoned and its location is temporarily banned to prevent deadlocks.
Conversely, when the \ac{UAV} successfully descends to the height of \SI{0.8}{\meter}, it switches its localization system to the direct brick visual servoing (the green states within \reffig{fig:grasping_state_machine}).
The \ac{UAV} then realigns itself again using only the detected brick as a source of the state estimate.
This second alignment starts with \SI{3}{\centi\meter} alignment criterion and relaxes the distance with time.
This ensures that the \ac{UAV} eventually attempts to grasp the brick even if the control accuracy is low.
The final grasping maneuver is also performed using the visual localization of the \ac{UAV} relative to the brick.
The process of adaptive and repeated switching from \ac{GNSS}-based control into visual servoing and back is the main contribution of this part for general object manipulation in demanding outdoor conditions.



\section{Brick placing\label{sec:placing}}


\subsection{Placing location detection}

\begin{figure}
  \centering
  \subfloat {\begin{tikzpicture}
    \node[anchor=south west,inner sep=0] (img) at (0,0) {\includegraphics[width=0.30\textwidth]{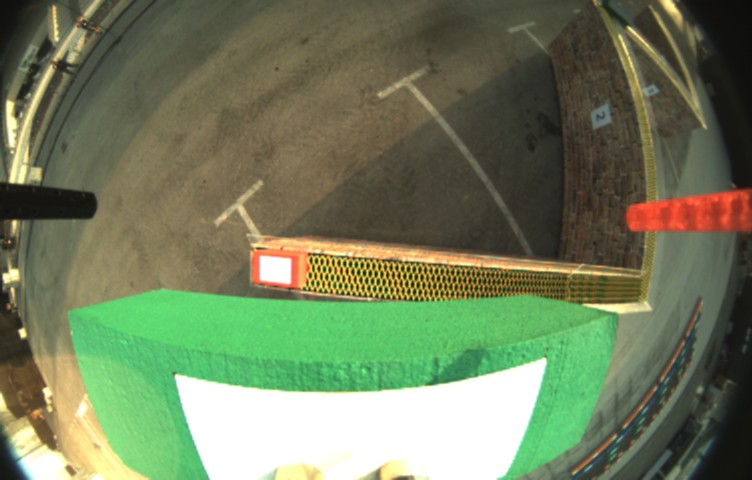}};
    \begin{scope}[x={(img.south east)},y={(img.north west)}]
      \fill[draw=black, draw opacity=0.5, fill opacity=0] (0,0) rectangle (1, 1);
      \node[imgletter,text=black] (label) at (img.south west) {(a)};
    \end{scope}
  \end{tikzpicture}} \\
  \vspace{-0.55em}
  \hfill
  \subfloat {\begin{tikzpicture}
    \node[anchor=south west,inner sep=0] (img) at (0,0) {\includegraphics[width=0.23\textwidth]{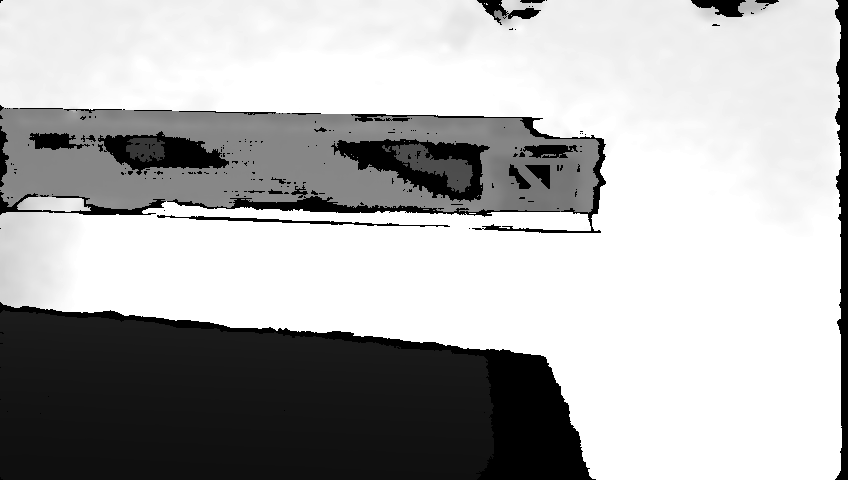}};
    \begin{scope}[x={(img.south east)},y={(img.north west)}]
      \fill[draw=black, draw opacity=0.5, fill opacity=0] (0,0) rectangle (1, 1);
      \node[imgletter,text=black] (label) at (img.south west) {(b)};
    \end{scope}
  \end{tikzpicture}}
  \hfill%
  \subfloat {\begin{tikzpicture}
    \node[anchor=south west,inner sep=0] (img) at (0,0) {\includegraphics[width=0.23\textwidth]{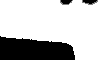}};
    \begin{scope}[x={(img.south east)},y={(img.north west)}]
      \fill[draw=black, draw opacity=0.5, fill opacity=0] (0,0) rectangle (1, 1);
      \node[imgletter,text=black] (label) at (img.south west) {(c)};
    \end{scope}
  \end{tikzpicture}}
  \hfill%
  \caption{Original data from the color camera (a) and original depth data (b).
  The mask (c) for the depth data is used to remove the carried brick and the UAV leg in the top part of the mask.
  }
  \label{fig:wall_color}
\end{figure}

Place detection uses a similar approach as wall detection (see \refsec{sec:wall_detection}).
However, there are three main challenges for detection of locations for where to place the next brick.
The top surface of the wall is tiled with a repetitive pattern (see \reffig{fig:wall_color}) that makes the stereo camera detection difficult.
This problem is solved mainly by filtering minimal distances by reduction of the depth image size.
Additionally, the second solution uses a morphological \emph{closing} operation after thresholding.
The second challenging aspect is that the view of onboard camera to the wall is partially blocked by the brick attached to the gripper.
This problem is solved by applying an automatically created mask after successfully grasping the brick.
An example of such a mask is depicted in \reffig{fig:wall_color}.


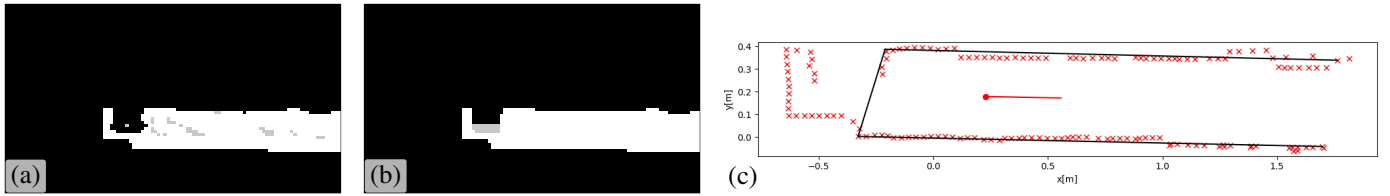
\begin{figure*}
  \centering
  \subfloat {
    \begin{tikzpicture}
      \node[anchor=south west,inner sep=0] (img) at (0,0) {\includegraphics[width=0.24\textwidth]{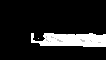}};

      \begin{scope}[x={(img.south east)},y={(img.north west)}]
        \fill[draw=black, draw opacity=0.5, fill opacity=0] (0,0) rectangle (1, 1);
        \node[imgletter,text=black] (label) at (img.south west) {(a)};
      \end{scope}

    \end{tikzpicture}}
  \hfill%
  \subfloat {
    \begin{tikzpicture}
      \node[anchor=south west,inner sep=0] (img) at (0,0) {\includegraphics[width=0.24\textwidth]{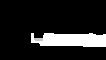}};
      \begin{scope}[x={(img.south east)},y={(img.north west)}]
        \fill[draw=black, draw opacity=0.5, fill opacity=0] (0,0) rectangle (1, 1);
        \node[imgletter,text=black] (label) at (img.south west) {(b)};
      \end{scope}
    \end{tikzpicture}}
  \hfill%
  \subfloat {
    \begin{tikzpicture}
      \node[anchor=south west,inner sep=0] (img) at (0,0) {\includegraphics[width=0.48\textwidth,trim=2.5cm 1.3cm 2.5cm 2.5cm, clip]{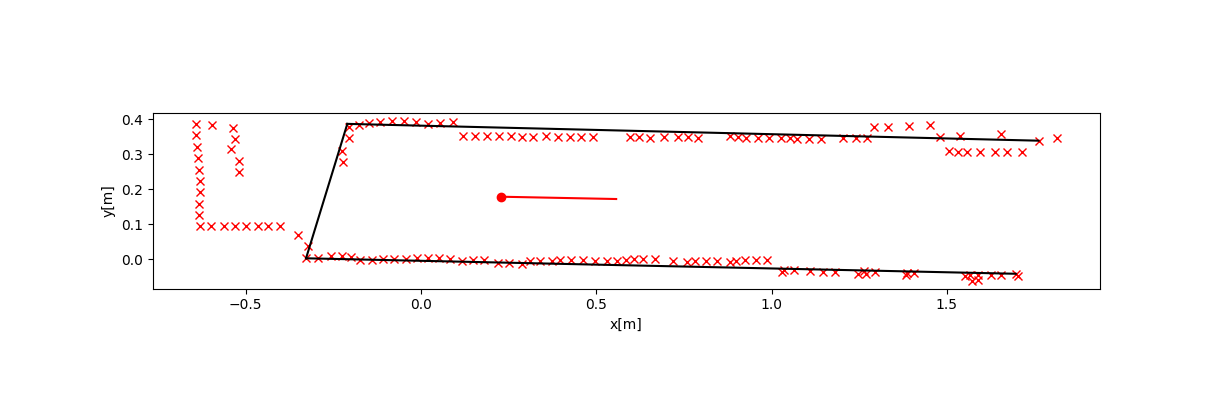}};
      \begin{scope}[x={(img.south east)},y={(img.north west)}]
        \node[imgletter,text=black] (label) at (img.south west) {(c)};
      \end{scope}
    \end{tikzpicture}}
  \caption{Thresholded depth data (a) and thresholded data after applying the morphological \emph{closing} operation (b).
  The result of place detection (c) where the red line demonstrates shift from the center of the wall to the leftmost position of the brick.}
  \label{fig:wall_place}
\end{figure*}

The last challenge is the transparent channel wings that were used to facilitate placement of bricks.
These transparent acrylic borders are rather randomly visible on depth measurements and may occasionally appear as a place free for placing a brick.
The proposed solution is to detect the free end of the wall by detecting the leftmost border point of the thresholded image (see \reffig{fig:wall_place}).

The method for detecting the placing spot on the wall assumes the wall segment is already aligned with the wider axis of the camera image.
The alignment is initially governed by the global planner which operates with necessary information obtained during the initial sweep.
The size of the brick being placed is known since the grasping procedure, and thus the method detects a place on the free wall at a correct distance from the leftmost border of the detected wall segment within the image.
As two layers of bricks can be built on the wall, the free area on the wall depends on the currently active layer.
In many cases, such free space contains a transparent acrylic border of the wall.
This border is removed from the detected wall by morphological \emph{erosion}.

The leftmost place on the wall is selected and if the left border of the wall is not visible, the \ac{UAV} moves to the leftmost part of a visible wall to find the correct edge.
The results of this algorithm are depicted in \reffig{fig:wall_place}.


\subsection{Placing state machine}
\label{sec:placing_state_machine}

\begin{figure}
  \centering

\tikzset{radiation/.style={{decorate,decoration={expanding waves,angle=90,segment length=4pt}}}}


\tikzset{
  state/.style={
    rectangle,
    inner sep=5pt,
    draw=black, very thick,
    minimum height=2em,
    text centered,
  },
  final_state/.style={
    rectangle,
    rounded corners,
    draw=black, very thick,
    minimum height=2em,
    text centered,
  },
  initial_state/.style={
    rectangle,
    double=white,
    double distance=1pt,
    inner sep=2pt,
    draw=black, very thick,
    minimum height=2em,
    text centered,
  },
  point/.style={
    circle,
    inner sep=0pt,
    minimum size=3pt,
    fill=red
  },
  adder/.style={
    circle,
    inner sep=2pt,
    minimum size=0.3in,
    draw=black, very thick,
    text centered
  },
  state_gray/.style={
    rectangle,
    draw=black, very thick,
    fill=gray!40,
    minimum height=1.0em,
    text centered,
    inner sep=0,
  },
  state_white/.style={
    rectangle,
    draw=black, very thick,
    fill=white,
    minimum height=1.0em,
    text centered,
    text=black,
    inner sep=0,
  },
  state_green/.style={
    rectangle,
    draw=black, very thick,
    fill=green!50,
    minimum height=1.0em,
    text centered,
    text=black,
  },
  state_red/.style={
    rectangle,
    draw=black, very thick,
    fill=red!70,
    minimum height=1.0em,
    text centered,
    text=black,
    inner sep=0,
  },
  state_blue/.style={
    rectangle,
    draw=black, very thick,
    fill=blue!40,
    minimum height=1.0em,
    text centered,
    text=black,
    inner sep=0,
  },
  final_state/.style={
    rectangle,
    rounded corners,
    draw=black, very thick,
    minimum height=2em,
    text centered,
  },
  initial_state/.style={
    rectangle,
    double=white,
    double distance=1pt,
    inner sep=2pt,
    draw=black, very thick,
    minimum height=2em,
    text centered,
  },
  point/.style={
    circle,
    inner sep=0pt,
    minimum size=3pt,
    fill=red
  },
}


\begin{tikzpicture}[->,>=stealth', node distance=1.5cm,scale=0.8, every node/.style={scale=0.8}]

  \node[state, shift = {(0, -0.5)}] (idle) {
      \begin{tabular}{c}
        \small Waiting for a wall \\
        \small $n \leftarrow 0$
      \end{tabular}
    };


  \node[state, below of = idle, shift = {(0, -0.4)}] (align1) {
      \begin{tabular}{c}
        \small Align horizontally \\
        \small in \SI{3.5}{\meter} height
      \end{tabular}
    }; 

  \node[final_state, below of = align1, shift = {(0.0, -0.15)}] (drop) {
      \begin{tabular}{c}
        \small Dropping brick
      \end{tabular}
    }; 

    \path[->] (idle.south) edge node[midway, left, shift = {(0.0, 0.0)}] {\small wall detected} (align1.north);
    \draw[->] (align1.south) -- (drop.north) node[midway, right] {\small Aligned};

    \path[->] (align1.east)+(0.0, 0.0) edge [bend left=40] node[midway, right, shift = {(-0.2, 0.0)}] {\begin{tabular}{c}
        \small wall lost
    \end{tabular}} ($(drop.east)+(0.0,0.0)$);


    \path[->] ($(idle.west)+(0,-0.0)$) edge [bend right=50] node[midway, right, shift = {(-1.65, 0.0)}] {
      \begin{tabular}{c}
        \small timeout \\
        \small reached
      \end{tabular}
  } ($(drop.west)+(0.00,-0.0)$);





  \end{tikzpicture}

  \caption{\ac{UAV} state machine for placing a brick on the wall. This state machine corresponds to state \ref{state:placing} within \reffig{fig:wall_sm}.}
  \label{fig:placing_state_machine}
\end{figure}
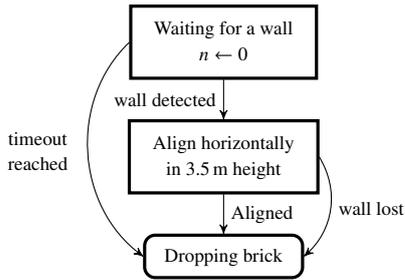

The action of placing a brick on the wall is governed by the state machine depicted in \reffig{fig:placing_state_machine}.
This lower-level state machine is responsible for guiding the \ac{UAV} above the spot designated for placement and controlling the descent to a desired height above the wall.
Since the outcome of placing a brick can rarely be influenced after releasing the brick from the magnetic gripper, we do not consider any actions in case of failure.
Moreover, numerous bricks are available in the grasping area, so grasping a misplaced brick or even repairing the wall is forfeit over continuing for a fresh brick.
Therefore, the placing state machine sequentially follows the actions of aligning horizontally with the wall and descending while aiming for the designated spot.
If anything fails, the held brick is dropped and the \ac{UAV} continues above the brick area to obtain a new brick.
This was chosen so as to not counteract any potential failure states, such as sudden misalignment (e.g., caused by localization drift).
A simpler yet capable approach was chosen due to the added complexity and less-deterministic execution of a more failure-proof solution.

\section{Experimental results}
\label{sec:results}

This section describes the results achieved during the \ac{MBZIRC} Challenge 2 competition trials using the proposed system.
The herein system for wall building by \acp{UAV}, developed as detailed above by the CTU-UPenn-NYU team, was able to win Challenge 2 by placing the far most number of bricks, mostly with the \acp{UAV}.
However, the \ac{UGV} deployed in the challenge also contributed by autonomously placing a brick during the second competition trial as described in~\cite{stibinger2020mobile}.
The CTU-UPenn-NYU team won by scoring 8.24 points, while the second Nimbro Team (University of Bonn) scored 1.33 points, and the third (Technical University of Denmark) 0.89 points.
Figure~\ref{fig:ch2_stand} depicts the team at the winner stand.

\begin{figure*}
  \centering
  \subfloat {
    \includegraphics[width=0.32\textwidth]{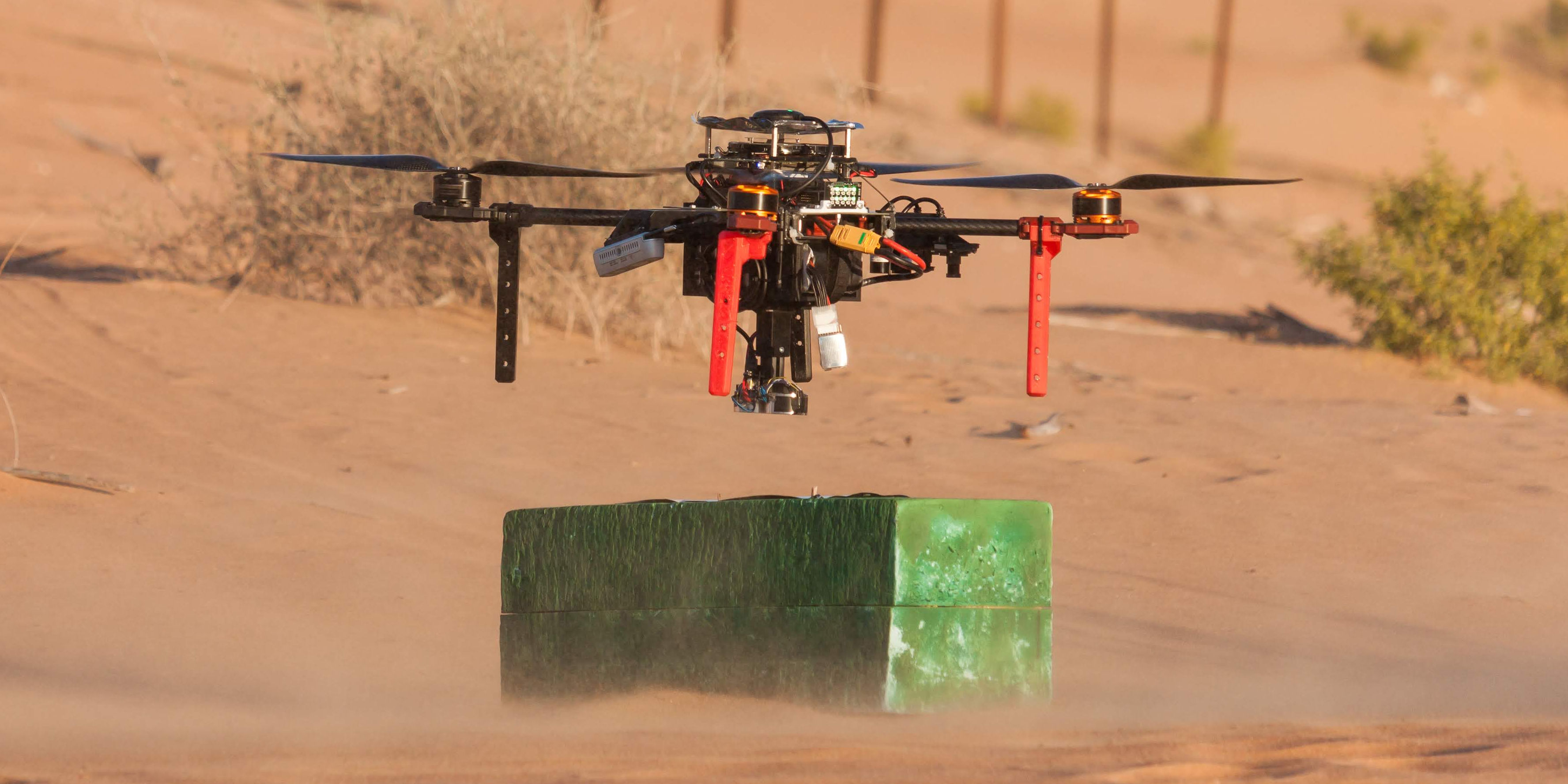}
  }
  \subfloat {
    \includegraphics[width=0.32\textwidth]{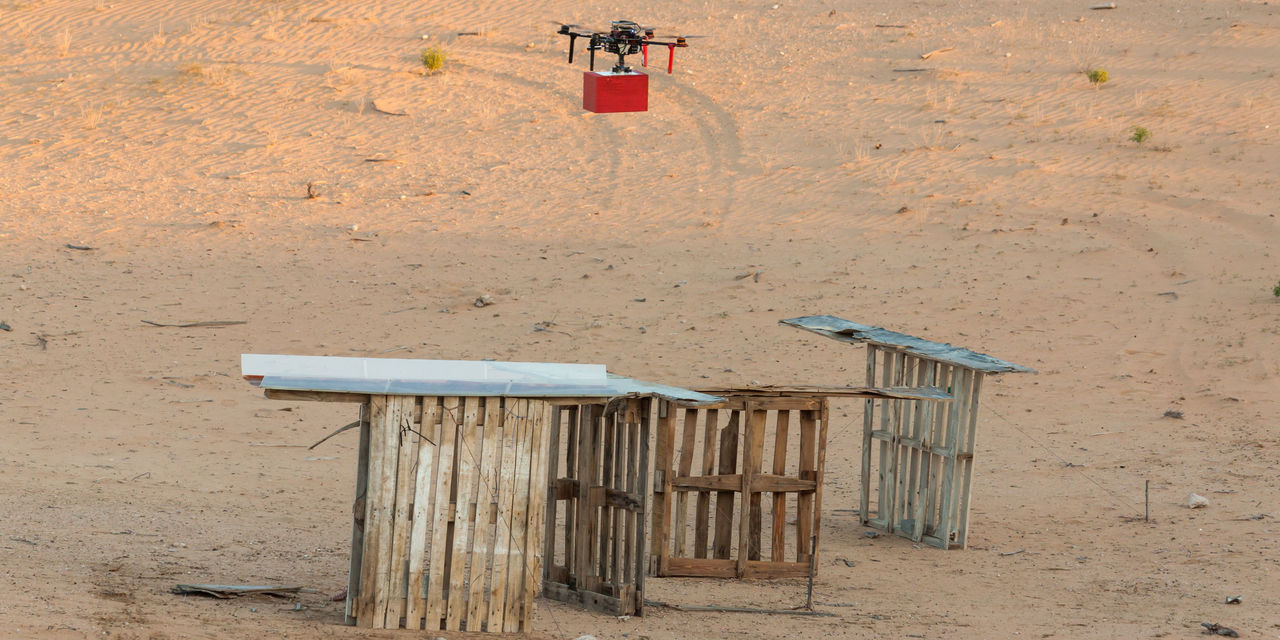}
  }
  \subfloat {
    \includegraphics[width=0.32\textwidth]{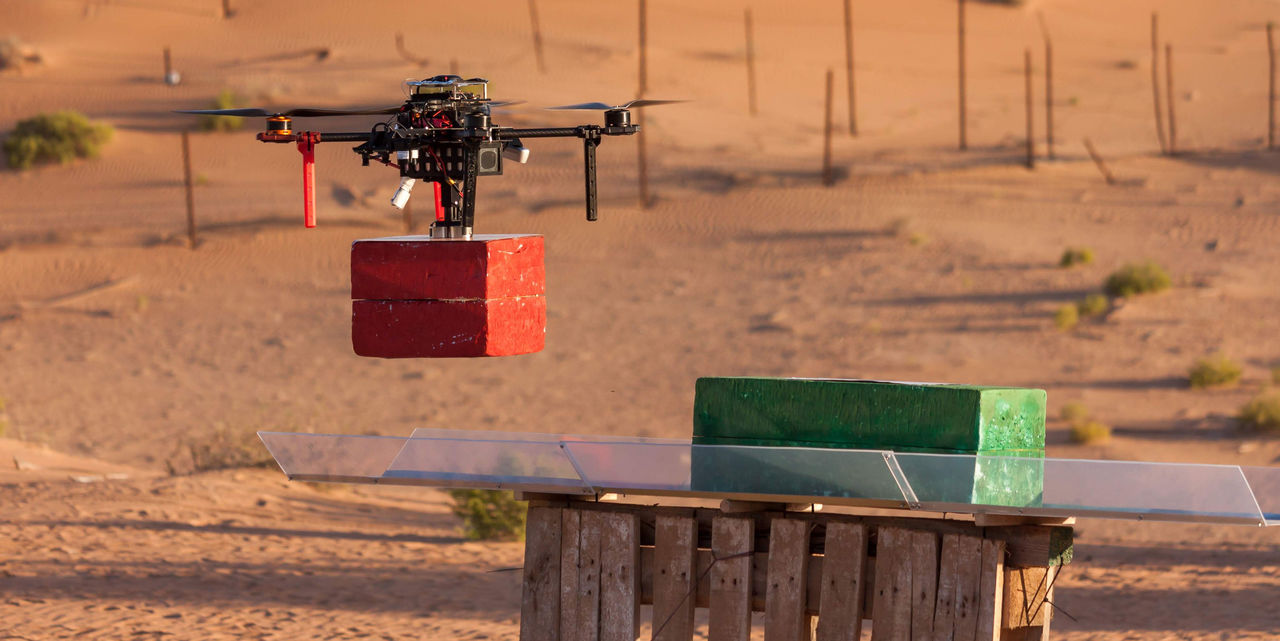}
  }
  \caption{Photos from preparation in the Abu Dhabi desert. Videos are available at \url{http://mrs.felk.cvut.cz/mbzirc-2020-uav-wall}.}
  \label{fig:desert_photos}
\end{figure*}

Prior to the competition, the team dedicated over a month for preparation and experimental evaluation in a desert near Abu Dhabi.
Real world experiments could not be conducted in the Czech Republic in the final months of preparations due to the winter weather conditions.
Therefore, the team decided to conduct final preparation near the competition venue, despite the unforgiving high temperatures, sand, and wind conditions of the coastal United Arab Emirates.
This preparation phase proved to be crucial in securing first place in the competition just as it was in 2017.
Figure~\ref{fig:desert_photos} depicts photos from the desert experiments. Videos from the experiments are available at \url{http://mrs.felk.cvut.cz/mbzirc-2020-uav-wall}.

Table~\ref{tab:results_bricks} shows the overall performance of the system during the two competition trials, each lasting 25 minutes.
It shows the types of bricks, denoted as `R' for RED and `G' for GREEN, in the order they were grasped during the individual trials.
The grasping of BLUE bricks was not attempted during the competition trials, although the \acp{UAV} were capable of carrying them, mainly due a significantly higher detachment probability of a grasped brick caused by the pendulum effect of the longer and heavier BLUE brick.
The grasping of ORANGE bricks was also not attempted for similar reasons and difficulty of cooperative carrying.
The competition did not require delivering any ORANGE bricks to qualify for obtaining points.
The grasped bricks are further displayed per individual restarts within the trials, where the restart had the possibility of keeping the already placed bricks and running the system again with all robots in their initial positions.






\begin{table}
  \centering
  \caption{Grasped bricks during the two competition trials of Challenge 2.}
  \label{tab:results_bricks}
  \setlength{\tabcolsep}{9pt}
  \renewcommand{\arraystretch}{1.0}
  \begin{tabular}{ccll}
    \noalign{\hrule height 1.1pt}\noalign{\smallskip}
    \bf{Trial} & \bf{Restart} & \multicolumn{1}{c}{\bf{UAV1}} & \multicolumn{1}{c}{\bf{UAV3}} \\ \hline \\[-1em]
    \multirow{4}{*}{one} & 1 &  R$^b$, R, G, G$^b$ & G \\
    & 2 & R, R, G$^b$         & R, G$^b$  \\
    & 3 & R$^b$, R$^b$        & R, R\\
    & 4 & G                   & R$^s$, R$^b$\\
    \hline
    \multirow{1}{*}{two} & 1 &  G, G, R$^b$, R$^b$, R & -- \\
    \noalign{\hrule height 1.1pt}\noalign{\smallskip}
  \end{tabular}
\end{table}


Table~\ref{tab:results_bricks} shows all grasped bricks, although not all were successfully placed on a wall.
The bricks denoted with `$b$' index did not have successful placement, which in most cases was due to the brick bouncing off the wall after release.
The single case of a placed RED brick during the fourth restart of the first trial, denoted with `$^s$' index, is a placement into the second layer of a particular channel which was also not achieved by any other team.

Notice that only UAV1 and UAV3 were used for wall building during the competition trials.
The strategy of having UAV2 wait for a collision or battery depletion of another UAV was not required during the competition.
However, this reliable multi-\ac{UAV} strategy was successfully tested during the pre-competition trials.


Ten bricks were successfully placed on the wall during the first trial consisting of seven RED and three GREEN bricks.
Seven bricks bounced off the wall during the same trial.
The main focus of the CTU-UPenn-NYU team during the second trial was to autonomously place at least one brick using the \ac{UGV} as was required for winning the challenge.
Therefore, many restarts were done to ensure this goal and only before the first restart were the \acp{UAV} used to grasp and place bricks as shown in \reftab{tab:results_bricks}.
In the rest of this section, we focus on the individual (i.e. scanning, grasping, placing) wall building subtasks and the overall performance achieved during the first trial, in which \acp{UAV} were used throughout the entire trial.

\subsection{Scanning for bricks and wall placement\label{sec:results_scanning}}

Scanning of the arena was the first subtask of the proposed UAV wall building approach performed in order to find the location of the brick stack and wall position designed for the \acp{UAV}.
The arena scanning was planned using a zig-zag path within the predefined arena space (defined by arena corners and safety area) as described in \refsec{sec:arena_scanning_path}.

\begin{figure}
  \centering
  \includegraphics[width=0.99\linewidth]{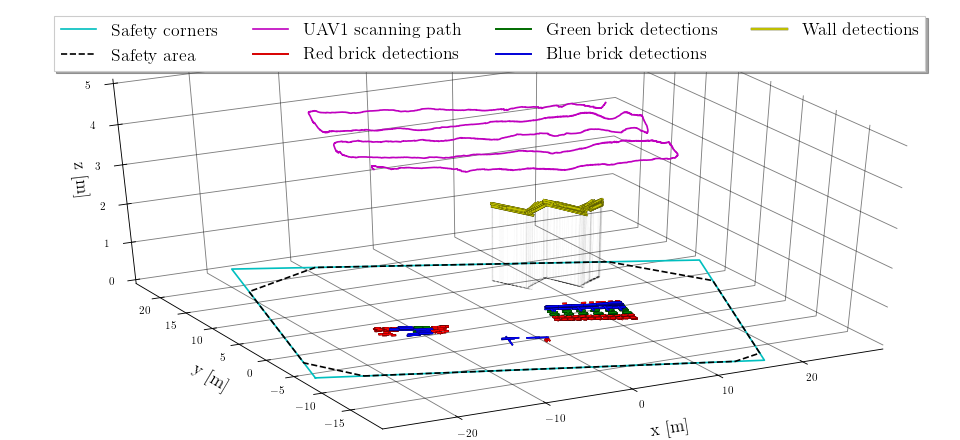}
  \caption{Path of UAV1 while scanning the arena and detections of the bricks and UAV wall channels.\label{fig:scanning}}
\end{figure}

Figure~\ref{fig:scanning} shows all brick detections (RED, GREEN and BLUE- ORANGE bricks were not considered) as well as the detections of the wall channels.
The zig-zag scanning path is shown within the arena boundaries as recorded by the onboard \ac{GPS}.
The employed brick detection using the onboard RGB camera was able to detect bricks in the range of $\approx$ \SI{10}{\meter} $\times$ \SI{5}{\meter} in x $\times$ y coordinates of the camera when flying at the scanning \SI{4.5}{\meter} height.
However, robust detections of bricks were obtained in range of $\approx$ \SI{5}{\meter} $\times$ \SI{3}{\meter} of x $\times$ y coordinates.
This required an overlapping camera field of view while following the zig-zag pattern path.


Figure \ref{fig:scanning} clearly shows the `W' letter shape of the four wall channels created by the wall detections.
The brick detections form two large clusters being the \ac{UGV} and \ac{UAV} brick stacks. The \ac{UAV} stack being wider was further used for the \ac{PCA} analysis.
Figure~\ref{fig:scanning} also features a set of false positive brick detections between the two brick stacks corresponding to the position of waiting UAV2 and the starting takeoff area marked as a white rectangle on the ground.

\begin{figure}[!htb]
  \centering
  \begin{tikzpicture}
    \node[anchor=south west,inner sep=0] (a) at (0,0) {
      \begin{tabular}{c}
        \subfloat{
          \includegraphics[height=12em]{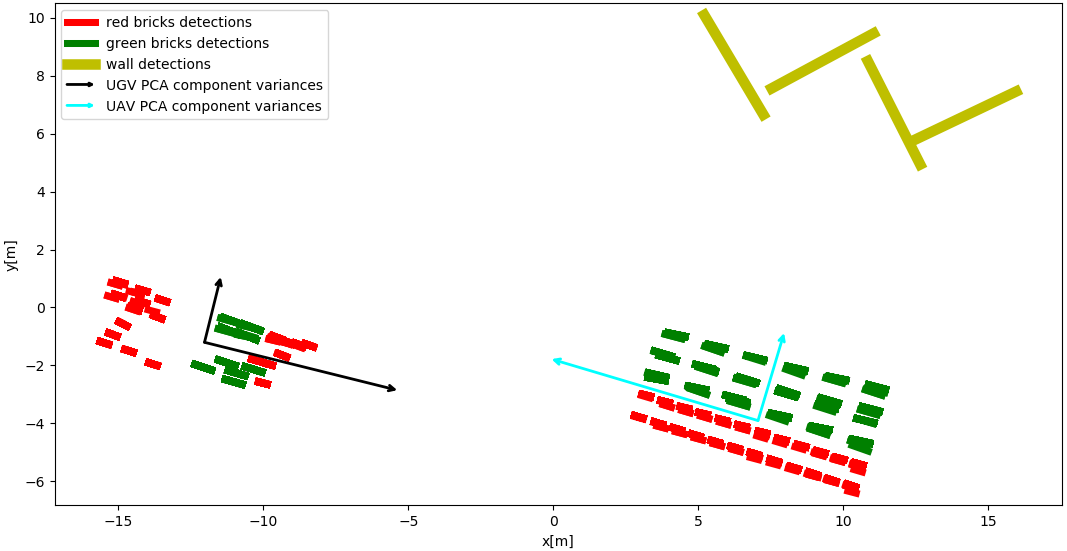}
          \label{fig:mapping_detections_all}
        }
        \vspace{-1.5em}
        \\
        \hfill\subfloat{
          \includegraphics[height=12em]{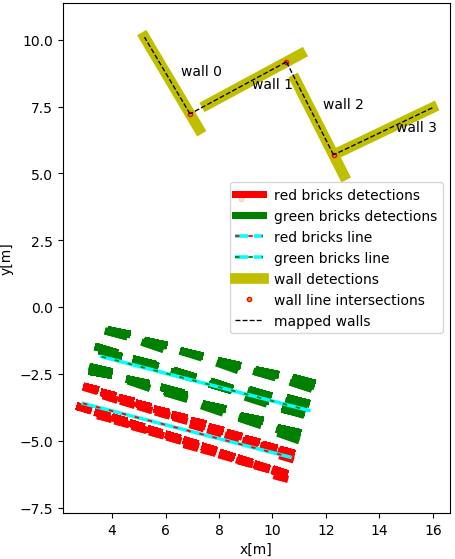}
          \label{fig:mapping_detections_uav}
        }
      \end{tabular}
    };


    \begin{scope}[x={(a.south east)},y={(a.north west)}]

      \draw (0.32,0.45) node [text=black] {
        \setlength{\tabcolsep}{3pt}
        \begin{tabular}{ll}
          \small (a) & \small All wall detections and brick\\
          & \small clusters with PCA analysis.
        \end{tabular}
      };
      \draw (0.43,0.13) node [text=black] {
        \setlength{\tabcolsep}{3pt}
        \begin{tabular}{rr}
          \small (b) & \small Mapped UAV\\
          & walls and\\
          & bricks.
        \end{tabular}
      };


    \end{scope}
  \end{tikzpicture}
  \vspace{-1em}
  \centering
  \caption{Topological map based on scanning of the arena.}
  \label{fig:mapping_detections}
\end{figure}

After completion of the scanning path, the wall and brick detections are processed to create a topological map of the arena.
The map contains positions of the individual wall channels ordered in a `W' letter chain as well as lines along the particular brick types that can be deterministically divided among the three \acp{UAV}.
Figure~\ref{fig:mapping_detections} shows the brick and wall detections already filtered out by the number of reoccurrences during the scanning.
Figure~\reffig{fig:mapping_detections_all} shows all the wall detections and the two bricks \ac{GMM} clusters with corresponding \ac{PCA} component variances proportional to width and height of the two brick stacks.
Figure~\reffig{fig:mapping_detections_uav} features only the \ac{UAV} stack together with the topological map consisting of `wall 0'--`wall 3' and the red and green brick lines.
The positions of the mapped walls and brick lines are then shared among \acp{UAV} and used to distribute the wall building task, as was detailed before in \refsec{sec:multirobot_aspects}.


By comparing \reffig{fig:scanning} and \reffig{fig:mapping_detections}, it can been seen that the initial reoccurrences-based filter removes the false positive detections present in the takeoff area.
However, the blue bricks present in the \reffig{fig:scanning} are also filtered out and are instead incorrectly labeled as green bricks in most detections.
The \ac{PCA} analysis indeed selects the correct brick stack for \acp{UAV} having the larger width, with the smaller variance component being larger for the \ac{UAV} stack.
Figure~\ref{fig:mapping_detections_uav} shows that the wall detections are correctly recognized as `wall 0'--`wall 3' based on the intersections of line approximations of the wall detections.
Moreover, the wall detection and mapping shows a great performance by creating the individual channels of almost the same size based only on the detections and analysis of the intersections.
Finally, by comparing the raw wall detections and the mapped walls, we can see a \emph{shadow effect} of the wall detection in \SI{1.7}{\meter} when projected to the ground plane due to the left-to-right scanning trajectory above the wall.


\begin{figure*}
  \centering

  \tikzset{
    partlabel/.style={
      rectangle,
      inner sep=2pt,
      rounded corners=.1em,
      draw=white,
      text=black,
      minimum height=1em,
      text centered,
      fill=white,
      fill opacity=.9,
      text opacity=1,
      anchor=north west,
    },
    partarrow/.style={
      draw=green, thick,
    },
  }

  \renewcommand{\tabcolsep}{2pt}
  \begin{tabular}{ccc}
    \multirow{2}{*}[7.0em]{\hspace{1.0em}\includegraphics[width=0.35\linewidth]{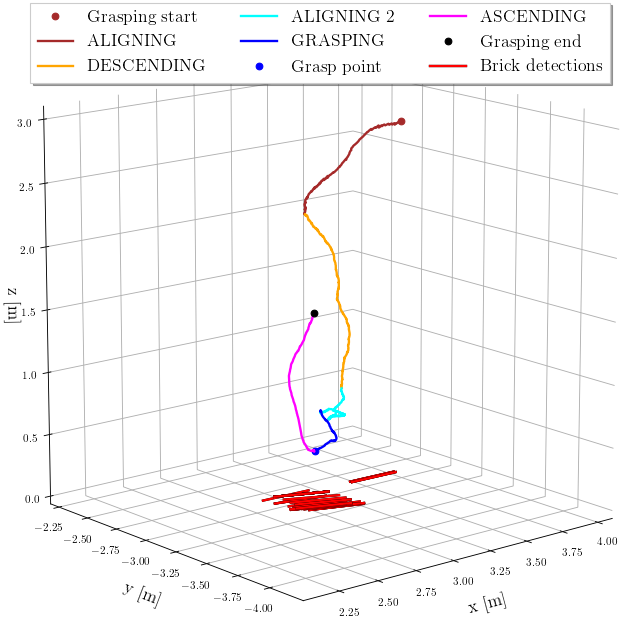}\hspace{1.0em}} & \begin{tikzpicture}
      \node[anchor=south west,inner sep=0] (img) at (0,0) {\includegraphics[width=0.25\linewidth]{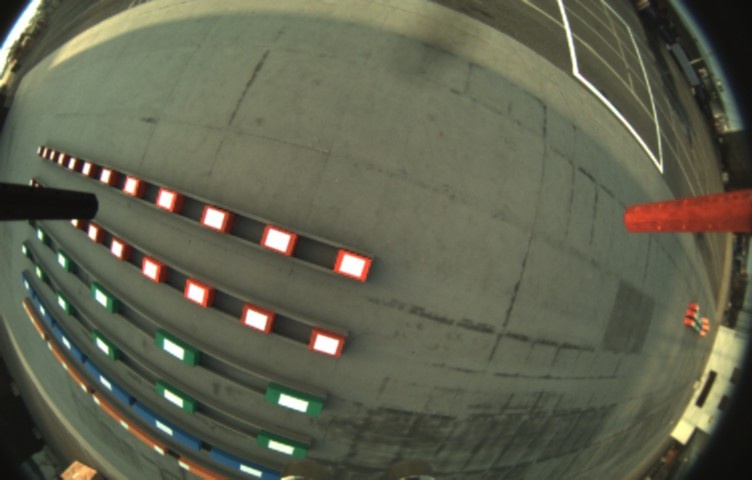}};
      \begin{scope}[x={(img.south east)},y={(img.north west)}]
        \node[partlabel,text=orange] (grasping_descending_label) at (img.north west) {(a) Descending start};
        \draw (img.north east) rectangle (img.south west);
      \end{scope}
    \end{tikzpicture} & \begin{tikzpicture}
      \node[anchor=south west,inner sep=0] (img) at (0,0) {\includegraphics[width=0.25\linewidth]{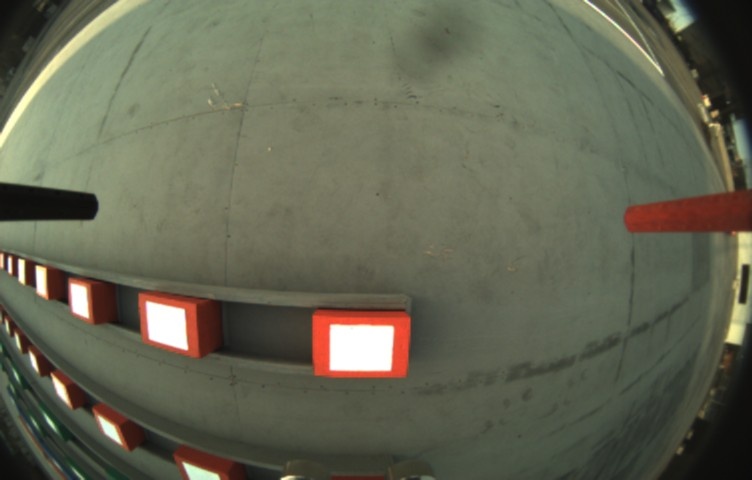}};
      \begin{scope}[x={(img.south east)},y={(img.north west)}]
        \node[partlabel,text=cyan] (grasping_align2_start_label) at (img.north west) {(b) Align 2 start};
        \draw (img.north east) rectangle (img.south west);
      \end{scope}
    \end{tikzpicture} \\
    & \begin{tikzpicture}
      \node[anchor=south west,inner sep=0] (img) at (0,0) {\includegraphics[width=0.25\linewidth]{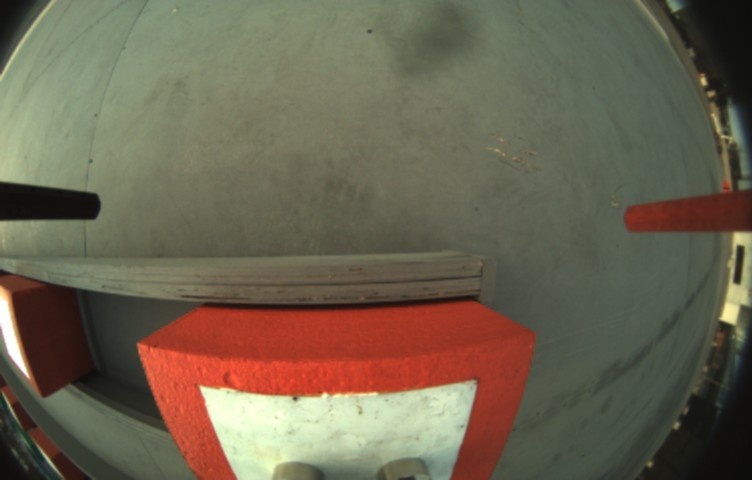}};
      \begin{scope}[x={(img.south east)},y={(img.north west)}]
        \node[partlabel,text=blue] (grasping_grasp_point_label) at (img.north west) {(c) Grasp point};
        \draw (img.north east) rectangle (img.south west);
      \end{scope}
    \end{tikzpicture} & \begin{tikzpicture}
      \node[anchor=south west,inner sep=0] (img) at (0,0) {\includegraphics[width=0.25\linewidth]{fig/grasping_ascending_start.jpg}};
      \begin{scope}[x={(img.south east)},y={(img.north west)}]
        \node[partlabel,text=black] (grasping_end_label) at (img.north west) {(d) Grasping end};
        \draw (img.north east) rectangle (img.south west);
      \end{scope}
    \end{tikzpicture}
  \end{tabular}
  \caption{Grasping of a red brick with color indicated stages of the grasping procedure.\label{fig:grasping}}
\end{figure*}


\subsection{Brick grasping}

The brick grasping is another key capability required for competing in the wall building task.
The grasping procedure~\ref{state:grasping} consists of the lower-level grasping state machine described in \refsec{sec:grasping_state_machine}.
Figure~\ref{fig:grasping} shows the evolution of the grasping states with respect to the \ac{UAV} position for the first successful grasp of the red brick with UAV1 during the second restart of first trial.


Figure~\ref{fig:grasping} also features the images taken by the mvBlueFOX camera that used for RGB detection of the bricks during various stages of the grasping state machine.
The brick detections are shown as measured during the grasping.
The rest of the detections have a mean position for x, y, heading being \SI{3.022}{\meter}, \SI{-3.215}{\meter},  \SI{2.825}{\radian}, respectively, and a corresponding standard deviation of \SI{0.029}{\meter}, \SI{0.077}{\meter} , \SI{0.077}{\radian}, respectively.
However, the absolute localization of the grasped brick is only relevant within the first two stages of the grasping manoeuvre (the first alignment and descent), where the \ac{UAV} is guided using these estimated \ac{GPS} coordinates of the brick.
The later stages use direct visual servoing to estimate the \ac{UAV} states using the brick detections which outperforms the accuracy of standard \ac{GPS} by an order of magnitude.
Our \acp{UAV} were able to target the magnetic plates on the bricks reliably within centimeter precision.

\subsection{Brick placement}

\begin{figure*}
  \centering

  \tikzset{
    partlabel/.style={
      rectangle,
      inner sep=2pt,
      rounded corners=.1em,
      draw=white,
      text=black,
      minimum height=1em,
      text centered,
      fill=white,
      fill opacity=.8,
      text opacity=1,
      anchor=north west,
    },
    partarrow/.style={
      draw=green, thick,
    },
  }

  \renewcommand{\tabcolsep}{2pt}
  \begin{tabular}{cccc}
    \multicolumn{2}{c}{\multirow{2}{*}[7.0em]{\hspace{1.6em}\includegraphics[width=0.33\linewidth]{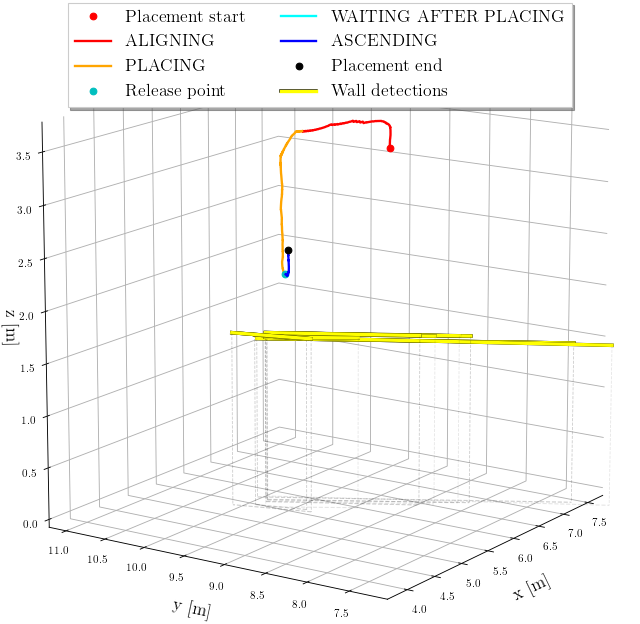}\hspace{1.6em}}} &
    \begin{tikzpicture}
      \node[anchor=south west,inner sep=0] (img) at (0,0) {\includegraphics[width=0.25\linewidth]{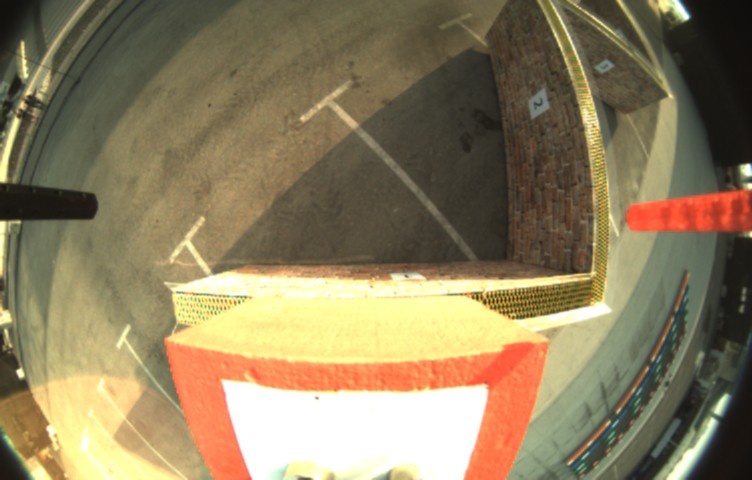}};
      \begin{scope}[x={(img.south east)},y={(img.north west)}]
        \node[partlabel,text=red] (placement_align_start_label) at (img.north west) {(a) Align start};
        \draw (img.north east) rectangle (img.south west);
      \end{scope}
    \end{tikzpicture} &
    \begin{tikzpicture}
      \node[anchor=south west,inner sep=0] (img) at (0,0) {\includegraphics[width=0.25\linewidth]{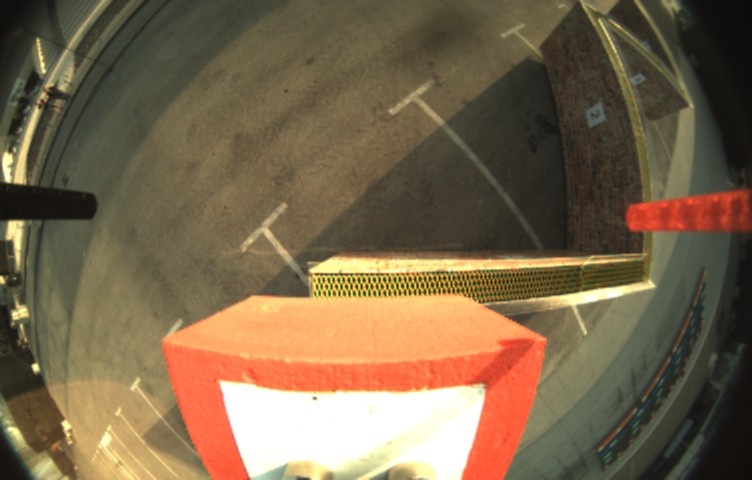}};
      \begin{scope}[x={(img.south east)},y={(img.north west)}]
        \node[partlabel,text=orange] (placement_placing_start_label) at (img.north west) {(b) Placing start};
        \draw (img.north east) rectangle (img.south west);
      \end{scope}
    \end{tikzpicture} \\
    \multicolumn{2}{c}{} &
    \begin{tikzpicture}
      \node[anchor=south west,inner sep=0] (img) at (0,0) {\includegraphics[width=0.25\linewidth]{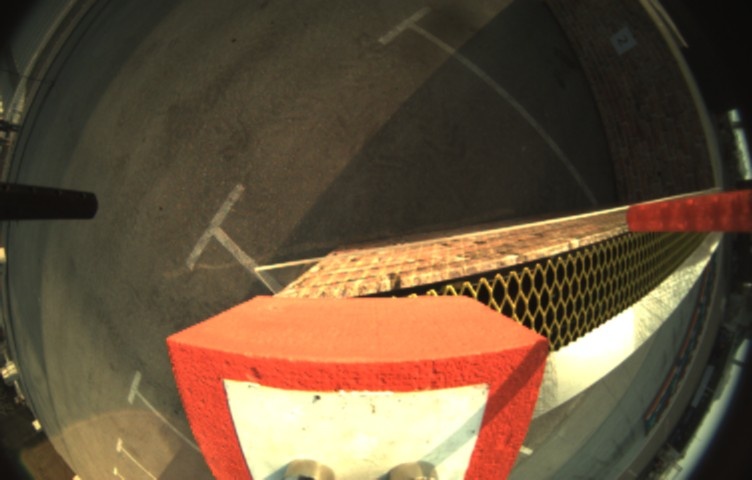}};
      \begin{scope}[x={(img.south east)},y={(img.north west)}]
        \node[partlabel,text=cyan] (gplacement_release_point_label) at (img.north west) {(c) Release point};
        \draw (img.north east) rectangle (img.south west);
      \end{scope}
    \end{tikzpicture} &
    \begin{tikzpicture}
      \node[anchor=south west,inner sep=0] (img) at (0,0) {\includegraphics[width=0.25\linewidth]{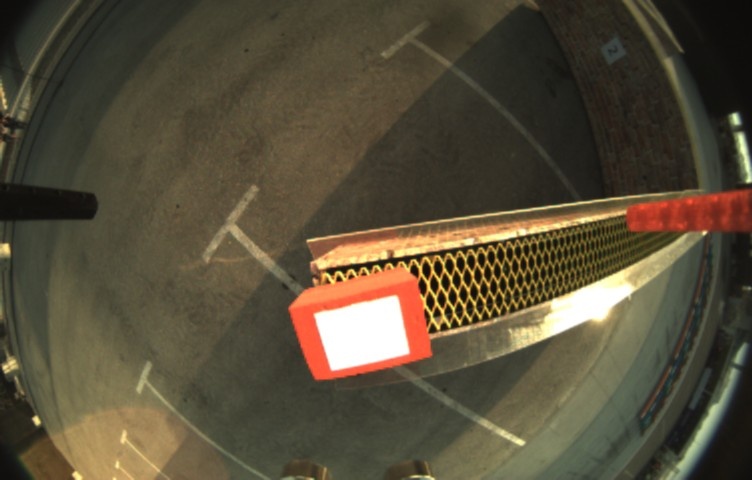}};
      \begin{scope}[x={(img.south east)},y={(img.north west)}]
        \node[partlabel,text=black] (placement_end_label) at (img.north west) {(d) Placement end};
        \draw (img.north east) rectangle (img.south west);
      \end{scope}
    \end{tikzpicture} \\
    \multicolumn{4}{c}{
      \subfloat{
        \begin{tikzpicture}
          \node[anchor=south west,inner sep=0] (img) at (0,0) {\includegraphics[width=0.22\linewidth]{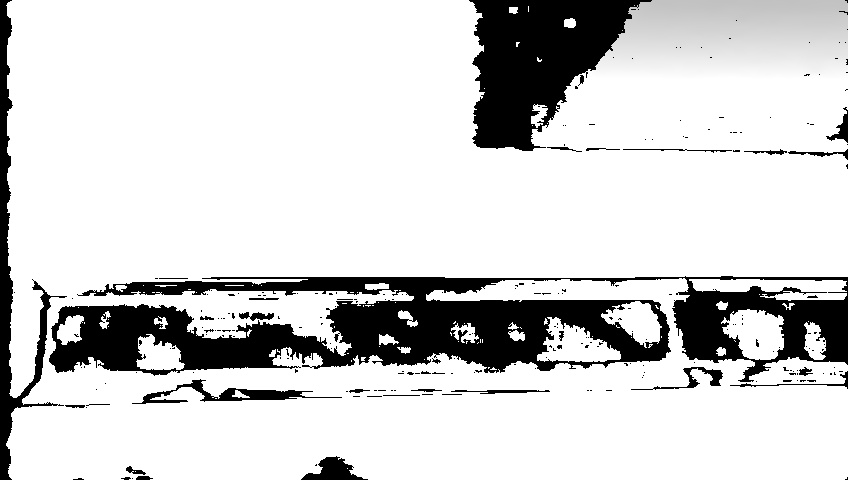}};
          \begin{scope}[x={(img.south east)},y={(img.north west)}]
            \node[partlabel,text=red] (placement_align_start_depth_label) at (img.north west) {(e) Align start};
            \draw (img.north east) rectangle (img.south west);
          \end{scope}
        \end{tikzpicture}
      }
      \subfloat{
        \begin{tikzpicture}
          \node[anchor=south west,inner sep=0] (img) at (0,0) {\includegraphics[width=0.22\linewidth]{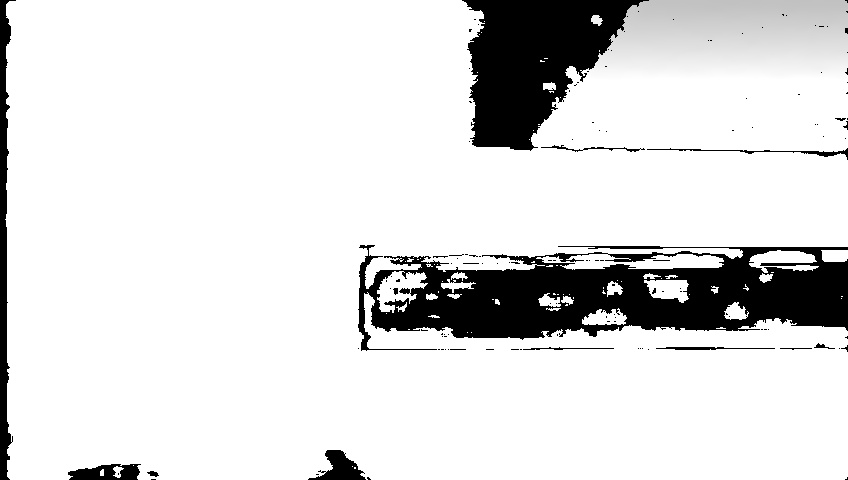}};
          \begin{scope}[x={(img.south east)},y={(img.north west)}]
            \node[partlabel,text=orange] (placement_placing_start_depth_label) at (img.north west) {(f) Placing start};
            \draw (img.north east) rectangle (img.south west);
          \end{scope}
        \end{tikzpicture}
      }
      \subfloat{
        \begin{tikzpicture}
          \node[anchor=south west,inner sep=0] (img) at (0,0) {\includegraphics[width=0.22\linewidth]{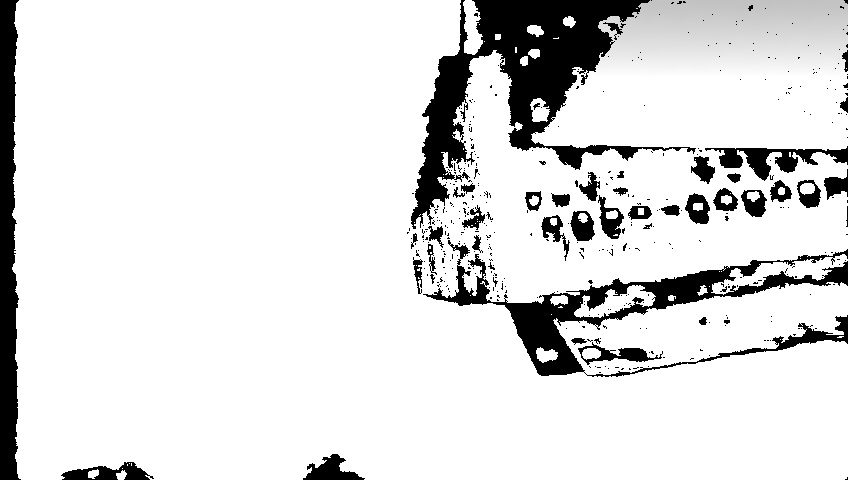}};
          \begin{scope}[x={(img.south east)},y={(img.north west)}]
            \node[partlabel,text=cyan] (placement_release_point_depth_label) at (img.north west) {(g) Release point};
            \draw (img.north east) rectangle (img.south west);
          \end{scope}
        \end{tikzpicture}
      }
      \subfloat{
        \begin{tikzpicture}
          \node[anchor=south west,inner sep=0] (img) at (0,0) {\includegraphics[width=0.22\linewidth]{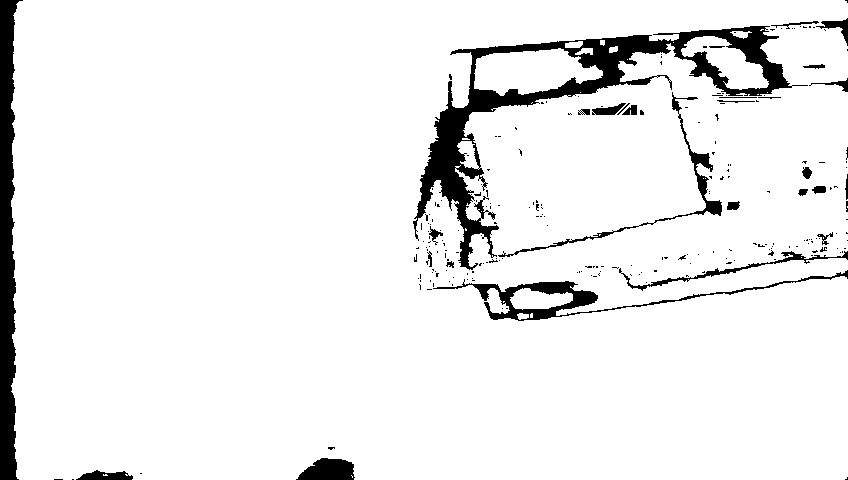}};
          \begin{scope}[x={(img.south east)},y={(img.north west)}]
            \node[partlabel,text=black] (placement_end_depth_label) at (img.north west) {(h) Placement end};
            \draw (img.north east) rectangle (img.south west);
          \end{scope}
        \end{tikzpicture}
      }
    }
  \end{tabular}
  \caption{Placement of a red brick on the beginning of the first wall channel shown with placement stages and wall detections, together with images from the RGB camera (a)--(d) and depth camera (e)--(h). Videos from experimental testing and from the competition are available at \url{http://mrs.felk.cvut.cz/mbzirc-2020-uav-wall}.}
  \label{fig:placement}
\end{figure*}
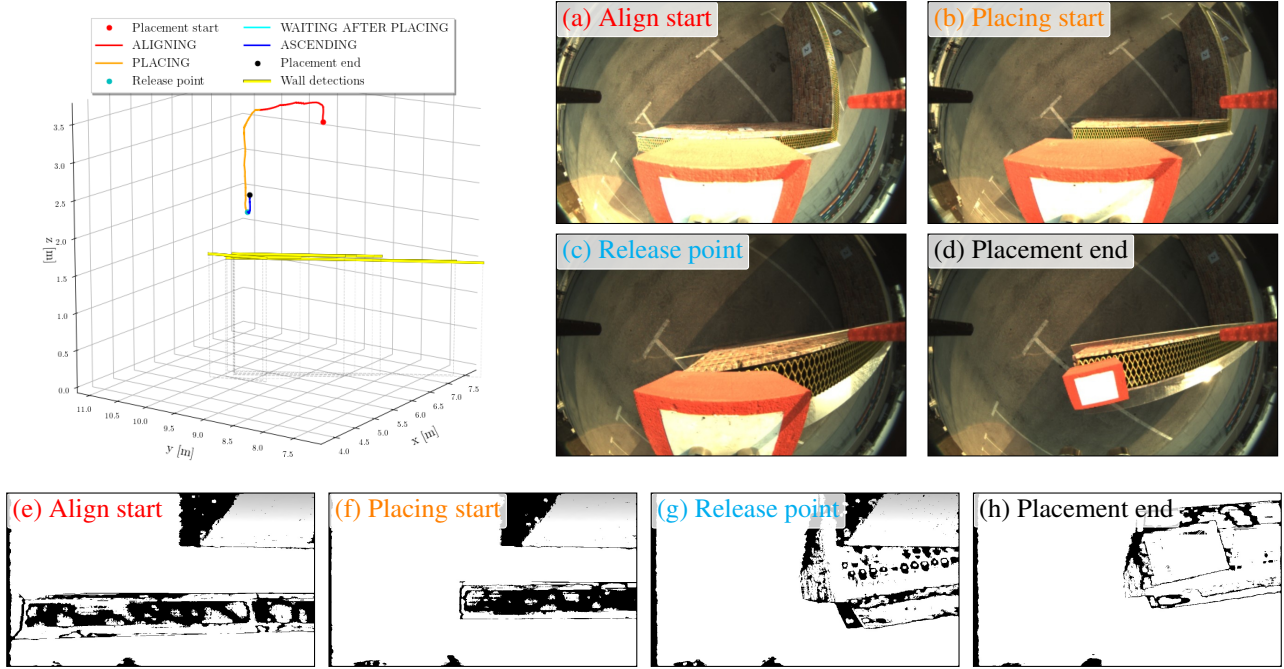

Brick placement together with grasping is one of the most important wall building capabilities.
Figure~\ref{fig:placement} depicts the successful placement of the same brick being grasped in \reffig{fig:grasping}.
After successful grasping, the \ac{UAV} flies above its assigned wall channel and switches to the lower-level placement state machine using place detection (both described in \refsec{sec:placing}) to guide the \ac{UAV} above the release point on the appropriate location of the wall.
Note that the pattern (i.e. sequence of bricks in both layers) forming individual channels was given for each trial.
The brick building sequence was planed as a consecutive placement of a brick into the next unoccupied position in the wall segment --- either the leftmost position in the completely-free segment or neighbouring position to the rightmost brick on the segment.
Figure~\ref{fig:placement} shows the stages of the lower-level placement state machine together with the \ac{UAV} positions and detections of the wall channel during alignment and placing states.
Additionally, the images from both the RGB mvBlueFOX camera and the depth images from the RealSense camera are show in various stages of placement.


In \reffig{fig:placement}, it can be seen that placement starts at approximately the middle of the assigned wall channel. During the alignment state, the \ac{UAV} moves along the channel to the leftmost position on the empty wall.
The wall detections are clearly shown to be of various lengths as the smaller portion of the wall is visible once descending during the placing state.
However, the left corner of the wall is measured during the detections with a mean position for x, y, heading being \SI{5.374}{\meter}, \SI{9.790}{\meter}, \SI{-1.110}{\radian}, respectively, with corresponding standard deviation of \SI{0.139}{\meter}, \SI{0.093}{\meter}, \SI{0.113}{\radian}, respectively.


The brick placing state machine is comparably simpler than the grasping state machine, since the grasping state machine needs to cover various failure stages during the grasping process.
This was required to deal with failures in the grasping stage, because the \ac{UAV} adds no value to the mission outcome if it does not succeed with grasping.
Moreover, the grasping manoeuvre is sensitive to control accuracy and timing while also being more dangerous for the \ac{UAV}.
On the other hand, placing allows for a significant slack in the control of the \ac{UAV} thanks to the width of the wall channel and the possibility of dropping the brick from a higher height without any physical interaction with the wall.
The wall detection also worked more reliably thanks to more prominent features in sensory input as the \ac{UAV} rarely lost the wall from its field of vision during testing.
Lastly, we did not consider any possible correcting action for instances of improperly placed bricks.

\subsection{Wall building performance}

\begin{figure*}[!htb]
  \centering
  \includegraphics[width=0.60\linewidth]{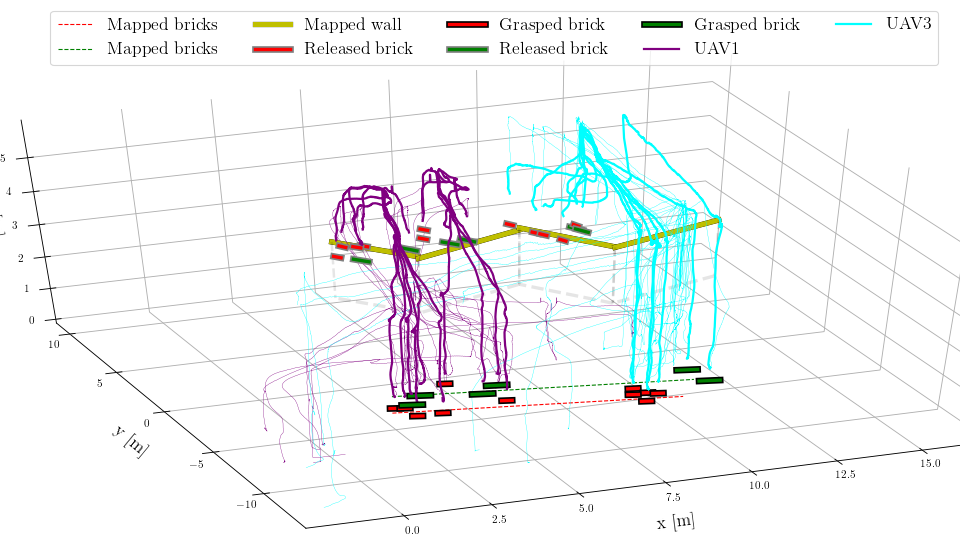}
  \caption{Visualization of wall building performance during the first trial with shown grasping/placing positions and highlighted UAV trajectories by the thicker lines when carrying a brick.}
  \label{fig:all_uavs_odom}
\end{figure*}

Finally, the performance of the wall building stage using the proposed system is depicted in \reffig{fig:all_uavs_odom} showing the recorded positions of both the UAV1 and UAV3 throughout the first trial with four performed restarts.
All 17 grasped bricks are shown with their respective release positions, however only ten bricks stayed on the wall without bouncing off as detailed in \reftab{tab:results_bricks}.
The recorded positions of the \acp{UAV} shown do not include the initial scanning of UAV1 already discussed in the \refsec{sec:results_scanning}. Instead, the positions of the \acp{UAV} once carrying bricks is highlighted.
The mapped wall and brick locations are based on the scanning in the second restart as already shown in \reffig{fig:scanning} and \reffig{fig:mapping_detections}.


See video of the CTU-UPenn-NYU team \url{https://youtu.be/1-aRtSarYz4} with summary of the preparations in the Abu Dhabi desert, of competition rehearsals, and of the actual Challenge 2 \ac{MBZIRC} 2020 competition trials.

\subsection{Lessons learned}


The most important factor driving development was the need for safety and reliability of the multi-\ac{UAV} system.
Interaction with the environment during grasping and placing is potentially dangerous and can easily damage the \ac{UAV}.
The use of real-time weight and force estimation for the detection of potentially dangerous situations was of significant benefit.
A common approach of relying solely on \ac{UAV} position estimation to drive the decision-making process would not be sufficient.

One of the tunable parameters of the grasping/placing maneuvers was the speed of the descent.
Speed too slow increases the overall duration of the \ac{UAV} being in a potentially dangerous location and allows the ground effect to build up (aerodynamic effects caused by the rotor downwash close to a ground).
Alternatively, too fast of speed increases the risk of damaging the \ac{UAV} due to the sudden bump caused by interaction with the brick, the wall, or the ground.
On several occasions during our preparations, we experienced a complete stall of motors due to the sudden impact at higher speed, which subsequently caused a loss of onboard power and an uncontrollable, unstoppable tumbling of the \ac{UAV}.
Therefore, we advise caution when working with \acp{UAV} if sharp acceleration spikes may be transmitted to the \ac{UAV} body.
Finally, we settled on the descending speed of \SI{0.25}{\meter\per\second} which showed to be the most reliable and rewarding.

This proposed wall building system depends on successful arena scanning and creation of a topological map used for planning further brick pickup and placement.
As such, the brick and wall detections need to be robust and without excessive false positives that could influence the topological map creation based on statistical analysis of detections.
The task was further challenging due to the stack of UGV bricks present that could not be used safely by \acp{UAV} and thus had to be recognized among the detections.
During the preparations, a rather high number of false detections forced implementing significant detection filtering during scanning, requiring a minimal number of corrections in the detection map and used the iterative median filtering of detections to remove outliers.
However, during the competition rehearsals, the creation of the topological map had to be further fine-tuned to, e.g., filter out already placed bricks from previous restarts.

Finally, the \ac{GPS} drift significantly influenced the entire system deployment as arena borders were defined in \ac{GPS} coordinates, with either the brick stack or the wall channels possibly placed too close to the borders.
This prevented \acp{UAV} from flying too close to the net-protected borders and suggests that an additional \ac{LiDAR} or camera-based detection of the border (or even sensor-based fix of the GPS drift) would significantly improve deployment robustness in similar competitions.
An \ac{RTK}-based localization is technically a possible solution to this problem, but it was not used by the team due to the penalization of \ac{RTK} in scoring.
However, after discussion with potential industrial partners, a solution using both the \ac{RTK} \ac{GPS} and the onboard local sensors would be preferred.
Important construction locations could be pre-measured using the \ac{RTK} (as it is nowadays common on construction sites). Combining this with the proposed onboard sensor capabilities would yield a robust cognitive system capable of reacting to changes in the environment.

The execution time of brick detection for visual servoing is critical for smooth real-time \ac{UAV} control and navigation.
The Intel NUC computer, by its design similar to laptop hardware, had a power-saving mode that caused irregular execution times of our methods.
This problem was detected before the competition and resolving it improved the robustness of the whole system significantly.



\section{Conclusions\label{sec:conclusions}}

In this paper, an autonomous system developed by CTU-UPenn-NYU team for wall building with a team of \acp{UAV} was introduced.
The examined task was part of Challenge~2 of the \ac{MBZIRC} 2020 where three \acp{UAV} were assigned to find, pickup, and place color-typed bricks on a prepared wall structure.
The goal of the task was to maximize collected points for placing the bricks while following the prescribed wall pattern.
This paper presents the key parts of the \ac{UAV} system developed for the competition, including the \ac{UAV} control system, the algorithms for brick and wall detection, the single robot state machine, and the multi-robot distributed approach for the task.
The core autonomous capabilities of scanning the arena for wall/brick locations, autonomous grasping using visual servoing technique, and precise placement of bricks on the wall structure are described in detail.
We further report the experimental results achieved during the competition trials showing the performance of the core autonomous capabilities and of the entire system.
The proposed approach performed the best among all participants with 22 successfully grasped bricks in which 13 of these bricks were successfully placed during the two trials of the Challenge 2 \ac{MBZIRC} competition.
The entire system for the wall building task is open-sourced for the community to be used for possible deployment and future development.
It can also serve as a useful reference for future robotic challenges, such as the MBZIRC that indeed serves as a great verification of robotic research.



\section*{Acknowledgements}

\begin{figure}
  \centering
  \includegraphics[width=0.99\linewidth]{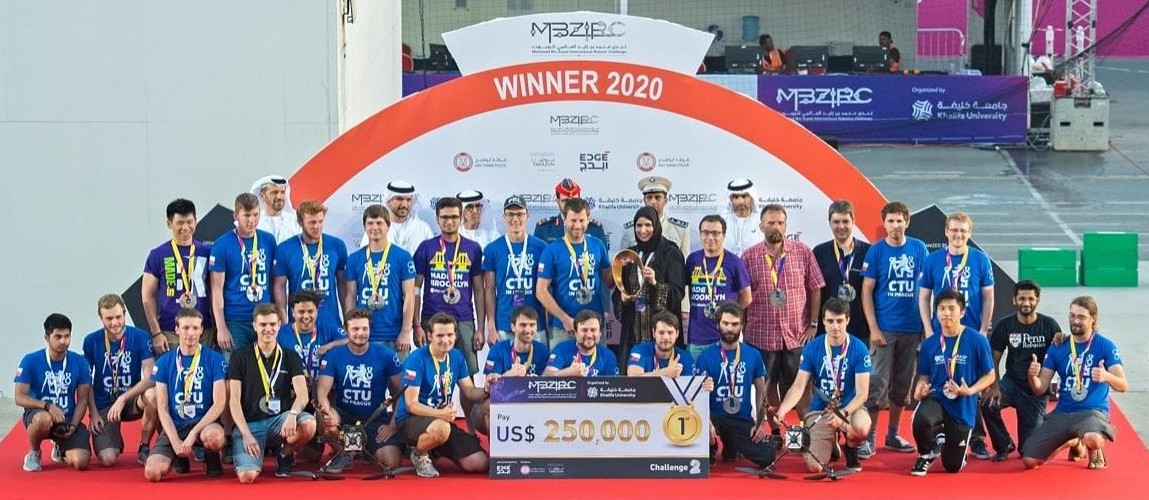}
  \caption{CTU-UPenn-NYU team after winning the wall building Challenge 2 of \ac{MBZIRC} 2020.}
  \label{fig:ch2_stand}
\end{figure}

We thank the \acl{UPenn} and the \acl{NYU} for the collaboration as the members of the CTU-UPenn-NYU team (see \reffig{fig:ch2_stand}).
This work would not have been possible without the tireless dedication of all the team members and support and patience from their families.
The presented work has been supported by Khalifa University and by CTU grant no. SGS20/174/OHK3/3T/13.

\bibliographystyle{elsarticle-num}
\bibliography{bib_common,bib_tomas}

\begin{thebibliography}{10}
\expandafter\ifx\csname url\endcsname\relax
  \def\url#1{\texttt{#1}}\fi
\expandafter\ifx\csname urlprefix\endcsname\relax\def\urlprefix{URL }\fi
\expandafter\ifx\csname href\endcsname\relax
  \def\href#1#2{#2} \def\path#1{#1}\fi

\bibitem{Tatum2017UASapplicationsinConstruction}
M.~C. Tatum, J.~Liu, Unmanned aircraft system applications in construction,
  Procedia Engineering 196 (2017) 167--175.

\bibitem{ham2016visual}
Y.~Ham, K.~K. Han, J.~J. Lin, M.~Golparvar-Fard, {Visual monitoring of civil
  infrastructure systems via camera-equipped Unmanned Aerial Vehicles (UAVs): a
  review of related works} 4~(1) (2016) 1--8.

\bibitem{Howard2018uavinconstruction}
J.~Howard, V.~Murashov, C.~M. Branche, Unmanned aerial vehicles in construction
  and worker safety, American Journal of Industrial Medicine 61~(1) (2018)
  3--10.
\newblock \href {https://doi.org/10.1002/ajim.22782}
  {\path{doi:10.1002/ajim.22782}}.

\bibitem{mbzirc_description_page}
{Khalifa University}, Mohamed bin zayed international robotics challenge 2020,
  available online, {\url{https://www.mbzirc.com/challenge/2020}}, accessed on,
  citied on 2020/11/16.

\bibitem{baca2018model}
T.~Baca, D.~Hert, et~al., {Model Predictive Trajectory Tracking and Collision
  Avoidance for Reliable Outdoor Deployment of Unmanned Aerial Vehicles}, in:
  2018 IEEE/RSJ IROS, IEEE, 2018, pp. 1--8.

\bibitem{lee2010geometric}
T.~Lee, M.~Leoky, et~al., {Geometric tracking control of a quadrotor UAV on
  SE(3)}, in: {2010 IEEE CDC}, IEEE, 2010, pp. 5420--5425.

\bibitem{baca2020mrs}
T.~Baca, M.~Petrlik, M.~Vrba, V.~Spurny, R.~Penicka, D.~Hert, M.~Saska, {The
  MRS UAV System: Pushing the Frontiers of Reproducible Research, Real-world
  Deployment, and Education with Autonomous Unmanned Aerial Vehicles},
  {submitted to JINT} (8 2020).
\newblock \href {http://arxiv.org/abs/2008.08050v2}
  {\path{arXiv:2008.08050v2}}.

\bibitem{zink2015bridgeinspection}
J.~Zink, B.~Lovelace, Unmanned aerial vehicle bridge inspection demonstration
  project, Tech. rep. (2015).

\bibitem{lindsey2011construction_cubic_structures}
Q.~Lindsey, D.~Mellinger, V.~Kumar, Construction of cubic structures with
  quadrotor teams, Proc. Robotics: Science \& Systems VII (2011).

\bibitem{lindsey2013distributed}
Q.~Lindsey, et~al., Distributed construction of truss structures, in:
  Algorithmic Foundations of Robotics X, Springer, 2013, pp. 209--225.

\bibitem{augugliaro2013building}
F.~Augugliaro, A.~Mirjan, et~al., {Building tensile structures with flying
  machines}, in: 2013 IEEE/RSJ IROS, IEEE, 2013, pp. 3487--3492.

\bibitem{mirjan2016building}
A.~Mirjan, F.~Augugliaro, R.~D’Andrea, F.~Gramazio, M.~Kohler, Building a
  bridge with flying robots, in: Robotic Fabrication in Architecture, Art and
  Design 2016, Springer, 2016, pp. 34--47.

\bibitem{alejo2014collisionfree}
D.~Alejo, J.~A. Cobano, et~al., {Collision-Free 4D Trajectory Planning in
  Unmanned Aerial Vehicles for Assembly and Structure Construction}, Journal of
  Intelligent \& Robotic Systems 73~(1) (2014) 783--795.

\bibitem{augugliaro2014flight}
F.~Augugliaro, S.~Lupashin, M.~Hamer, C.~Male, M.~Hehn, M.~W. Mueller, J.~S.
  Willmann, et~al., The flight assembled architecture installation: Cooperative
  construction with flying machines, IEEE Control Systems Magazine 34~(4)
  (2014) 46--64.

\bibitem{ARCAS_project}
{Aerial Robotics Cooperative Assembly system}, retrieved from
  \url{http://www.arcas-project.eu}, 2020/08/28 (2020).

\bibitem{ruggiero2018aerial}
F.~Ruggiero, V.~Lippiello, A.~Ollero, {Aerial manipulation: A literature
  review}, IEEE Robotics and Automation Letters 3~(3) (2018) 1957--1964.

\bibitem{Kondak13_Aerial_arm_manipulation}
K.~Kondak, K.~Krieger, A.~Albu-Schaeffer, M.~Schwarzbach, M.~Laiacker, I.~Maza,
  et~al., {Closed-Loop Behavior of an Autonomous Helicopter Equipped with a
  Robotic Arm for Aerial Manipulation Tasks}, International Journal of Advanced
  Robotic Systems 10~(2) (2013) 145--154.

\bibitem{kondak2014aerial}
K.~Kondak, F.~Huber, M.~Schwarzbach, M.~Laiacker, et~al., Aerial manipulation
  robot composed of an autonomous helicopter and a 7 degrees of freedom
  industrial manipulator, in: 2014 IEEE ICRA, IEEE, 2014, pp. 2107--2112.

\bibitem{ryll20176d}
M.~Ryll, G.~Muscio, F.~Pierri, E.~Cataldi, G.~Antonelli, F.~Caccavale,
  A.~Franchi, {6D physical interaction with a fully actuated aerial robot}, in:
  2017 IEEE ICRA, IEEE, 2017, pp. 5190--5195.

\bibitem{munoz2015assembly}
J.~Munoz-Morera, I.~Maza, C.~J. Fernandez-Aguera, F.~Caballero, A.~Ollero,
  {Assembly planning for the construction of structures with multiple UAS
  equipped with robotic arms}, in: 2015 IEEE ICUAS, IEEE, 2015, pp. 1049--1058.

\bibitem{thomas2014toward}
J.~Thomas, G.~Loianno, et~al., Toward image based visual servoing for aerial
  grasping and perching, in: 2014 IEEE ICRA, IEEE, 2014, pp. 2113--2118.

\bibitem{ramon2017detection}
P.~Ramon~Soria, B.~C. Arrue, A.~Ollero, {Detection, location and grasping
  objects using a stereo sensor on UAV in outdoor environments}, Sensors 17~(1)
  (2017) 103.

\bibitem{gawel2017aerial}
A.~Gawel, M.~Kamel, T.~Novkovic, J.~Widauer, D.~Schindler, B.~P.
  Von~Altishofen, R.~Siegwart, J.~Nieto, {Aerial picking and delivery of
  magnetic objects with MAVs}, in: 2017 IEEE ICRA, IEEE, 2017, pp. 5746--5752.

\bibitem{feng2020packages}
K.~Feng, W.~Li, S.~Ge, F.~Pan, {Packages delivery based on marker detection for
  UAVs}, in: 2020 IEEE CCDC, IEEE, 2020, pp. 2094--2099.

\bibitem{loianno2018localization}
G.~Loianno, V.~Spurny, J.~Thomas, T.~Baca, D.~Thakur, D.~Hert, R.~Penicka,
  T.~Krajnik, A.~Zhou, A.~Cho, M.~Saska, et~al., {Localization, Grasping, and
  Transportation of Magnetic Objects by a team of MAVs in Challenging Desert
  like Environments}, IEEE Robotics and Automation Letters 3~(3) (2018)
  1576--1583.

\bibitem{spurny2019cooperative}
V.~Spurny, T.~Baca, M.~Saska, R.~Penicka, T.~Krajnik, J.~Thomas, D.~Thakur,
  G.~Loianno, et~al., {Cooperative Autonomous Search, Grasping and Delivering
  in a Treasure Hunt Scenario by a Team of UAVs}, {Journal of Field Robotics}
  {36}~({1}) (2019) {125--148}.

\bibitem{castano2019robotics}
A.~R. Castano, F.~Real, P.~Ramon-Soria, J.~Capitan, V.~Vega, B.~C. Arrue,
  et~al., {Al-Robotics team: A cooperative multi-unmanned aerial vehicle
  approach for the Mohamed Bin Zayed International Robotic Challenge}, Journal
  of Field Robotics 36~(1) (2019) 104--124.

\bibitem{nicadrone_epmv3}
{NicaDrone}, {Electro Permanent Magnet OpenGrab EPM V3}, retrieved August 22,
  2020, from https://nicadrone.com/products/epm-v3 (2020).

\bibitem{bahnemann2019eth}
R.~Bahnemann, M.~Pantic, M.~Popovic, D.~Schindler, M.~Tranzatto, M.~Kamel,
  M.~Grimm, J.~Widauer, R.~Siegwart, J.~Nieto, {The ETH-MAV Team in the MBZ
  International Robotics Challenge}, Journal of Field Robotics 36~(1) (2019)
  78--103.

\bibitem{kamel2017robust}
M.~Kamel, J.~Alonso-Mora, R.~Siegwart, J.~Nieto, {Robust Collision Avoidance
  for Multiple Micro Aerial Vehicles Using Nonlinear Model Predictive Control},
  in: 2017 IEEE/RSJ IROS, IEEE, 2017, pp. 236--243.

\bibitem{beul2019team}
M.~Beul, M.~Nieuwenhuisen, J.~Quenzel, R.~A. Rosu, J.~Horn, D.~Pavlichenko,
  S.~Houben, S.~Behnke, {Team NimbRo at MBZIRC 2017: Fast landing on a moving
  target and treasure hunting with a team of micro aerial vehicles}, Journal of
  Field Robotics 36~(1) (2019) 204--229.

\bibitem{beul2017fast}
M.~Beul, S.~Behnke, Fast full state trajectory generation for multirotors, in:
  2017 IEEE ICUAS, IEEE, 2017, pp. 408--416.

\bibitem{ROSMelodic}
{Open Robotics}, {Robotic Operating System}, retrieved July 22, 2020, from
  https://www.ros.org (2020).

\bibitem{schillinger2016human}
P.~Schillinger, S.~Kohlbrecher, O.~von Stryk, Human-robot collaborative
  high-level control with application to rescue robotics, in: 2016 IEEE ICRA,
  IEEE, 2016, pp. 2796--2802.

\bibitem{Bohren2010SMACH}
J.~Bohren, S.~Cousins, The smach high-level executive [ros news], IEEE Robotics
  Automation Magazine 17~(4) (2010) 18--20.

\bibitem{nimbro}
M.~Schwarz, nimbro\_network - ros transport for high-latency, low-quality
  networks, available online, \url{https://github.com/AIS-Bonn/nimbro_network},
  accessed on 2020/11/16.

\bibitem{CPPsurvey2013}
E.~Galceran, M.~Carreras, A survey on coverage path planning for robotics,
  Robotics and Autonomous Systems 61~(12) (2013) 1258--1276.

\bibitem{realsense_depth}
A.~Grunnet-Jepsen, J.~N. Sweetser, J.~Woodfill, {Best-Known-Methods for Tuning
  Intel{\textregistered} RealSense™ D400 Depth Cameras for Best Performance},
  Tech. rep., Intel Corporation: Satan Clara, CA, USA (2018).

\bibitem{opencv_library}
G.~Bradski, {The OpenCV Library}, Dr. Dobb's Journal of Software Tools (2000).

\bibitem{Ocam2006}
D.~Scaramuzza, A.~Martinelli, R.~Siegwart, A toolbox for easily calibrating
  omnidirectional cameras, in: 2006 IEEE/RSJ IROS, IEEE, 2006, pp. 5695--5701.

\bibitem{Reynolds2009_GMM}
D.~Reynolds, Gaussian mixture models, Encyclopedia of Biometrics 741 (2009)
  659--663.

\bibitem{Jolliffe2011_PCA}
I.~Jolliffe, Principal Component Analysis, 2011, pp. 1094--1096.

\bibitem{stibinger2020mobile}
P.~Stibinger, G.~Broughton, F.~Majer, Z.~Rozsypalek, A.~Wang, K.~Jindal,
  A.~Zhou, D.~Thakur, G.~Loianno, T.~Krajnik, M.~Saska, {Mobile Manipulator for
  Autonomous Localization, Grasping and Precise Placement of Construction
  Material in a Semi-structured Environment}, Submitted to Robotics and
  Automation Letters (2020).
\newblock \href {http://arxiv.org/abs/2011.07972} {\path{arXiv:2011.07972}}.

\end{thebibliography}


\end{document}